\documentclass[11pt, oneside]{article}   	
\usepackage{geometry}                		
\pdfoutput=1						

\usepackage[parfill]{parskip}    			
\usepackage{graphicx}				
\usepackage{amssymb}
\usepackage{amsmath}

\usepackage{bm} 					
\usepackage{setspace}
\usepackage{natbib}					
\setcitestyle{aysep={}}  				
\usepackage[ngerman, english]{babel}
\usepackage{booktabs} 				
\usepackage[table]{xcolor}			
\usepackage{tcolorbox}
\newcommand{\maxf}[1]{{\cellcolor[gray]{0.8}} #1}
\usepackage{multirow} 
\usepackage{arydshln} 
\usepackage{enumitem} 
\setlist{topsep=0pt, leftmargin=*}
\usepackage{float}
\usepackage{titlesec} 
\titlespacing*{\section}
{0pt}{2ex}{0ex}
\titlespacing*{\subsection}
{0pt}{2ex}{0ex}
\titlespacing*{\subsubsection}
{0pt}{2ex}{0ex}

\usepackage{xurl} 
\usepackage[bookmarks]{hyperref}
\hypersetup{colorlinks=true, citecolor=black, linkcolor=black, urlcolor=black} 

\usepackage{sectsty} 
\usepackage[font={sf, small}, labelfont={sf, bf, small}, labelsep=period]{caption}
\allsectionsfont{\sffamily}

\title{{\bfseries\sffamily{Introduction to Neural Transfer Learning with Transformers for Social Science Text Analysis}}}
\author{Sandra Wankm\"uller \\
\emph{Ludwig-Maximilians-Universit\"at M\"unchen} \\
sandra.wankmueller@gsi.lmu.de \\
\small{https://orcid.org/0000-0002-4003-1704}}

\date{}							

\begin{document}
\maketitle

\setlength{\abovedisplayskip}{3pt}
\setlength{\belowdisplayskip}{3pt}

{\bfseries\sffamily{Abstract.}}
Transformer-based models for transfer learning have the potential to achieve high prediction accuracies on text-based supervised learning tasks with relatively few training data instances. These models are thus likely to benefit social scientists that seek to have as accurate as possible text-based measures but only have limited resources for annotating training data. To enable social scientists to leverage these potential benefits for their research, this paper explains how these methods work, why they might be advantageous, and what their limitations are. Additionally, three Transformer-based models for transfer learning, BERT \citep{Devlin2019}, RoBERTa \citep{Liu2019}, and the Longformer \citep{Beltagy2020}, are compared to conventional machine learning algorithms on three applications. Across all evaluated tasks, textual styles, and training data set sizes, the conventional models are consistently outperformed by transfer learning with Transformers, thereby demonstrating the benefits these models can bring to text-based social science research.


{\bfseries\sffamily{Source Code.}} The code of this study is openly available at \url{https://doi.org/10.6084/m9.figshare.14394173}

{\bfseries\sffamily{Acknowledgements.}} I am very grateful to Paul W.~Thurner, Christian Heumann, Matthias A{\ss}enmacher, the participants of the colloquium at the chair of Paul W.~Thurner, and three anonymous reviewers for their highly valuable guidance and helpful comments on this work.

{\bfseries\sffamily{Funding.}} This work was supported by a scholarship I received from the German Academic Scholarship Foundation.

\newpage

\section{Introduction: Why Neural Transfer Learning with Transformers?} \label{sec:intro}

In social science, supervised learning techniques have been employed to measure a vast range of application-specific (and often complex, latent, and multidimensional) concepts from texts, such as e.g.~tonality \citep{Rudkowsky2018, Barbera2020, Fowler2020}, inequality \citep{Nelson2021}, populism \citep{DiCocco2021}, attitudes \citep{Ceron:2014wr, Mitts2019}, policy topics \citep{Osnabruegge2021, Seboek2020}, and events \citep{DOrazio2014, Zhang2019, Muchlinski2021}. In such supervised learning settings, the training data encode how the concept (e.g.~attitude, inequality, event) is to be operationalized and the text analysis method is the measurement method that is deployed to assign the textual units to the values of the variable. 

If a researcher is applying a supervised learning method on text data for the purpose of measuring an a priori-specified concept, her aim---as in any measurement process---will be to have a valid measure that captures the concept it is devised to measure. And consequently---because when working with text data humans are usually seen as the ``the ultimate arbiter of the `validity' of any research exercise'' \citep[p.~470]{Benoit2019b}---the aim for the researcher is to have a supervised learning technique that as closely as possible can imitate human codings \citep[p.~270, 279]{Grimmer:2013fp}.\footnote{\citet[p.~470]{Benoit2019b} points out that research indicates that humans are not very reliable coders of text data \citep[see also e.g.][]{Mikhaylov2012, Ennser2018}. This, in turn, raises the question of how valid human judgments can be \citep[p.~553]{Song2020}. Nevertheless, in this study---and in concordance with the literature (\citealp[p.~470]{Benoit2019b}; \citealp[p.~204-205]{Nelson2021})---the comparison of human codings to the predictions of a supervised learning method is considered the best available procedure for validation.} After having trained a model on human annotated training data, the researcher thus will hope that the trained model as accurately as possible predicts human codings on data that have not been used in training \citep[p.~271, 279]{Grimmer:2013fp}. If this is the case and hence the model can be said to generalize well, this indicates that the model's predictions will provide a valid measure of the concept under study \citep[p.~271, 279]{Grimmer:2013fp}.\footnote{This focus on prediction performance is a major deviation from the usual social science focus on making causal inferences. In a causal inference setting, modeling is theory-based and interpretable models are used to identify the effects of single independent variables. But in order to test hypotheses about causal relations between concepts, the concepts have to be translated into measurable variables that constitute valid measures of the concepts under study. And if for the process of measurement a supervised learning method is used, then the goal is to as closely as possible replicate human coding as this indicates validity \citep[p.~271, 279]{Grimmer:2013fp}. So here, for the very purpose of measurement, the aim is not causal inference but precise prediction.}

In the field of natural language processing (NLP), the usage of deep learning models (as compared to conventional machine learning algorithms) has enabled researchers to learn better generalizing mappings from textual inputs to task-specific outputs and hence has enabled researchers to more accurately perform a wide spectrum of prediction tasks such as text classification, machine translation, or reading comprehension (\citealp[p.~347-348]{Goldberg2016}; \citealp{Ruder2020b}). Despite the fact that deep learning techniques tend to exhibit higher prediction accuracies in text-based supervised learning tasks compared to traditional machine learning algorithms \citep{Socher2013, Iyyer2014, Budhwar2018, Ruder2020b}, they are not yet a standard tool for social science researchers that use supervised learning for text analysis. Although there are exceptions \citep[e.g.][]{Rudkowsky2018, Zhang2019, Amsalem2020, Chang2019, Muchlinski2021, Wu2021}, 
social scientists typically resort to bag-of-words-based representations of texts that serve as an input to conventional machine learning models such as support vector machines (SVMs), naive Bayes, random forests, boosting algorithms, or regression with regularization 
\citep[see e.g.][]{Diermeier2011, Colleoni2014, DOrazio2014, Ceron2015, Theocharis2016, Welbers2017, Kwon2018, Greene2019, Katagiri2019, Mitts2019, Pilny2019, Ramey2019, RonaTas2019, Anastasopoulos2020, Miller2020, Park2020, Barbera2020, DiCocco2021, Fowler2020, Osnabruegge2021, Seboek2020}.\footnote{This is not to say that social scientists would not have started to leverage the foundations of deep learning approaches in NLP: During the last years, the use of real-valued vector representations of terms, known as word embeddings, enabled social scientists to explore new research questions or to study old research questions by new means \citep[e.g.][]{Rheault2016, Han2018, Kozlowski2019, Rheault2020, Rodman2020, Watanabe2020}. Moreover, there is a small but increasing number of publications in social science journals that apply deep neural networks to texts \citep[e.g.][]{Rudkowsky2018, Zhang2019, Amsalem2020, Chang2019, Muchlinski2021, Wu2021}. Yet applications of deep neural networks (let alone deep neural networks plus transfer learning) are not widely used by social scientists. And thus, implementations of deep neural networks and modern NLP techniques on texts that are relevant for social science research up til now are mostly conducted by research teams that are not primarily social science trained \citep[see e.g.][]{Iyyer2014, Zarrella2016, Glavas2017, Budhwar2018, Meidinger2021} and/or are published via platforms and venues (e.g.~important NLP conferences such as the EMNLP, ACL, or NAACL) that social scientists typically do not closely monitor \citep[e.g.][]{Kim2021a, Rehbein2021a, Rehbein2021}.}  

One among several likely reasons why deep learning methods so far have not been widely used for text-based supervised learning tasks by social scientists might be that deep learning models have considerably more parameters to be learned in training than classic machine learning models. Consequently, deep learning models are computationally highly intensive and require substantially larger numbers of training examples. \citet[p.~20]{Goodfellow2016} stated that ``As of 2016, a rough rule of thumb is that a supervised deep learning algorithm will generally achieve acceptable performance with around 5,000 labeled examples per category''.\footnote{Yet how much training data instances are really needed depends on the width and depth of the deep neural network, the task, and training data quality. Thus, precise numbers on the amounts of parameters and required training examples cannot be specified. To nevertheless put the sizes in relation, note that an SVM with a linear kernel that learns to construct a hyperplane in a 3,000-dimensional feature space which separates instances into two categories based on 1,000 support vectors has around 3 million parameters. 
The Transformer-based models presented in this article, in contrast, have well above 100 million parameters.} For research questions relating to domains in which it is difficult to access or label large enough numbers of training data instances, deep learning becomes infeasible or prohibitively costly.

Recent developments within NLP on transfer learning alleviate this problem. Transfer learning is a set of learning procedures in which knowledge that has been learned from training on a source task in a source domain is used to improve learning on the target task in the target domain (where the target task is the task of interest that a researcher

\begin{tcolorbox}[width=\linewidth, sharp corners=all, colback=white!95!black]
{\bfseries\sffamily{What kind of NLP tasks are there?}} In the field of NLP, a large spectrum of diverse tasks are addressed. There are NLP tasks that operate at the linguistic level (e.g.~part-of-speech (POS) tagging, syntactic parsing) \citep[p.~4-11]{Smith2011}, and there are tasks that operate at the semantic level and focus on natural language understanding (e.g.~
information extraction, sentiment analysis, or question answering) \citep{MacCartney2014}. Furthermore, there are natural language generation tasks (e.g.~machine translation, text summarization) \citep{Gatt2018}, and there are multimodal tasks in which the inputs to be processed can be of different modalities (e.g.~text plus image, audio, or video). Additionally, these tasks can be approached in different formats. Sentiment analysis, for example, can be conducted as a document classification task \citep{Pang:2002gt}, a sequence tagging task \citep{Mitchell2013}, or a span extraction task \citep{Hu2019}. Especially with regard to natural language understanding, however, many NLP tasks can be framed as binary or multi-class classification tasks in which the model's task is to assign one out of two or one out of several class labels to each text input \citep[see e.g.][]{Wang2019a}. This matches well with text-based research in social science where the measurement of an a priori-defined concept via supervised learning is very frequently implemented as a text classification task.\footnote{This is not to say that all supervised learning in social science is classification. Especially in political science, supervised techniques that estimate values for documents on latent continuous dimensions have been developed \citep{Laver2003, Perry2017}. For a new technique see \citet{Wankmueller2021a}.}
\end{tcolorbox}

actually seeks to conduct) \citep[p.~1347]{Pan2010}. In sequential transfer learning---which is one common type of transfer learning---the aim when training on a source task is to acquire a highly general, close to universal language representation model \citep[p.~64]{Ruder2019}. The pretrained general-purpose representation model then can be used as an input to a target task of interest \citep[p.~63-64]{Ruder2019}. This practice of using a pretrained language model as an initialization for training on a target task has been shown to improve the prediction performances on a large variety of NLP target tasks (\citealp{Ruder2020b}; \citealp[p.~22-23]{Bommasani2021}). Moreover, adapting a pretrained language model to a target task requires fewer target training examples than when not using transfer learning and training the model from scratch on the target task \citep[p.~334]{Howard2018}.

In addition to the efficiency and performance gains from research on transfer learning, the introduction of the attention mechanism \citep{Bahdanau2015} and the self-attention mechanism \citep{Vaswani2017} has significantly improved the ability of deep learning NLP models to capture contextual information from texts. (Self-)attention mechanisms learn a token representation by capturing information from other tokens, and thereby encode textual dependencies and context-dependent meanings. (Self-)attention mechanisms constitute the core building blocks of the Transformer---a type of deep learning model that has been presented by \citeauthor{Vaswani2017} in 2017. During the last years, several Transformer-based models that are used in a transfer learning setting have been introduced \citep[e.g.][]{Devlin2019, Liu2019, Yang2019}. These models substantively outperform previous state-of-the-art models across a large variety of NLP tasks (\citealp{Ruder2020b}; \citealp[p.~22-23]{Bommasani2021}).

Due to the likely increases in prediction accuracy, as well as the efficient and less resourceful adaptation phase, transfer learning with deep (e.g.~Transformer-based) language representation models seems promising to social science researchers. It seems especially promising to researchers that seek to have as accurate as possible text-based measures but lack the resources to annotate large amounts of data or are interested in specific domains in which only small corpora and few training instances exist. 
In order to equip social scientists to use the potential of transfer learning with Transformer-based models for their research, this paper provides an 
introduction to transfer learning and the Transformer.

The following Section \ref{sec:tradvsdl} compares conventional machine learning to deep learning by focusing on the question of how textual features (e.g.~characters, terms, symbols) and larger textual units (e.g.~sentences, paragraphs, tweets, comments, speeches, ... here named: documents) tend to be represented in conventional vs.~deep learning approaches. The subsequent Section \ref{sec:tl} on transfer learning provides an answer to the question of what transfer learning is and explains in more detail in what ways transfer learning might be beneficial. The then following Section \ref{sec:attention} introduces the attention mechanism and the Transformer and elaborates on how the Transformer has advanced the study of text. Afterward, an overview of Transformer-based models for transfer learning is provided (Section \ref{sec:tltbm}). Here, a special focus will be given to the seminal Transformer-based language representation model BERT (standing for Bidirectional Encoder Representations from Transformers) \citep{Devlin2019}. Additionally, the changes in NLP and artificial intelligence (AI) research, that these models have caused, are outlined and problematic aspects are discussed. Finally, three Transformer-based models for transfer learning, BERT \citep{Devlin2019}, RoBERTa \citep{Liu2019}, and the Longformer \citep{Beltagy2020}, are compared to traditional learning algorithms based on three classification tasks using data from speeches in the UK parliament \citep{Duthie2018}, tweets regarding the legalization of abortion \citep{Mohammad2017}, and comments from Wikipedia Talk pages \citep{Jigsaw2018} (Section \ref{sec:applications}). The final Section \ref{sec:conclusion2} concludes with a discussion on task-specific factors and research goals for which neural transfer learning with Transformers is highly beneficial vs.~rather limited. Throughout the paper, it is assumed that readers 
know core elements of neural network architectures, and are also familiar with recurrent neural networks (RNNs) as well as with optimization via stochastic gradient descent with backpropagation. For readers that feel not sufficiently acquainted with these deep learning concepts see Appendix \ref{appendix:introdl}. Also note that a document is an ordered sequence of tokens and here is denoted as $d_i = (a_1, \dots, a_t, \dots, a_T)$. A token $a_t$ is an instance of a type, which is the set of all tokens that are made up of the same string of characters \citep[p.~22]{Manning:2008vf}. A type that is used for analysis is named term or feature and here is given as $z_u$. The set of features that are used in an analysis is $\{z_1, \dots, z_u \dots, z_U\}$.


\section{Conventional Machine Learning vs.~Deep Learning} \label{sec:tradvsdl}

\subsection{Conventional Machine Learning}

Given raw input data $D = (d_1, \dots, d_i, \dots, d_N)$ (e.g.~a corpus comprising $N$ raw text files) 
and a corresponding output variable $\bm y = [y_1, \dots, y_i, \dots, y_N]^{\top}$ (e.g.~class labels), 
the aim in supervised machine learning is to 
find the parameters $\bm \theta$ of a function $f$ that captures the general systematic relation between $D$ and $\bm y$ such that 
the trained model will generalize well and generate accurate predictions for new, yet unseen data $D_{test}$ (\citealp[p.~30]{James2013}; \citealp[ch.~1.1.3]{Chollet2020}).

When applying a machine learning algorithm in order to learn a function that as accurately as possible maps from text data inputs to provided outputs, the algorithm, however, will not take as an input raw text documents. The raw text units first have to be converted into a format that is suitable for data analysis \citep[p.~463-464]{Benoit2019b}. This is achieved by transforming each raw data unit $d_i$ into an abstracted representation of $d_i$ \citep[p.~463-464]{Benoit2019b}.
Learning in supervised machine learning hence 
essentially is a two-step process \citep[p.~10]{Goodfellow2016}: The first step is to create or learn representations of the data, and the second step is to learn mappings from these representations of the data to the output. For a single document $d_i$, the first step 
can be described as $\mathsf{f}_l(d_i, \bm{\hat \theta}_l)$ and the entire process as
\begin{equation}
\hat{y}_i = \mathsf{f}(d_i, \bm{\hat \theta}) = \mathsf{f}_o(\mathsf{f}_l(d_i, \bm{\hat \theta}_l),  \bm{\hat \theta}_o)
\end{equation}
where the subscript $l$ indicates the mapping from raw data to a representation and the subscript $o$ indicates the mapping from the representation to the output. Conventional machine learning algorithms cover the second step: They learn a function mapping data representations to the output. This in turn implies that the first step falls to the researcher who has to (manually) generate representations of the data herself. 

\begin{figure}
\begin{center}
\large
\includegraphics[width=1\textwidth]{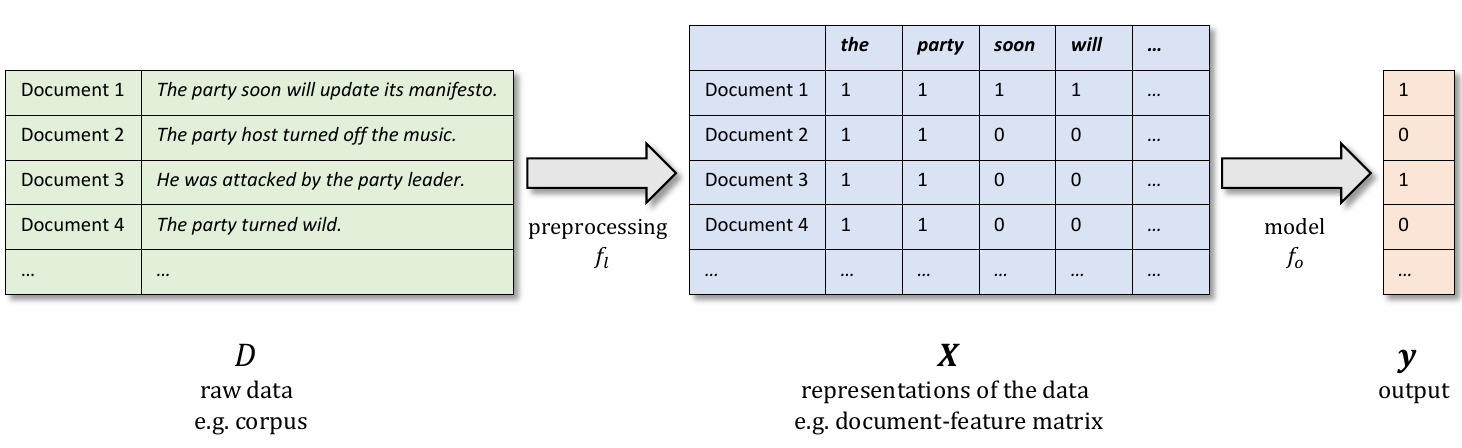}\\
\caption[Learning as a Two-Step Process]{\textbf{Learning as a Two-Step Process.} \small{In text-based applications of conventional machine learning approaches, the raw data $D$ first are (manually) preprocessed such that each example is represented as a feature vector in the document-feature matrix $\bm X$. Second, these representations of the data are fed as inputs to a traditional machine learning algorithm that learns a mapping between data representations $\bm X$ and outputs $\bm y$.}}
\label{fig:ml1}
\end{center}
\end{figure}

When working with texts, the raw data $D$ are typically a corpus of text documents. A very common approach in social science is to transform the raw text files via multiple preprocessing procedures into a document-feature matrix $\bm X = [\bm x_1| \dots| \bm x_i| \dots| \bm x_N]^\top$ (see Figure \ref{fig:ml1}) \citep[p.~464]{Benoit2019b}. In a document-feature matrix, each document is represented as a feature vector $\bm x_i = [x_{i1}, \dots, x_{iu}, \dots, x_{iU}]$ \citep[p.~147]{Turney2010}. Element $x_{iu}$ in this vector gives the value of the $i$th document on the $u$th textual feature---and typically is the (weighted) number of times that the $u$th feature occurs in the $i$th document \citep[p.~143, 147]{Turney2010}. To conduct the second learning step, the researcher then commonly applies a conventional machine learning algorithm on the document-feature matrix to learn the relation between the document-feature representation of the data $\bm X$ and the provided response values $\bm y$.

There are three difficulties with this approach. 
The first is that it may be hard for the researcher to a priori know which features are useful for the task at hand \citep[p.~3-5]{Goodfellow2016}. The performance of a supervised learning algorithm will depend on the representation of the data in the document-feature matrix \citep[p.~3-4]{Goodfellow2016}. In a classification task, features that capture observed linguistic variation that helps in assigning the texts into the correct categories are more informative and will lead to a better classification performance than features that capture variation that is not helpful in distinguishing between the classes \citep[p.~3-5]{Goodfellow2016}. Yet determining which sets of possibly highly abstract and complex features are informative (and which are not) is highly difficult \citep[p.~3-5]{Goodfellow2016}. A researcher can choose from a multitude of possible preprocessing steps such as stemming, lowercasing, removing stopwords, adding POS tags, or applying a sentiment lexicon.\footnote{For a more detailed list of possible steps see \citet[p.~153 ff.]{Turney2010} and \citet[p.~170-172]{Denny2018}. Note that not only the set of selected preprocessing steps but also the order in which they are implemented define the way in which the texts at hand are represented and thus affect the research findings \citep{Denny2018}.} Social scientists may be able to use some of their domain knowledge in deciding upon a few specific preprocessing decisions (e.g.~whether it is likely that excluding a predefined list of stopwords will be beneficial because it reduces dimensionality or will harm performance because the stopword list includes terms that are important). Domain knowledge, however, is most unlikely to guide researchers regarding all possible permutations of preprocessing steps. Simply trying out each possible preprocessing permutation in order to select the best performing one for a supervised task is not possible given the massive number of permutations and limited researcher resources.

Second, the document-feature matrix defines a representational space in which each feature constitutes one separate and independent dimension of the space \citep[p.~349-350]{Goldberg2016}. Accordingly, if there are $U$ features, $\{z_1, \dots, z_u, \dots, z_U\}$, then each feature $z_u$ defines one dimension of the representational space. This implies that each feature is represented to be as distant (and thus as dissimilar) to one feature as to each other feature \citep[p.~351]{Goldberg2016}. The terms \emph{`excellent'} and \emph{`outstanding'} are treated as (dis)similar to each other as the terms \emph{`excellent'} and \emph{`terrible'}. Moreover, as---even after feature exclusion and feature normalization
---the number of features in any text-based analysis typically tends to be high, the document representation vectors $\bm{x}_i$ tend to be high-dimensional and sparse. (This is, $\bm{x}_i$ is likely to be a vector with a large number of elements, most of which will be zero.) By defining such a high-dimensional and sparse feature space, a document-feature matrix brings about the curse of dimensionality: There are much more combinations of feature values than can be covered by the training data, therefore making it difficult to generalize to regions of the space for which no or only few training data are observed (\citealp[p.~1137-1138, 1139-1140]{Bengio2003}).

The third problem is that in a document-feature matrix each document is represented as a bag-of-words \citep[p.~147]{Turney2010}. Bag-of-words-based representations disregard word order and syntactic or semantic dependencies between words in a sequence \citep[p.~147]{Turney2010}.\footnote{By counting the occurrence of word sequences of length $N$, $N$-gram models extend unigram-based bag-of-words models and allow for capturing information from small contexts around words. However, by including $N$-grams as features, the dimensionality of the feature space increases, thereby increasing the problem of high dimensionality and sparsity. Moreover, texts often exhibit dependencies between words that are positioned much farther apart than what could be captured with $N$-gram models \citep[p.~395]{Chang2019}.} Yet text is contextual and sequential by nature. Word order carries meaning. And the context, in which a word is embedded in, is essential in determining the meaning of a word. When represented as a bag-of-words, the sentence \emph{`The opposition party leader attacked the prime minister.'} cannot be distinguished from the sentence \emph{`The prime minister attacked the opposition party leader.'}. Moreover, the fact that the word \emph{`party'} here refers to a political party rather than a festive social gathering only becomes clear from the context.

\subsection{Deep Learning and Embeddings} \label{sec:dle5678}

These stated problems are overcome by deep neural networks and the real-valued vector representations that typically accompany deep neural networks \citep{Goldberg2016, Goodfellow2016}. In contrast to conventional machine learning algorithms, deep learning models can be considered to conduct both learning steps: They learn representations of the data \emph{and} a function mapping data representations to the output. In deep learning models, an abstract representation of the data is learned by applying the data to a stack of several simple (typically nonlinear) functions \citep[p.~5, 164-165]{Goodfellow2016}. Each function takes as an input the representation of the data created by (the sequence of) previous functions and generates a new representation:
\begin{equation}  \label{eq:dl1}
\mathsf{f}(d_i,  \bm{\hat \theta}) = \mathsf{f}_o(\dots \mathsf{f}_{l_3}(\mathsf{f}_{l_2}(\mathsf{f}_{l_1}(d_i,  \bm{\hat \theta}_{l_1}),  \bm{\hat \theta}_{l_2}),  \bm{\hat \theta}_{l_3}) \dots,  \bm{\hat \theta}_o)
\end{equation}
When applying deep neural networks to text-based applications, they, however, do not take as an input the raw text documents. They still have to be fed with a data format they can read. Neural networks usually operate on real-valued vector representations of entities, named embeddings \citep[p.~349-351]{Goldberg2016}. Frequently, the embedded entities are unique vocabulary terms \citep[p.~5]{Pilehvar2021}. (In this case, embeddings are referred to as word embeddings.) Yet embeddings also can be learned for smaller or larger textual units such as characters \citep{Akbik2018}, subwords \citep{Bojanowski2017}), sentences or documents \citep{Le2014, Reimers2019}, and even for entities of a different nature, e.g.~word senses \citep{Rothe2015} or the nodes in a network \citep{Kipf2017}.

When working with text data and having a set of $U$ textual features (e.g.~$U$ vocabulary terms in a corpus), which are given by $\{z_1, \dots, z_u, \dots, z_U\}$, then each feature $z_u$ can be represented as an embedding---a $K$-dimensional real-valued vector $\bm{z}_u \in \mathbb{R}^K$. Whereas in a document-feature matrix representation $z_u$ is a dimension of a $U$-dimensional feature space, now $z_u$ is represented as a dense vector $\bm{z}_u$ that is embedded in a $K$-dimensional continuous space (where typically $K << U$) \citep[p.~350-351]{Goldberg2016}. The positioning of the embedding vectors within this $K$-dimensional space reflects the information that the embeddings encode about the features. For example, if the embeddings encode the feature's semantics, then features that are semantically similar are likely to have close embedding vectors and thus are likely to be positioned close in space \citep[p.~4-5, 39]{Pilehvar2021}. 
(The terms \emph{`excellent'} and \emph{`outstanding'} then are likely to be close together and far from \emph{`terrible'}.) Learning real-valued vector representations for textual features and documents implies that one obtains relatively low-dimensional and dense (rather than high-dimensional and sparse) representations \citep[p.~349-351]{Goldberg2016}. This, in turn, much facilitates generalization via the employment of local smoothness assumptions \citep[p.~1137-1140]{Bengio2003}. 

In text-based applications, the feature representation vectors can be collectively kept in an embedding matrix $\bm E$, which is a $U \times K$ matrix that stores for each of the $U$ unique features its $K$-dimensional embedding $\bm{z}_u$ \citep[p.~360]{Goldberg2016}. Therefore, if a researcher wants to feed a text document, $d_i = (a_1, \dots, a_t, \dots, a_T)$, to a neural network, then for each token $a_t$, the respective feature embedding $\bm{z}_{[a_t]}$ is retrieved from the embedding matrix $\bm E$ \citep[p.~360]{Goldberg2016}. In the end, the document $(a_1, \dots, a_t, \dots, a_T)$ is mapped to a sequence of embeddings $(\bm{z}_{[a_1]}, \dots, \bm{z}_{[a_t]}, \dots, \bm{z}_{[a_T]})$ which is the input representation entering the network \citep[p.~33]{Ruder2019}.
A researcher that has a corpus of raw text documents at his disposal thus merely has to extract features $\{z_1, \dots, z_u, \dots, z_U\}$ for which vector representations will be learned \citep[p.~349-353]{Goldberg2016}. 
In practice, this typically involves tokenization 
and sometimes normalization (e.g.~lowercasing). Other than that, no text preprocessing steps are required. The values of the elements of the embedding vector $\bm{z}_u$ of each feature are treated as usual parameters and are learned jointly with the other model parameters in the optimization process \citep[p.~349, 361]{Goldberg2016}. 
The representation $\bm{z}_u$ 
thus does not have to be manually prefabricated by the researcher. 

Nevertheless, it is common practice to initialize the representation vectors $\bm{z}_u$ with pretrained embeddings \citep[p.~365]{Goldberg2016}. Continuous bag-of-words (CBOW) \citep{Mikolov2013b}, Skip-gram \citep{Mikolov2013b, Mikolov2013c}, and Global Vectors (GloVe) \citep{Pennington2014}, are early seminal models that learn (pretrained) word embeddings. In these models, the embedding for a target term $z_u$ is learned on the basis of words that occur in a context window surrounding instances of term $z_u$ \citep[p.~1533-1535]{Pennington2014}. In CBOW, for example, the self-supervised learning task is to predict a word given its context words \citep[p.~4-5]{Mikolov2013b}. In Skip-gram, surrounding context words are predicted given a target word \citep[p.~4-5]{Mikolov2013b}. And GloVe seeks to find a representation for term $z_u$ and context term $z_j$ such that the dot product of their representation vectors, $\bm{z}_{u}^\top \tilde{\bm{z}}_{j}$, has a minimal squared difference to the logged number of times that $z_j$ occurs in a context window around $z_u$ \citep[p.~1535]{Pennington2014}. By utilizing the contexts of a term to learn a representation for this term, these models implement the distributional hypothesis \citep{Firth1957} according to which the meaning of a term can be inferred from its context (\citealp[p.~365]{Goldberg2016}; \citealp[p.~4]{Spirling2020}). Similar terms are expected to be observed in similar contexts and, consequently, semantically or syntactically similar terms are expected to be positioned close in the embedding space (\citealp[p.~365]{Goldberg2016}; \citealp[p.~27]{Pilehvar2021}). 

Representations learned by these early word embedding models such as CBOW and GloVe, however, have two shortcomings.
First, these models learn for each feature $z_u$ a single vector representation $\bm{z}_u \in \mathbb{R}^K$ that encodes one information \citep[p.~74]{Ruder2019}. For models to deduce complex meanings from sequences of tokens, however, several different information types that build on top of each other are likely to be required (e.g.~morphological, syntactic, and semantic information) \citep{Peters2018a, Tenney2019}. In NLP, therefore, deep neural networks are now being used to learn deep (i.e.~multi-layered) representations (\citealp[p.~2233-2234]{Peters2018}; \citealp[p.~74]{Ruder2019}). In deep neural networks, each layer learns one vector representation for a feature \citep[p.~2228]{Peters2018}. Hence, a single feature is represented by several vectors---one vector from each layer. Although it cannot be specified a priori which information is encoded in which hidden layer in a specific model trained on a specific task, research suggests that information encoded in lower layers is less complex and more general whereas information encoded in higher layers is more complex and more task-specific \citep{Yosinski2014, Tenney2019}. The representations learned by a deep neural language model thus may, for example, encode morphological information about core textual elements at lower layers, syntactic aspects at middle layers, and semantic information in higher layers \citep{Peters2018a, Jawahar2019, Tenney2019}. 
Consequently, while previously often only the first embedding layer $\bm E$ of a deep neural network had been initialized with pretrained word embeddings (e.g.~from Skip-gram or GloVe), the standard procedure in NLP now is to pretrain an entire deep neural network 
(\citealp[p.~74-75]{Pilehvar2021}; \citealp[p.~74]{Ruder2019}). Then, the pretrained model (including its pretrained parameters) is used as the starting point for training on the target task of interest \citep[p.~64, 77]{Ruder2019}. In general, this procedure 
is called sequential transfer learning \citep[p.~45]{Ruder2019} and will be introduced in more detail in Section \ref{sec:tl} below.

The second issue with the early word embedding models is that by representing each feature $z_u$ with a single vector $\bm{z}_u$, distinct meanings of one feature are fused into one representation vector \citep[p.~60]{Pilehvar2021}. This is known as the meaning conflation deficiency \citep[p.~60]{Pilehvar2021}. For example, the term \emph{`class'} can denote a group of people with a similar status but also a course taken at an educational institution \citep{Wordnet2010}. A single vector is likely to blend these two meanings (having the effect that the vector will be located somewhere between the two different meanings in space) \citep[p.~102]{Schuetze1998}. In recent years, this issue has been addressed in NLP by learning contextualized representations \citep[p.~74]{Pilehvar2021}. Contextualized representations account for the observation that the (exact) meaning of a token arises from its context \citep[p.~82]{Pilehvar2021}. A contextualized representation is a representation of a token $a_t$ (not a feature $z_u$) and 
is a function of the tokens that precede and/or proceed token $a_t$ \citep[p.~82]{Pilehvar2021}. Hence, two identical tokens that occur in different contexts, will have a different representation. As contextualized representations capture information from surrounding tokens, they also allow encoding information on syntactic or semantic dependencies between tokens \citep[p.~74]{Pilehvar2021}. Deep and contextualized representations are learned by deep RNNs \citep{Elman1990} 
(and derived architectures such as deep long short-term memory (LSTM) models \citep{Hochreiter1997}) and the Transformer \citep{Vaswani2017}.  
Currently, especially Transformer-based models are widely used to learn deep contextualized representations.

To wrap up and to sum up: Because they are composed of a stack of nonlinear functions that map from one vector representation to the next,
deep learning models tend to have a high capacity \citep[p.~5, 168]{Goodfellow2016}. This is, they can approximate a large variety of complex functions \citep[p.~110]{Goodfellow2016}. On less complex data structures, large deep learning models may risk overfitting and conventional machine learning approaches with lower expressivity may be more suitable. The ability to express complicated functions, the ability to automatically learn multi-layered representations, and the ability to encode information on dependencies between tokens and to encode context-dependent meanings of tokens, however, seem important when working with text data: In most areas of NLP, bag-of-words-based representations coupled with conventional machine learning does not constitute the state-of-the-art for some time now \citep{Goldberg2016}. Moreover, models that learn deep and contextualized representations tend to generalize better across a wide spectrum of specific target tasks compared to the one-layer representations from early word embedding architectures \citep[see e.g.][]{McCann2018}. Consequently, over the last two decades, the field of NLP moved from sparse, high-dimensional representations of single textual features and documents to dense, relatively low-dimensional, deep, and contextualized representations. Today, models that can learn deep contextualized representations and that can be transferred (and then put to use) across learning tasks and domains are at the heart of many modern NLP approaches \citep{Bommasani2021}. How and why models are transferred across tasks and domains is described in the next section.


\section{Transfer Learning} \label{sec:tl}

The classic approach in supervised learning is to have a training data set containing a large number of annotated instances, $(\bm{x}_i, y_i)_{i=1}^N$, that are provided to a model that learns a function relating the $\bm x_i$ to the $y_i$ \citep[p.~2]{Ruder2019}. If the train and test data instances have been drawn from the same distribution over the feature space, the trained model can be expected to make accurate predictions for the test data, i.e.~to generalize well \citep[p.~42]{Ruder2019}. Given another task (i.e.~another set of labels to learn and thus another function $f$ to approximate) or another domain (e.g.~another set of documents with a different thematic focus and thus another distribution over the feature space), the standard supervised learning procedure would be to sample and create a new training data set for this new task and domain \citep[p.~42]{Ruder2019}. Yet the (manual) labeling of thousands to millions of training instances for each new task makes supervised learning highly resource intensive and prohibitively costly to be applied for all potentially useful and interesting tasks \citep[p.~2-3]{Ruder2019}. In situations, in which the number of annotated training examples is restricted or the researcher lacks the resources to label a sufficiently large number of training instances, classic supervised learning fails \citep[p.~2-3]{Ruder2019}. This is where transfer learning comes in. Transfer learning refers to a set of learning procedures in which knowledge, that has been obtained by training on a source task in a source domain, is transferred to the learning process of the target task in a task domain, where either the target task is not the same task as the source task or the target domain is not the same as the source domain (\citealp[p.~1347]{Pan2010}; \citealp[p.~42-43]{Ruder2019}).

\subsection{A Taxonomy of Transfer Learning} \label{sec:tltaxonomy}

\citet[p.~44-46]{Ruder2019} provides a taxonomy of transfer learning scenarios in NLP: In transductive transfer learning, source and target domains differ, and annotated training examples are typically only available for the source domain \citep[p.~46]{Ruder2019}. Here, knowledge is transferred across domains (domain adaptation); or---if source and target documents are from different domains in the sense that they are from different languages---knowledge is transferred across languages (cross-lingual learning) \citep[p.~46]{Ruder2019}. In inductive transfer learning, source and target tasks differ, but the researcher has at least some labeled training samples of the target task \citep[p.~46]{Ruder2019}. In this setting, tasks can be learned simultaneously (multitask learning) or sequentially (sequential transfer learning) \citep[p.~46]{Ruder2019}.

\subsection{Sequential Transfer Learning} \label{sec:sequentialtl}

In this article, the focus is on sequential transfer learning, which is a frequently employed type of transfer learning. In sequential transfer learning, two stages are distinguished: First, a model is pretrained on a source task (pretraining phase) \citep[p.~64]{Ruder2019}. Subsequently, the knowledge gained in the pretraining phase is transferred to the learning process on the target task (adaptation phase) \citep[p.~64]{Ruder2019}. In NLP, the \emph{knowledge} that is transferred are typically the parameter values learned during training the source model \citep[p.~43]{Ruder2019}. The model parameters define how token representations are computed from inputs and define how token representations are transformed into updated versions of token representations in deeper layers.

The common procedure in sequential transfer learning in NLP is to select a source task that is likely to learn a model that constitutes a widely applicable language representation tool and thus is likely to provide an effective input for a large spectrum of specific target tasks \citep[p.~64]{Ruder2019}. Because many training instances are required to learn such a general model, training a source model in the sequential transfer learning setting is highly expensive \citep[p.~64]{Ruder2019}. 
Yet adapting a once pretrained model to a target task is often fast and cheap as transfer learning procedures require only a small proportion of the annotated target data required by standard supervised learning procedures in order to achieve the same level of performance \citep[p.~334]{Howard2018}. In \citet[p.~334]{Howard2018}, for example, training the deep learning model ULMFiT 
from scratch on the target task requires 5 to 20 times more labeled training examples to reach the same error rate than when adapting a pretrained ULMFiT model to the target task.

When a model whose parameter values have been learned by training on a suitable task and data set is used as a pretrained input to the training process on a target task, this is likely to increase the prediction performance on the target task---even if only few target training instances are used (\citealp[p.~334-335]{Howard2018}; \citealp[p.~65]{Ruder2019}). The smaller the target task training data set size, the more salient the pretrained model parameters become. When decreasing the number of target task training set instances, the prediction performance of deep neural networks that are trained from scratch on the target task declines \citep[p.~334]{Howard2018}. For models that are used in a transfer learning setting and are pretrained on a source task before being trained on the target task, prediction performance levels also decline; yet performance levels decrease more slowly and more slightly \citep[p.~334]{Howard2018}. Hence, for medium-sized or small training data sets, the prediction performance increase achieved by transfer learning is likely to be larger than for very large training data sets \citep[p.~334-335]{Howard2018}.

\subsection{Pretraining} \label{sec:pretraining}

In order to learn a general, all-purpose language representation model, that is relevant for a wide spectrum of tasks within an entire discipline, two things are required: (1) a pretraining data set that contains a large number of training samples and is representative of the feature distribution studied across the discipline and (2) a suitable pretraining task (\citealp{Ruder2018}; \citealp[p.~65]{Ruder2019}). The most fundamental pretraining approaches in NLP are self-supervised \citep[p.~68]{Ruder2019}. Among these, a very common pretraining task is language modeling \citep{Bengio2003}. A language model models the probability for a sequence of tokens \citep[p.~1138]{Bengio2003}. As the probability for a sequence of $T$ tokens, $P(a_1, \dots, a_t, \dots, a_T)$, can be computed as 
\begin{equation} \label{eq:lm}
P(a_1, \dots, a_t, \dots, a_T) = \prod_{t=1}^T P(a_t|a_1,\dots,a_{t-1})
\end{equation}
or as
\begin{equation} \label{eq:lm2}
P(a_1, \dots, a_t, \dots, a_T) = \prod_{t=T}^1 P(a_t|a_T,\dots,a_{t+1})
\end{equation}
language modeling involves predicting the conditional probability of token $a_t$ given all its preceding tokens, $P(a_t|a_1,\dots,a_{t-1})$, or implicates predicting the conditional probability of token $a_t$ given all its succeeding tokens, $P(a_t|a_T,\dots,a_{t+1})$ (\citealp[p.~1138]{Bengio2003}; \citealp[p.~2229]{Peters2018}). A forward language model models the probability in Equation \ref{eq:lm}, a backward language model computes the probability in Equation \ref{eq:lm2} \citep[p.~2228-2229]{Peters2018}. When being trained on a forward and/or backward language modeling task in pretraining, a model learns general structures and aspects of language, such as long-range dependencies, compositional structures, semantics, and sentiment, that are relevant for a wide spectrum of possible target tasks \citep{Howard2018, Peters2018a, Ruder2018}. Hence, language modeling can be considered a well-suited pretraining task \citep[p.~329-330]{Howard2018}.\footnote{The text corpora that are employed for pretraining vary widely regarding the number of tokens they contain as well as their accessibility \citep[p.~3-4]{Assenmacher2020}. (A detailed and systematic overview of these data sets is provided by \citet{Assenmacher2020}.) Most models are trained on a combination of different corpora. Several models \citep[e.g.][]{Devlin2019, Yang2019, Lan2020, Liu2019} use the English Wikipedia and the BooksCorpus Dataset \citep{Zhu2015}. Many models \citep[e.g.][]{Liu2019, Radford2019a, Yang2019, Brown2020} additionally also use pretraining corpora made up of web documents obtained from crawling the web.}

\subsection{Adaptation: Feature Extraction vs.~Fine-Tuning}  \label{sec:adaption}

There are two basic ways how to implement the adaptation phase in transfer learning: feature extraction vs.~fine-tuning \citep[p.~77]{Ruder2019}. In a feature extraction approach, the parameters learned in the pretraining phase are frozen and not altered during adaptation \citep[p.~77]{Ruder2019}. In fine-tuning, on the other hand, the pretrained parameters are updated in the adaptation phase \citep[p.~77]{Ruder2019}.

An example of a feature extraction approach is ELMo 
\citep{Peters2018}. After pretraining, ELMo is applied without further adaptations on each target task sequence to produce for each token in each sequence three layers of representation vectors \citep[p.~2229-2230]{Peters2018}. For each token, the representation vectors then are extracted to serve as the input for a new target task-specific model that learns a linear combination of the three layers of representation vectors \citep[p.~2229-2230]{Peters2018}. Here, only the weights of the linear model but not the parameters extracted from the pretrained model are trained \citep[p.~2229-2230]{Peters2018}. 

In fine-tuning---which now is the standard adaptation procedure in sequential transfer learning \citep{Ruder2021}---typically the same model architecture used in pretraining is also used for adaptation \citep[p.~8]{Peters2019}. Merely a task-specific output layer is added to the model \citep[p.~8]{Peters2019}. The parameters learned in the pretraining phase serve as initializations for the model in the adaptation phase \citep[p.~77]{Ruder2019}. When training the model on the target task, the gradients are allowed to backpropagate to the pretrained parameters and thus induce changes on these pretrained parameters \citep[p.~77]{Ruder2019}. In contrast to the feature extraction approach, the pretrained parameters hence are allowed to be fine-tuned to capture task-specific adjustments \citep[p.~77]{Ruder2019}.\footnote{A central parameter in fine-tuning is the learning rate $\eta$ with which the gradients are updated during training on the target task (see Equation \ref{eq:learn} in Appendix \ref{appendix:introdl}). Too much fine-tuning (i.e.~a too high learning rate) can lead to catastrophic forgetting---a situation in which the parameters learned during pretraining are overwritten and therefore forgotten when fine-tuning the model (\citealp{Kirkpatrick2017}; \citealp[p.~330-332]{Howard2018}). 
A too careful fine-tuning scheme (i.e.~a too low learning rate), in contrast, may lead to a very slow convergence process \citep[p.~330-332]{Howard2018}. In general, it is recommended that the learning rate should be lower than the learning rate used in pretraining such that the parameters learned during pretraining are not altered too much \citep[p.~78]{Ruder2019}.} 
When fine-tuning BERT on a target task, for example, a target task-specific output layer is put on top of the pretraining architecture \citep[p.~4173]{Devlin2019}. Then the entire architecture is trained, meaning that \emph{all} parameters 
are updated \citep[p.~4173]{Devlin2019}.

\subsection{Cross-Lingual Learning}

\defcitealias{dbmdz2021}{dbmdz 2021}
One reason for having only a limited amount of target task training data (or limited resources for labeling target task training data) could be that the target task texts are in a language other than English.\footnote{I am grateful to one of the reviewers for pointing this out to me.} In this case, transfer learning offers two solutions. One solution is to implement sequential transfer learning with a model that has been pretrained on a monolingual corpus in the target language.\footnote{Examples of non-English pretrained language representation models are, for example, the French CamemBERT \citep{Martin2020}, 
the Vietnamese PhoBERT \citep{Nguyen2020}, or German \citepalias{dbmdz2021} 
and Chinese BERT models \citep{Devlin2019a}. An overview of language-specific pretrained models is provided by the website https://bertlang.unibocconi.it/ which is introduced in \citet{Nozza2020}.} If, however, no monolingual pretrained model exists for the target language and/or no labeled target task training data in the language of interest are available, then another type of transfer learning---namely cross-lingual learning---provides a possible solution. In cross-lingual learning, source and target domains differ in the sense that source and target documents come from different languages \citep[p.~45]{Ruder2019}. Moreover, labeled training data are usually only available for the source language but not the target language \citep{Ruder2019e}. One way to conduct cross-lingual learning is as follows \citep{Ruder2019e}: (1) Cross-lingual representations are learned. This can be achieved by pretraining a 
model on text data from multiple languages (see 
e.g.~
\citealp{Devlin2019a}, 
\citealp{Conneau2020}, and 
\citealp{Xue2021}). (2) The labeled training examples in the source language are used to learn task-specific parameters that map from the cross-lingual representations to the task-specific outputs. (3) The pretrained model (containing cross-lingual representations plus task-specific parameters) is directly applied---without any adaptation step---on data in the target language to make predictions for target language data. So far, research suggests that the prediction performance of pretrained monolingual 
models on downstream target tasks tends to exceed the performance of multilingual models \citep{Rust2021}. But if the multilingual pretraining corpus contains substantial amounts of text in the target task language and if target language-adapted tokenizers are used, the performance differences between monolingual and multilingual models can become small \citep{Rust2021}. For more information on cross-lingual learning see \citet{Ruder2019e}.


\subsection{Zero-Shot Learning} \label{sec:zsl}

A further strand of research within NLP aims at the development and pretraining of models that are able to make accurate predictions for a wide range of different target tasks without having been explicitly fine-tuned on those target tasks \citep{Radford2019a, Yin2019, Brown2020}. The aim is to have a model that performs well on a task it has not conducted before \citep{Davison2020}. This general idea is often referred to as zero-shot learning (but the precise definition of the term varies across research papers) \citep{Davison2020}. Here, following the \emph{Definition-Wild} of \citet[p.~3915]{Yin2019} zero-shot learning is considered a setting in which a model makes predictions for target task texts without having seen task-specific pairs $(\bm{x}_i, y_i)$ and without having seen the space of task-specific labels (e.g.~$\mathcal Y = \{positive, negative\}$) during training. One work in this context that has generated attention far beyond the boundaries of the field of NLP is the GPT-3 model \citep{Brown2020}. (For a note on GPT-3 see Appendix \ref{appendix:gpt}.) Zero-shot learning partly can achieve surprisingly high prediction performances on target tasks. Thus far, however, performance levels tend to be lower compared to state-of-the-art fine-tuned models \citep[see e.g.~the zero-shot GPT-3 in][]{Brown2020}.


\section{(Self-)Attention and the Transformer}  \label{sec:attention}

In NLP, the attention mechanism first has been introduced for Neural Machine Translation (NMT) by \citet{Bahdanau2015}. The attention mechanism allows to model dependencies between tokens irrespective of the distance between them \citep[p.~5999]{Vaswani2017}. The Transformer is a deep learning architecture that is based on attention mechanisms \citep[p.~5999]{Vaswani2017}. This section first explains the attention mechanism and then introduces the Transformer.

\subsection{The Attention Mechanism}  \label{sec:attentionmech}

\begin{figure}[htb]
\begin{center}
\footnotesize
\includegraphics[width=0.8\textwidth]{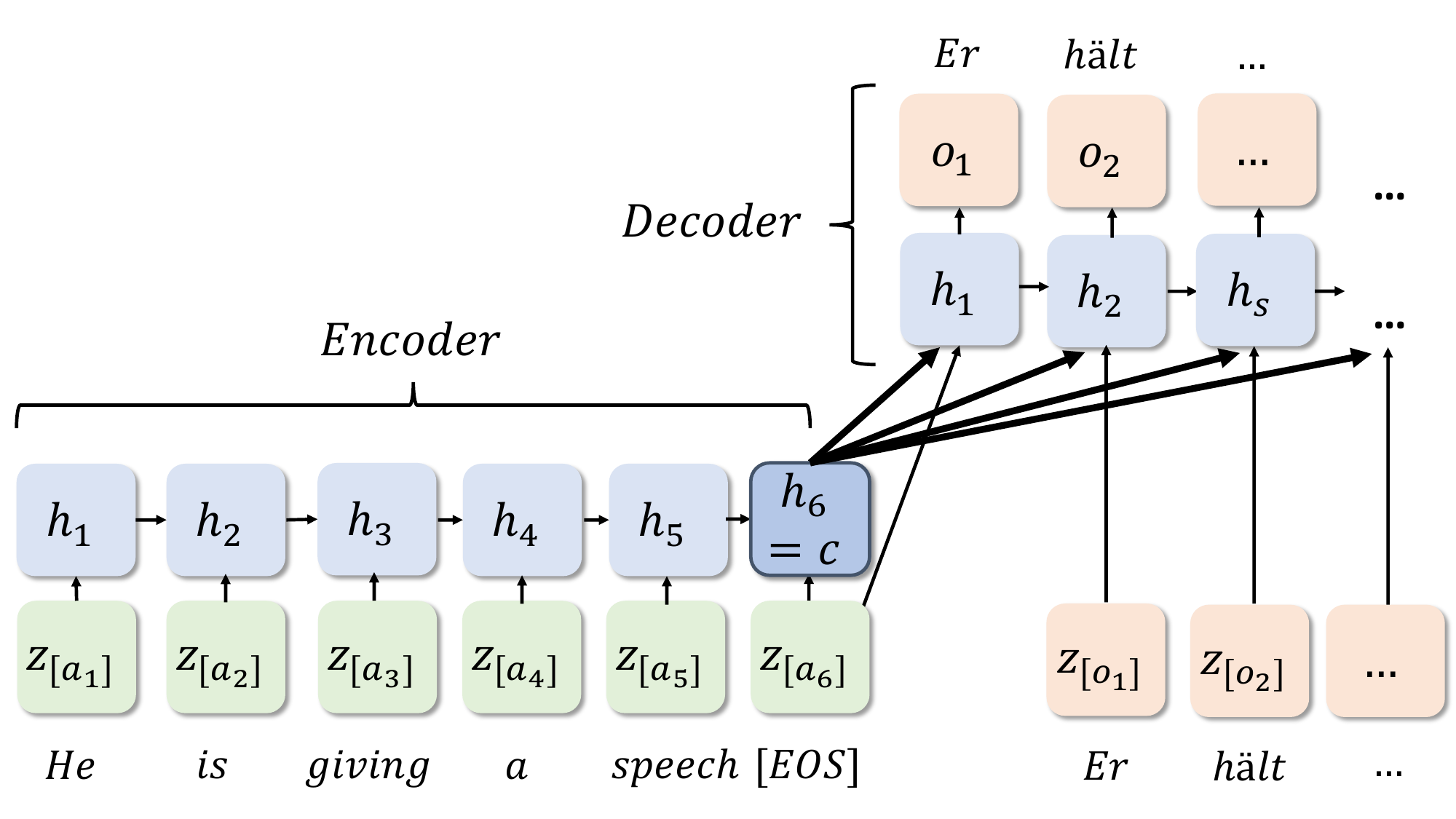}
\caption[Encoder-Decoder Architecture]{\textbf{Encoder-Decoder Architecture.} \small{Encoder-decoder structure in neural machine translation. In this example, the six token input sentence $(He, is, giving, a, speech, [EOS])$ is translated to German: $($\emph{Er, h{\"alt}, eine, Rede, [EOS]}$)$. The end-of-sentence symbol $[EOS]$ is used to signal to the model the end of a sentence. The recurrent encoder processes one input embedding $\bm{z}_{[a_t]}$ at a time and updates the input hidden state $\bm{h}_{t}$ at each time step. The last encoder hidden state $\bm{h}_{6}$ serves as context vector $\bm c$ that captures all the information from the input sequence. The decoder generates one translated output token at a time. Each output hidden state  $\bm{h}_{s}$ is a function of the preceding hidden state $\bm{h}_{s-1}$, the preceding predicted output token  embedding $\bm{z}_{[o_{s-1}]}$, and context vector $\bm c$. }}
\label{fig:encdec}
\end{center}
\end{figure}

The common task encountered in NMT is to translate a sequence of $T$ tokens in language $A$, $(a_1, \dots, a_t, \dots, a_T)$, to a sequence of $S$ tokens in language $O$, $(o_1, \dots, o_{s}, \dots, o_{S})$ \citep[p.~3106]{Sutskever2014}. The classic architecture to solve this task is an encoder-decoder structure (see Figure \ref{fig:encdec}) \citep[p.~3106]{Sutskever2014}.  In general, an encoder transforms input data into a representation and a decoder conducts the reverse operation: The decoder produces data output from an encoded representation. In the early NMT articles, the encoder maps the input tokens $(a_1, \dots, a_t, \dots, a_T)$ into a single context vector $\bm c$ of fixed dimensionality that is then provided to the decoder that generates the sequence of translated output tokens $(o_1, \dots, o_{s}, \dots, o_{S})$ from $\bm c$ \citep[p.~3106]{Sutskever2014}. 

Another characteristic of early NMT articles is that encoder and decoder are recurrent models \citep[p.~3106]{Sutskever2014} (on recurrent models see Appendix \ref{appendix:rnns}). Hence, the encoder processes each input embedding $\bm{z}_{[a_t]}$ step by step. The hidden state at time step $t$, $\bm{h}_{t}$, is a nonlinear function (here denoted by $\sigma$) of the previous hidden state, $\bm{h}_{t-1}$, and input embedding $\bm{z}_{[a_t]}$ \citep[p.~1725]{Cho2014}:  
\begin{equation}
\bm{h}_{t} = \sigma(\bm{h}_{t-1}, \bm{z}_{[a_t]})
\end{equation}
The last encoder hidden state, $\bm{h}_{T}$, corresponds to context vector $\bm c$ that then is passed on to the decoder which---given the information encoded in $\bm c$---produces a variable-length sequence output $(o_1, \dots, o_{s}, \dots, o_{S})$ \citep[p.~1725]{Cho2014}. The decoder also operates in a recurrent manner: Based on the current decoder hidden state $\bm{h}_{s}$, one output token $o_{s}$ is predicted at one time step \citep[p.~1725]{Cho2014}.\footnote{More precisely, in \citet[p.~1725]{Cho2014}, the decoder's prediction for the next output token is a function of the current decoder hidden state $\bm{h}_{s}$, the embedding of the previous output token $\bm{z}_{[o_{s-1}]}$, and context vector $\bm c$. The decoder produces a probability distribution over the vocabulary signifying the next predicted output token: $P(o_s | o_1, \dots, o_{s-1}, \bm{c}) = \sigma_o(\bm{h}_{s}, \bm{z}_{[o_{s-1}]}, \bm c)$ \citep[p.~1725]{Cho2014}.} In contrast to the encoder, the hidden state of the decoder at time step $s$, $\bm{h}_{s}$, is not only a function of the previous hidden state $\bm{h}_{s-1}$ but also the embedding of the previous output token $\bm{z}_{[o_{s-1}]}$, and context vector $\bm c$ (see also Figure \ref{fig:encdec}) \citep[p.~1725]{Cho2014}:
\begin{equation}
\bm{h}_{s} = \sigma(\bm{h}_{s-1}, \bm{z}_{[o_{s-1}]}, \bm{c})
\end{equation}
A problem with this traditional encoder-decoder structure is that all the information about the input sequence---regardless of the length of the input sequence---is captured in a single context vector $\bm c$ \citep[p.~1]{Bahdanau2015}. 

\begin{figure}[htb]
\begin{center}
\footnotesize
\includegraphics[width=0.8\textwidth]{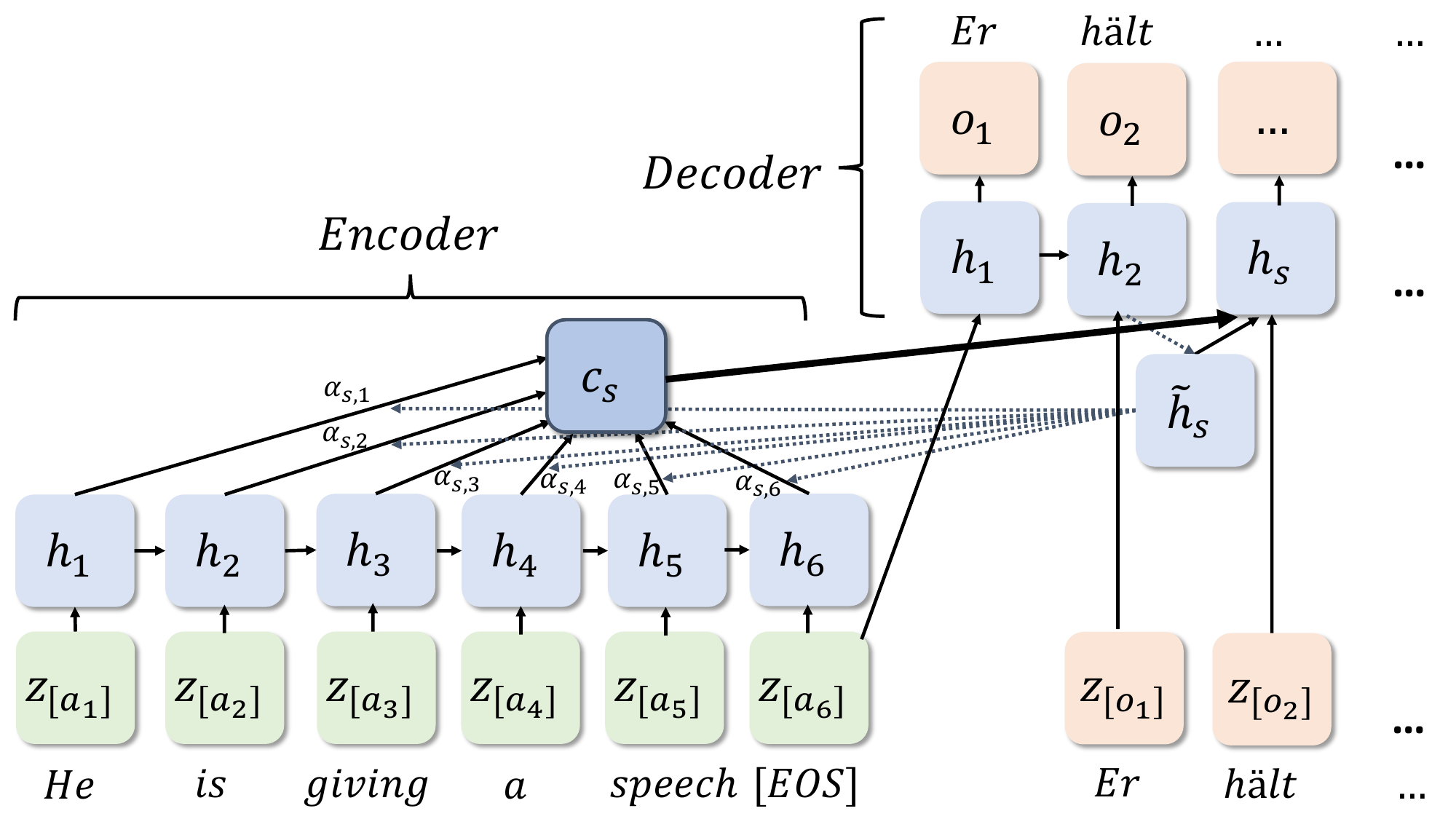}
\caption[Attention in an Encoder-Decoder Architecture]{\textbf{Attention in an Encoder-Decoder Architecture.} \small{Visualization of the attention mechanism in an encoder-decoder structure at time step $s$. In the attention mechanism, at each time step, i.e.~for each output token, there is a token-specific context vector $\bm{c}_s$. $\bm{c}_s$ is computed as the weighted sum over all input hidden states $(\bm{h}_{1}, \dots, \bm{h}_{6})$. The weights are $(\alpha_{s,1}, \dots, \alpha_{s,6})$. $\alpha_{s,1}$ results from a scoring function that captures the similarity between the $s$th output token, as represented by the initial output hidden state $\bm{\tilde{h}}_{s}$, and input token hidden state $\bm{h}_{1}$.}}
\label{fig:attention_encdec}
\end{center}
\end{figure}

The attention mechanism 
resolves this problem. In the attention mechanism, 
at each time step, the decoder can attend to, and thus derive information from, all encoder-produced input hidden states when computing its hidden state $\bm{h}_{s}$ (see Figure \ref{fig:attention_encdec}). More precisely, the decoder hidden state at time point $s$, $\bm{h}_{s}$, is a function of the initial decoder hidden state $\bm{\tilde{h}}_{s}$, the previous output token $\bm{z}_{[o_{s-1}]}$, and an output token-specific context vector $\bm{c}_{s}$ \citep[p.~1414]{Luong2015}.\footnote{Note that Equation \ref{eq:luong} blends the specifications of \citet[p.~1414]{Luong2015} and \citet[p.~3]{Bahdanau2015}. \citet[p.~1414]{Luong2015} do not include $\bm{z}_{[o_{s-1}]}$. \citet{Luong2015} also do not explicitly state how they compute $\bm{\tilde{h}}_{s}$. \citet[p.~3]{Bahdanau2015} use $\bm{h}_{s-1}$ instead of $\bm{\tilde{h}}_{s}$ to represent the state of the decoder at $s$ (or rather at the moment just before producing the $s$th output token).}
\begin{equation} \label{eq:luong}
\bm{h}_{s} = \sigma(\bm{\tilde{h}}_{s}, \bm{z}_{[o_{s-1}]}, \bm{c}_s)
\end{equation}
Note that now at each time step there is a context vector $\bm{c}_s$ that is specific to the $s$th output token \citep[p.~3]{Bahdanau2015}.
The attention mechanism rests in the computation of $\bm{c}_{s}$, which is a weighted sum over the input hidden states $(\bm{h}_1, \dots, \bm{h}_t, \dots, \bm{h}_T)$ \citep[p.~3]{Bahdanau2015}: 
\begin{equation} \label{eq:attentionscore}
\bm{c}_{s} = \sum_{t=1}^T \alpha_{s,t} \bm{h}_t
\end{equation}
The weight $\alpha_{s,t}$ is computed as
\begin{equation} \label{eq:attention}
\alpha_{s,t} = \frac{exp(score(\bm{\tilde{h}}_{s}, \bm{h}_{t}))}{\sum_{t^* = 1}^{T^*} exp(score(\bm{\tilde{h}}_{s}, \bm{h}_{t^*}))}
\end{equation}
where $score$ is a scoring function assessing the compatibility between output token representation $\bm{\tilde{h}}_{s}$ and input token representation $\bm{h}_{t}$ \citep[p.~1414]{Luong2015}. $score$ could be, for example, the dot product of $\bm{\tilde{h}}_{s}$ and $\bm{h}_{t}$ \citep[p.~1414]{Luong2015}. The attention weight $\alpha_{s,t}$ is a measure of the degree of alignment of the $t$th input token, represented by $\bm{h}_{t}$, with the $s$th output token, represented as $\bm{\tilde{h}}_{s}$ \citep[p.~3-4]{Bahdanau2015}. Input hidden states that do not match with output token representation $\bm{\tilde{h}}_{s}$ receive a small weight such that their contribution vanishes, whereas input hidden states that are relevant to output token $\bm{\tilde{h}}_{s}$ receive high weights, thereby increasing their contribution \citep{Alammar2018}. Hence, $\bm{c}_{s}$ considers all input hidden states and especially attends to those input hidden states that match with the current output token. As context vector $\bm{c}_{s}$ is constructed 
for each output token based on a weighted sum of \emph{all} input hidden states, the attention architecture allows for modeling dependencies between tokens irrespective of their distance \citep[p.~5999]{Vaswani2017}.

\subsection{The Transformer}  \label{sec:transformer}

\begin{figure}[htb]
\begin{center}
\footnotesize
\includegraphics[width=1\textwidth]{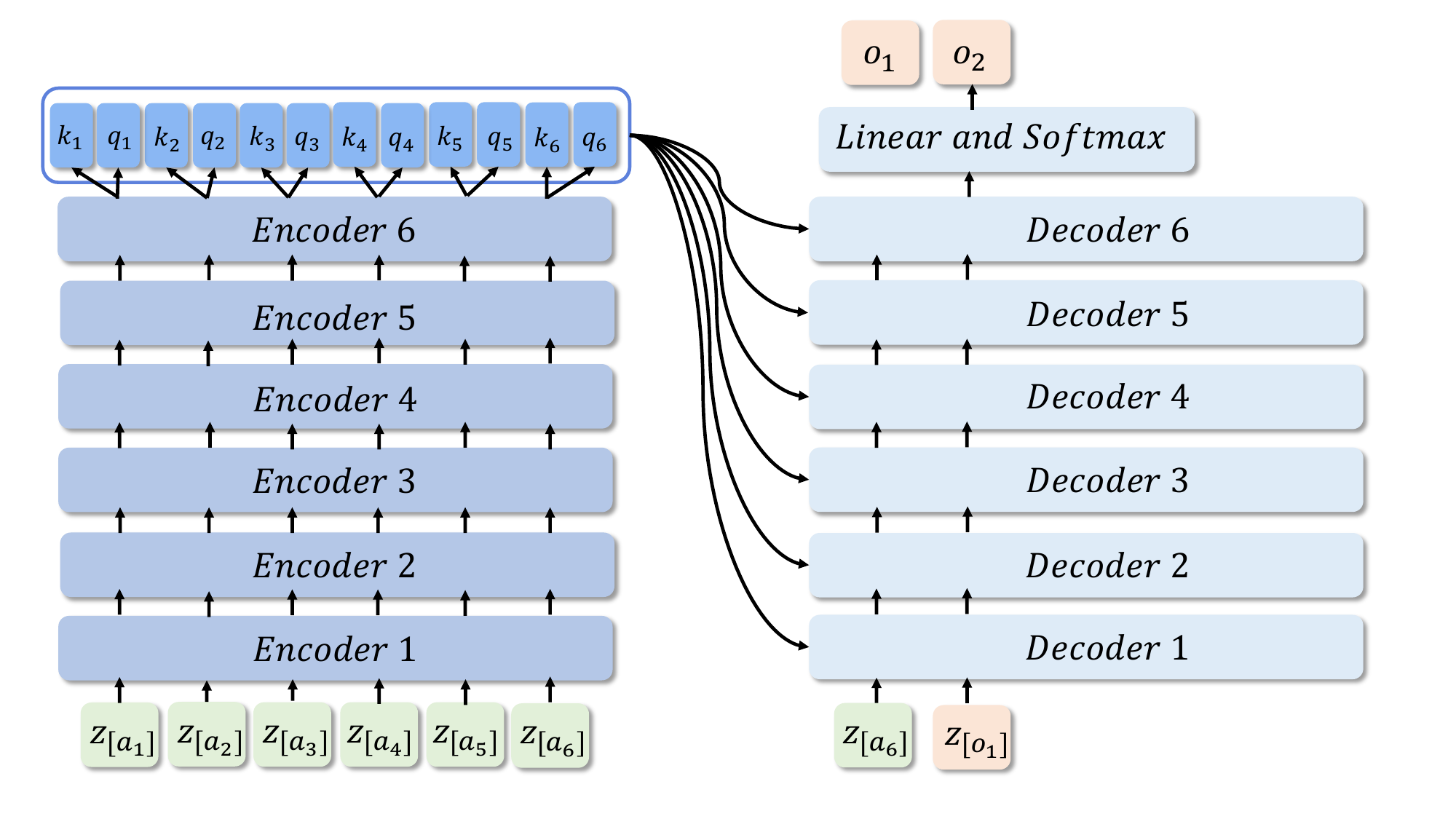}
\caption[Transformer Architecture]{\textbf{Transformer Architecture.} \footnotesize{In the original article by \citet{Vaswani2017}, the Transformer is made up of a stack of six encoders proceeded by a stack of six decoders. In contrast to recurrent architectures where each input token is handled one after another, a Transformer encoder processes the entire set of input token representations in parallel \citep[p.~5999]{Vaswani2017}. Here, the input embeddings are $(\bm{z}_{[a_1]}, \dots, \bm{z}_{[a_6]})$. The sixth encoder passes the key and query vectors of the input tokens, $(\bm{k}_1, \bm{q}_1, \dots, \bm{k}_6, \bm{q}_6)$, to each of the decoders. These key and query vectors from the last encoder are processed in each decoder's encoder-decoder attention layer \citep[p.~6002]{Vaswani2017}. The Transformer decoders operate in an autoregressive manner, meaning that the stack of decoders processes as an additional input the sequence of previous output tokens \citep[p.~6002]{Vaswani2017}. In the visualization here, output tokens are denoted with $(o_1, o_2, \dots)$ and the decoder predicts output token $o_2$ given the previous tokens $(a_6, o_1)$ (where $a_6$ is an end-of-sentence symbol). To predict the $t$th output token, the hidden state of the last decoder is processed through a linear layer and a softmax layer to produce a probability distribution over the terms in the vocabulary \citep[p.~6002]{Vaswani2017}}.}
\label{fig:transformer2}
\end{center}
\end{figure}

The original articles on attention use recurrent architectures in the encoder and decoder. The sequential nature of recurrent models implies that within each training example sequence each token has to be processed one after another---a computationally not efficient strategy \citep[p.~5999]{Vaswani2017}. To overcome this inefficiency and to enable parallel processing within training sequences, \citet{Vaswani2017} introduced the Transformer architecture that is built from attention mechanisms. The Transformer consists of a sequence of six encoders followed by a stack of six decoders (see Figure \ref{fig:transformer2}) \citep[p.~6000]{Vaswani2017}.\footnote{Note that the number of encoders and decoders, as well as the dimensionality of the input embeddings and the key, query and value vectors (introduced in the following), are Transformer hyperparameters that are simply set by the authors to specific values. Other suitable values could be used instead.} Each encoder consists of two components: a multi-head self-attention layer (to be explained below) and a feedforward neural network \citep[p.~6000]{Vaswani2017}. Each decoder also has a multi-head self-attention layer followed by a multi-head encoder-decoder attention layer and a feedforward neural network \citep[p.~6000]{Vaswani2017}. Instead of processing each token of each training example one after another, the Transformer encoder takes as an input the whole set of $T$ embeddings for one training example and processes this set of embeddings, $(\bm{z}_{[a_1]}, \dots, \bm{z}_{[a_t]}, \dots, \bm{z}_{[a_T]})$, in parallel \citep{Alammar2018a}. 

\begin{figure}[h!]
\begin{center}
\footnotesize
\includegraphics[width=0.5\textwidth]{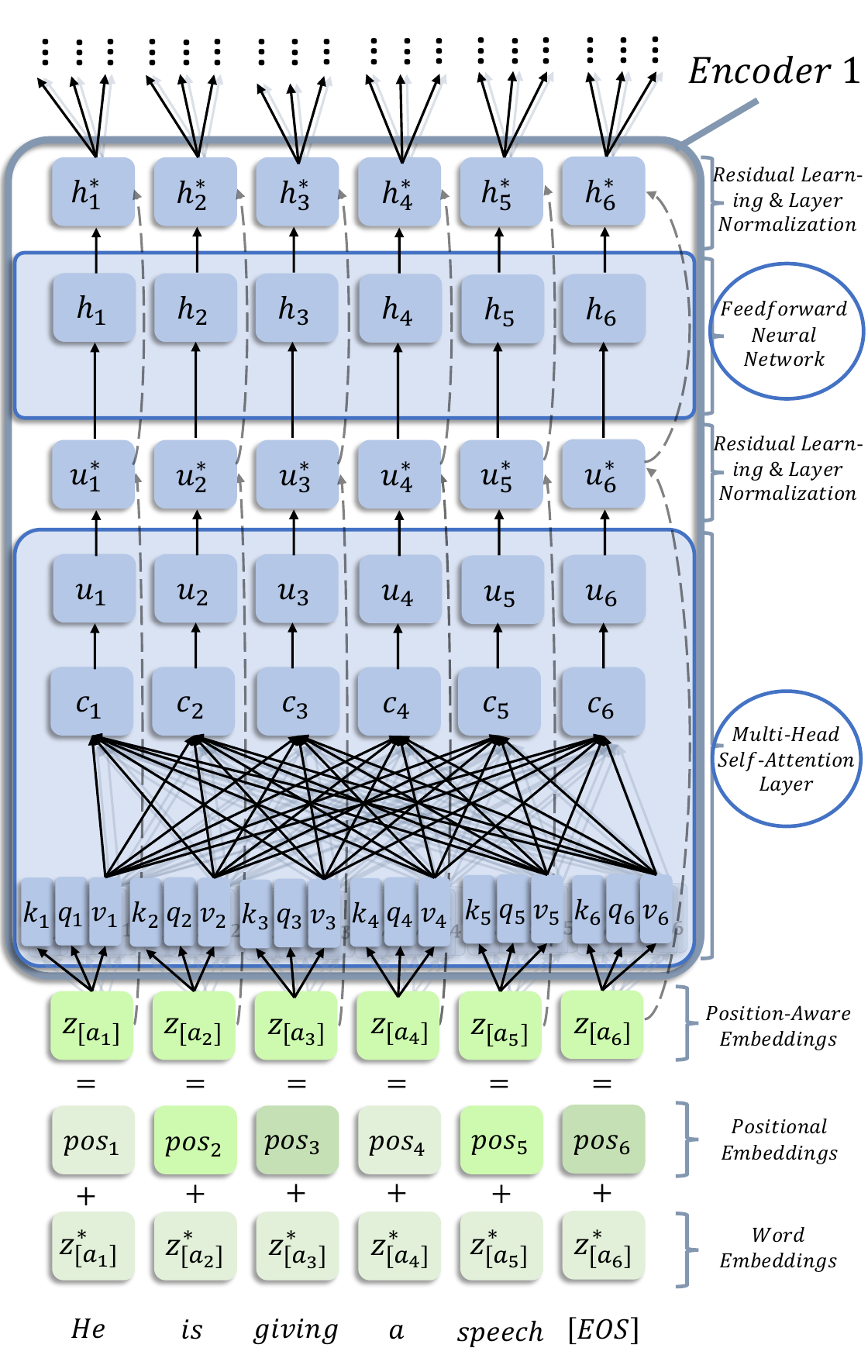}
\caption[Transformer Encoder Architecture]{\textbf{Transformer Encoder Architecture.} \footnotesize{This visualization details the processes in the first Transformer encoder. The encoder comprises a multi-head self-attention layer and a feedforward neural network, each followed by residual learning and layer normalization. The first encoder takes as an input position-aware embeddings, $(\bm{z}_{[a_1]}, \dots, \bm{z}_{[a_6]})$. (A position-aware embedding is the sum of a pure embedding vector and a positional encoding vector \citep[p.~6003]{Vaswani2017}. The positional encoding vector contains information on the position of the $t$th token within the input sequence, thereby making the model aware of token positions \citep[p.~6002-6003]{Vaswani2017}.) The position-aware embeddings then are transformed into eight sets of key, query and value vectors. One set is $(\bm{k}_1, \bm{q}_1,  \bm{v}_1, \dots, \bm{k}_6, \bm{q}_6, \bm{v}_6)$. These are processed in the multi-head self-attention layer to produce eight sets of context vectors (one set being $(\bm{c}_{1}, \dots, \bm{c}_{6})$). The sets then are concatenated and transformed linearly to become the updated representations $(\bm{u}_{1}, \dots, \bm{u}_{6})$. After residual learning and layer normalization, $(\bm{u^*}_{1}, \dots, \bm{u^*}_{6})$ enter the feedforward neural network, whose output---after residual learning and layer normalization---are the updated representations produced by the first Transformer encoder: $(\bm{h^*}_1, \dots, \bm{h^*}_6)$. The representations $(\bm{h^*}_1, \dots, \bm{h^*}_6)$ constitute the input to the next encoder, where they are first transformed to sets of key, query and value vectors.}}
\label{fig:transformer_encoderdetailed}
\end{center}
\end{figure}

The first element in a Transformer encoder is the multi-head self-attention layer. In the self-attention layer, the provided input sequence $(\bm{z}_{[a_1]}, \dots, \bm{z}_{[a_t]}, \dots, \bm{z}_{[a_T]})$ attends to itself. Instead of improving the representation of an output token by attending to tokens in the input sequence, the idea of self-attention is to improve the representation of a token $a_t$ by attending to the tokens in the same sequence in which $a_t$ is embedded in \citep{Alammar2018a}. For example, if \emph{`The company is issuing a statement as it is bankrupt.'} were a sentence to be processed, then the embedding for the token \emph{`it'} that enters the Transformer would not contain any information regarding which other token in the sentence \emph{`it'} is referring to. Is it the company or the statement? In the self-attention mechanism, the representation for \emph{`it'} is updated by attending to---and incorporating information from---other tokens in this sentence \citep{Alammar2018a}. It, therefore, is to be expected that after passing through the self-attention layers, the representation of \emph{`it'} absorbed some of the representation for \emph{`company'} and so encodes information on the dependency between \emph{`it'} and \emph{`company'} \citep{Alammar2018a}. 

\begin{figure}[htb]
\begin{center}
\footnotesize
\includegraphics[width=1\textwidth]{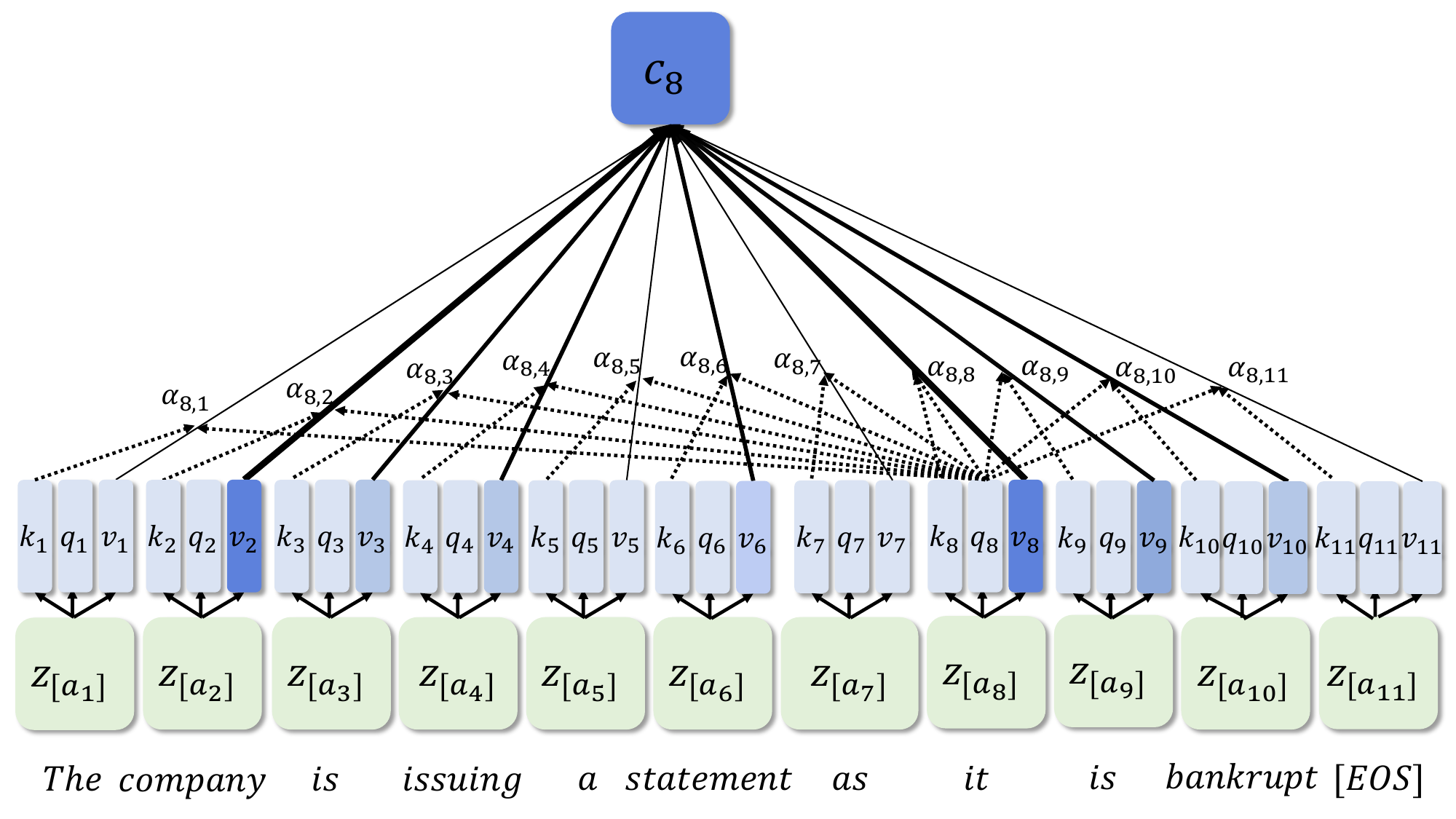}
\caption[Attention Mechanism in the Transformer]{\textbf{Attention Mechanism in the Transformer.} \small{Illustration of the attention mechanism in the first Transformer encoder for the $8$th token (\emph{`it'}) in the example sentence \emph{`The company is issuing a statement as it is bankrupt.'}. The arrows pointing from the value vectors $(\bm{v}_1, \dots, \bm{v}_{11})$ to context vector $\bm{c}_8$ are the weights $(\alpha_{8,1}, \dots, \alpha_{8,t^*}, \dots, \alpha_{8,11})$. A single weight $\alpha_{8,t^*}$ indicates the contribution of token $t^*$ to the representation of token $8$, $\bm{c}_8$. The larger $\alpha_{8,t^*}$ is assumed to be in this example, the thicker the arrow and the darker the corresponding value vector. The dotted lines symbolize the computation of the weights $(\alpha_{8,1}, \dots, \alpha_{8,t^*}, \dots, \alpha_{8,11})$.
}}
\label{fig:attention_large2}
\end{center}
\end{figure}

The first operation within a self-attention layer is that each input embedding $\bm{z}_{[a_t]}$ is transformed into three separate vectors, called key $\bm{k}_t$, query $\bm{q}_t$, and value $\bm{v}_t$  (see Figure \ref{fig:transformer_encoderdetailed}). The key, query, and value vectors are three different projections of the input embedding $\bm{z}_{[a_t]}$ \citep{Alammar2018a}. They are generated by matrix multiplication of $\bm{z}_{[a_t]}$ with three different weight matrices, $\bm{W}_k$, $\bm{W}_q$, and $\bm{W}_v$ \citep[p.~6002]{Vaswani2017}:\footnote{Note that in order to follow the notation in \citet{Vaswani2017}, vectors (which are indicated by bold letters) are treated as row vectors in the following.}
\begin{equation}
\bm{k}_t = \bm{z}_{[a_t]}\bm{W}_k \qquad \bm{q}_t = \bm{z}_{[a_t]}\bm{W}_q \qquad \bm{v}_t = \bm{z}_{[a_t]}\bm{W}_v
\end{equation}
Then, for each token $a_t$, an updated representation (named context vector $\bm{c}_t$) is computed as a weighted sum over the value vectors of \emph{all} tokens that are in the \emph{same} sequence as token $a_t$ (\citeauthor{Vaswani2017}, \citeyear{Vaswani2017}, p.~6000-6002):
\begin{equation}
\bm{c}_{t} = \sum_{t^*=1}^{T^*} \alpha_{t,t^*} \bm{v}_{t^*}
\end{equation}
The attention weight $\alpha_{t,t^*}$ is a function of the similarity between token $a_t$, represented by $\bm{q}_t$, and token $a_{t^*}$, that is represented as $\bm{k}_{t^*}$:
\begin{equation}
\alpha_{t,t^*} = \frac{exp(score(\bm{q}_{t}, \bm{k}_{t^*}))}{\sum_{t^* = 1}^{T^*} exp(score(\bm{q}_{t}, \bm{k}_{t^*}))}
\end{equation}
where $score$ is $(\bm{q}_{t}\bm{k}_{t^*}^\top)/\sqrt{|\bm{k}_{t^*}|}$ \citep[p.~6001]{Vaswani2017}. $\alpha_{t,t^*} $ indicates the contribution of token $a_{t^*}$ for the representation of token $a_t$. Thus, attention vector $\bm{c}_{t}$ is calculated as in a basic attention mechanism (see Equations \ref{eq:attentionscore} and \ref{eq:attention})---except that the attention now is with respect to the value vectors of the tokens that are part of the same sequence as $a_t$ (see also Figure \ref{fig:attention_large2}). 

The self-attention mechanism outlined so far is conducted eight times in parallel \citep[p.~6001-6002]{Vaswani2017}. Hence, for each token $a_t$, eight different sets of query, key and value vectors are generated and there will be not one but eight attention vectors $\{\bm{c}_{t,1}, \dots, \bm{c}_{t,8}\}$ \citep[p.~6001-6002]{Vaswani2017}. In doing so, each attention vector can attend to different tokens in each of the eight different representation spaces \citep[p.~6002]{Vaswani2017}. For example, in one representation space the attention vector for token $a_t$ may learn syntactic structures and in another representation space the attention vector may attend to semantic connections (\citealp[p.~6004]{Vaswani2017}; \citealp{Clark2019a}). Because the self-attention mechanism is implemented eight times in parallel and generates eight attention vectors (or heads), the procedure is called multi-head self-attention \citep[p.~6001]{Vaswani2017}. The eight attention vectors subsequently are concatenated into a single vector, $\bm{c}_t = [\bm{c}_{t,1}; \dots; \bm{c}_{t,8}]$, and multiplied with a corresponding weight matrix $\bm{W}_0$ to produce vector $\bm{u}_t$ \citep[p.~6002]{Vaswani2017}: $\bm{u}_t = \bm{c}_t \bm{W}_0$. Afterward, $\bm{u}_t$ is added to $\bm{z}_{[a_t]}$, thereby allowing for residual learning \citep{He2015}.\footnote{In residual learning, instead of leaning a new representation in each layer, merely the residual change is learned \citep{He2015}. Here $\bm{u}_t$ can be conceived of as the residual on the original representation $\bm{z}_{[a_t]}$. Residual learning has been shown to facilitate the optimization of very deep neural networks \citep{He2015}.}
Then, layer normalization as suggested in \citet{Ba2016} is conducted \citep[p.~6000]{Vaswani2017}.\footnote{In layer normalization, for each training instance, the values of the hidden units within a layer are standardized by using the mean and standard deviation of the layer's hidden units \citep{Ba2016}. Layer normalization reduces training time and enhances generalization performance due to its regularizing effects \citep{Ba2016}.}
\begin{equation}
\bm{u^*}_t = LayerNorm(\bm{u}_t + \bm{z}_{[a_t]})
\end{equation}
$\bm{u^*}_t$ then enters a feedforward neural network with a Rectified Linear Unit (ReLU) activation function \citep[p.~6002]{Vaswani2017}
\begin{equation}
\bm{h}_t = max(0, \bm{u^*}_t \bm{W}_{1} + \bm{b}_{1})\bm{W}_{2} + \bm{b}_{2}
\end{equation}
followed by a residual connection with layer normalization \citep[p.~6000]{Vaswani2017}:
\begin{equation}
\bm{h^*}_t = LayerNorm(\bm{h}_t + \bm{u^*}_t)
\end{equation}
$\bm{h^*}_t$ finally is the representation of token $a_t$ produced by the encoder. It constitutes an updated representation of input embedding $\bm{z}_{[a_t]}$. 
Due to the self-attention mechanism, $\bm{h^*}_t$ is a function of the other tokens in the same sequence and thus captures context-dependent information. Hence, $\bm{h^*}_t$ is a contextualized representation of token $a_t$. The same token in another sequence would obtain another token representation vector.

The entire sequence of representations, $(\bm{h^*}_1, \dots, \bm{h^*}_t, \dots, \bm{h^*}_T)$, that is produced as the encoder output, serves as the input for the next encoder that generates eight sets of query, key, and value vectors from each representation $\bm{h^*}_t$ to implement multi-head self attention and to finally produce an updated set of representations, $(\bm{h^*}_1, \dots, \bm{h^*}_t, \dots, \bm{h^*}_T)^*$, that are passed to the next encoder and so on. The last encoder from the stack of encoders passes the key and value vectors from its produced sequence of updated representations to each encoder-decoder multi-head attention layer in each decoder (see Figure \ref{fig:transformer2}) \citep[p.~6002]{Vaswani2017}. Except for the encoder-decoder attention layer in which the decoder pays attention to the encoder input, the architecture of each decoder is largely the same as those of the encoders \citep[p.~6000]{Vaswani2017}. Note, however, that the stack of decoders operates in an autoregressive manner \citep[p.~5999]{Vaswani2017}. This is, when making the prediction for the next output token $o_s$, the decoders have access to and process the sequence of previous output tokens, $(a_T, o_s, \dots, o_{s-1})$, as additional inputs (see Figure \ref{fig:transformer2}) \citep[p.~5999]{Vaswani2017}. In order to ensure that the decoders are autoregressive, self-attention in each decoder is masked, meaning that the attention vector for output token $o_s$ can only attend to output tokens preceding token $o_s$ \citep[p.~6000]{Vaswani2017}. To predict an output token, the hidden state of the last decoder is handed to a linear and softmax layer to produce a probability distribution over the vocabulary 
\citep[p.~6002]{Vaswani2017}.


\section{Transfer Learning with Transformer-Based Models} \label{sec:tltbm}

Taken together, the Transformer architecture in combination with transfer learning literally transformed the field of NLP \citep[p.~5]{Bommasani2021}. After the introduction of the Transformer by \citet{Vaswani2017}, several models for transfer learning that included elements of the Transformer were developed \citep[e.g.][]{Radford2018, Devlin2019, Yang2019, Clark2020, Raffel2020}. These models and their derivatives significantly outperformed previous state-of-the-art models. 

An important step within these developments was the introduction of BERT \citep{Devlin2019}. By establishing new state-of-the-art performance levels for eleven NLP tasks, BERT demonstrated the power of transfer learning \citep[p.~5]{Bommasani2021}. The introduction of BERT finally paved the way to a new transfer learning-based mode of learning in which it is common to use an already pretrained language model and adapt it to a specific target task as needed (\citealp{Alammar2018b}; \citealp[p.~5]{Bommasani2021}). Simultaneously with and independently of BERT, a wide spectrum of Transformer-based models for transfer learning have been developed. This section first introduces BERT and then provides an overview of further models.

\subsection{BERT}  \label{sec:bert}

BERT consists of a stack of Transformer encoders and comes in two different model sizes \citep[p.~4173]{Devlin2019}: BERT\textsubscript{BASE} consists of 12 stacked Transformer encoders, each with 12 attention heads. The dimensionality of the input embeddings and the updated hidden vector representations is 768. BERT\textsubscript{LARGE} has 24 Transformer encoders with 16 attention heads and a hidden vector size of 1024.\footnote{In the feedforward neural networks, \citet[p.~4183]{Devlin2019} employ the Gaussian Error Linear Unit (GELU) \citep{Hendrycks2016} instead of the ReLU activation function used in the original Transformer. BERT\textsubscript{BASE} has 110 million parameters. BERT\textsubscript{LARGE} has 340 million parameters.} As in the original Transformer, the first BERT encoder takes as an input a sequence of embedded tokens, $(\bm{z}_{[a_1]}, \dots, \bm{z}_{[a_t]}, \dots, \bm{z}_{[a_T]})$, processes the embeddings in parallel through the self-attention layer and the feedforward neural network to generate a set of updated token representations, $(\bm{h^*}_1, \dots, \bm{h^*}_t, \dots, \bm{h^*}_T)$, that are then passed to the next encoder that also generates updated representations to be passed to the next encoder and so on until the representations finally enter output layers for prediction \citep{Alammar2018b}.

The authors inventing BERT sought to tackle a disadvantage of the classic language modeling pretraining task (see Equations \ref{eq:lm} and \ref{eq:lm2}), namely that it is strictly unidirectional \citep[p.~4171]{Devlin2019}. A forward language model can only access information from the preceding tokens $(a_1, \dots, a_{t-1})$ but not from the following tokens $(a_{t+1}, \dots, a_{T})$. The same is true for a backward language model in which information can only be captured from succeeding tokens \citep[p.~5753]{Yang2019}. Assuming that a representation of token $a_t$ from a bidirectional model that simultaneously can attend to preceding and succeeding tokens may constitute a better representation of token $a_t$ than a representation stemming from a unidirectional language model, the authors of BERT invented an adapted variant of the traditional language modeling pretraining task, named masked language modeling, to learn deep contextualized representations that are bidirectional \citep[p.~4171-4172]{Devlin2019}.\footnote{The concatenation of representations learned by a forward language model with the representations of a backward language model does not generate representations that genuinely draw from left and right contexts \citep[p.~4172]{Devlin2019}. The reason is that the forward and backward representations are learned separately and each representation captures information only from a unidirectional context \citep[p.~5753]{Yang2019}.} 

\begin{figure}[htb]
\begin{center}
\footnotesize
\includegraphics[width=1.15\textwidth]{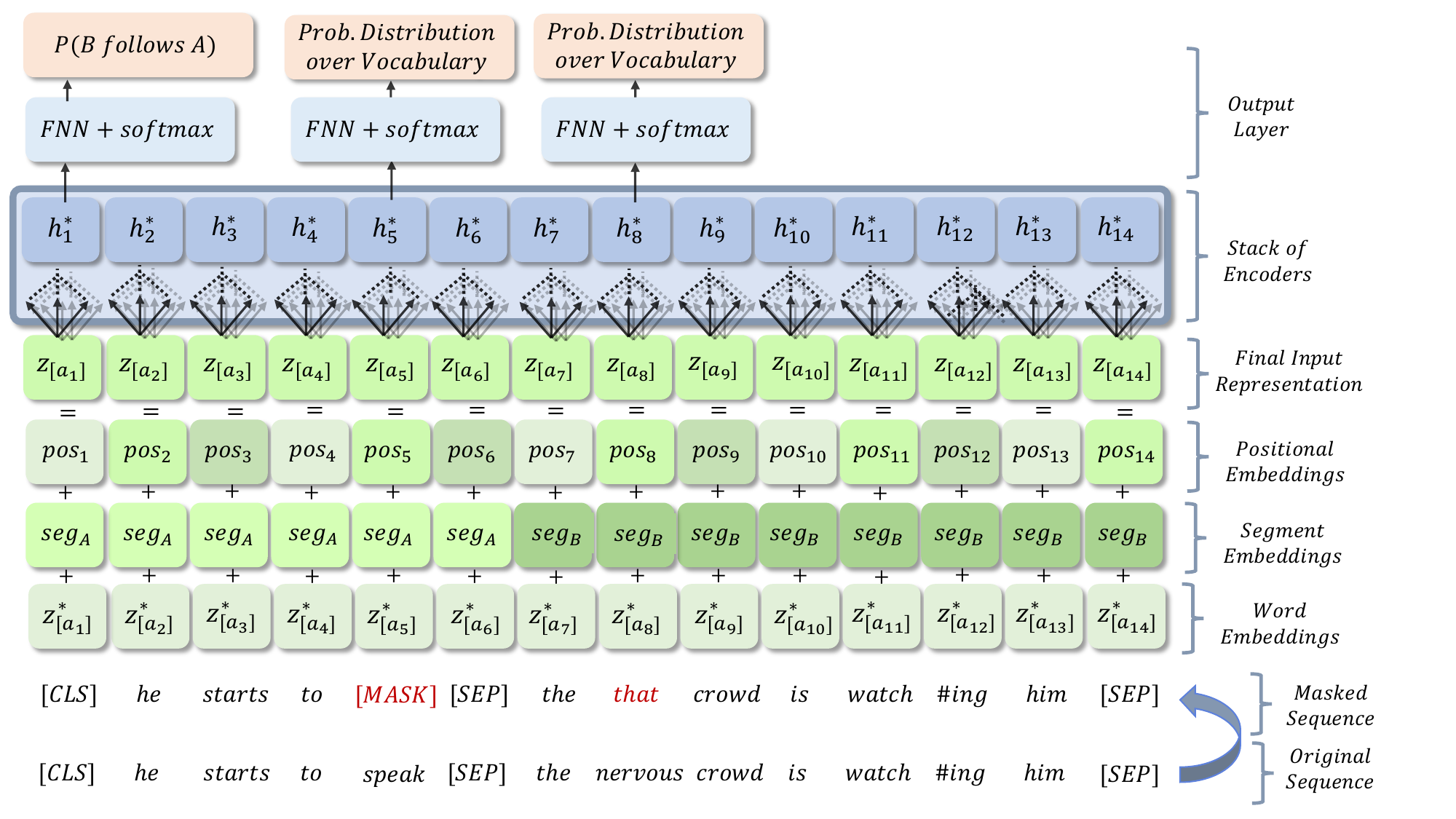}
\caption[Pretraining BERT]{\textbf{Pretraining BERT.} \small{Architecture of BERT in pretraining. Assume that in the lowercased example sequence consisting of the segment pair \emph{`he starts to speak. the nervous crowd is watch-ing him.'} the tokens \emph{`speak'} and \emph{`nervous'} were sampled to be masked. \emph{`speak'} is replaced by the \emph{`[MASK]'} token and \emph{`nervous'} is replaced by the random token \emph{`that'}. The model's task is to predict the tokens \emph{`speak'} and \emph{`nervous'} from the representation vectors it learns at the positions of the input embeddings of \emph{`[MASK]'} and \emph{`that'}. $P(B follows A)$ is the next sentence prediction task. FNN stands for feedforward neural network.}}
\label{fig:bert_pretrain}
\end{center}
\end{figure}

\begin{figure}[htb]
\begin{center}
\footnotesize
\includegraphics[width=1.15\textwidth]{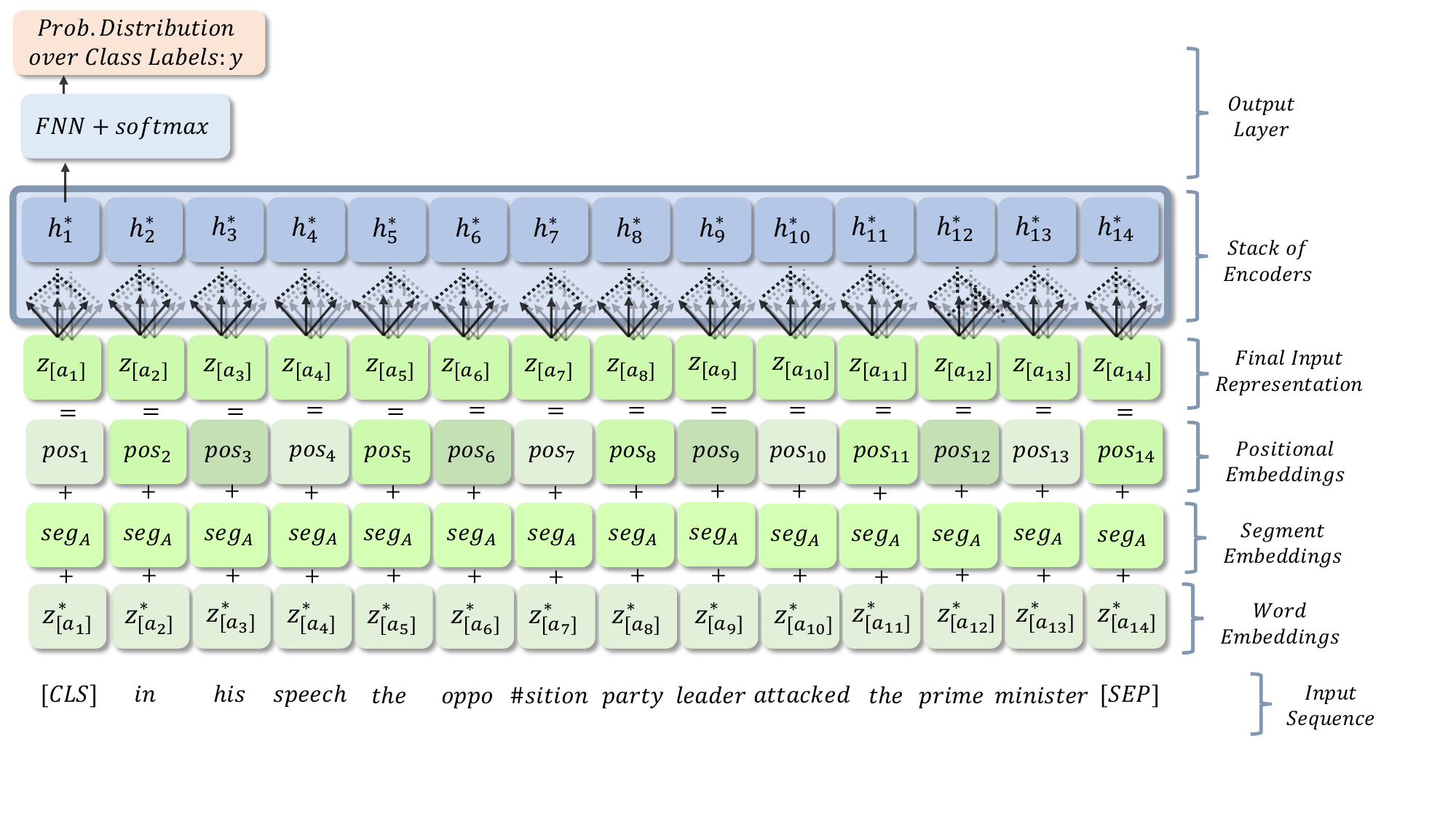}
\caption[Fine-Tuning BERT]{\textbf{Fine-Tuning BERT.} \small{Architecture of BERT during fine-tuning on a single sequence classification task.}}
\label{fig:bert_pretrain2}
\end{center}
\end{figure}

To conduct the masked language modeling task in the pretraining process of BERT, in each input sequence, $15\%$ of the input embeddings are selected at random \citep[p.~4174, 4183]{Devlin2019}. The selected tokens are indexed as $(1, \dots, q, \dots, Q)$ here. $80\%$ of the $Q$ selected tokens will be replaced by the \emph{`[MASK]'} token \citep[p.~4174]{Devlin2019}. $10\%$ of the selected tokens are supplanted with another random token, and $10\%$ of selected tokens remain unchanged \citep[p.~4174]{Devlin2019}. The task then is to correctly predict all $Q$ tokens sampled for the task based on their respective input token representation (for an illustration see Figure \ref{fig:bert_pretrain}) \citep[p.~4173-4174]{Devlin2019}. In doing so, self-attention is possible with regard to all---instead of only preceding or only succeeding---tokens in the same sequence, and thus the learned representations for all tokens in the sequence can capture encoded information from bidirectional contexts \citep[p.~4174, 4182]{Devlin2019}.

In addition to the masked language modeling task, BERT is also pretrained on a next sentence prediction task in which the model has to predict whether the second of two text segments it is presented with succeeds the first \citep[p.~4172, 4174]{Devlin2019}. The second pretraining task is hypothesized to serve the purpose of making BERT also a well-generalizing pretrained model for NLP target tasks that require an understanding of the association between two text segments (e.g.~question answering or natural language inference) \citep[p.~4172, 4174]{Devlin2019}.

To accommodate for the pretraining tasks and to prepare for a wide spectrum of downstream target tasks, the input format accepted by BERT consists of the following elements (see Figure \ref{fig:bert_pretrain}) (\citealp[p.~4174-4175, 4182-4183]{Devlin2019}): 
\begin{itemize}
\item Each sequence of tokens $(a_1, \dots, a_t, \dots, a_T)$ is set to start with the classification token \emph{`[CLS]'}. After fine-tuning, the \emph{`[CLS]'} token functions as an aggregate representation of the entire sequence and is used as an input for single sequence classification target tasks such as sentence sentiment analysis. 
\item The separation token \emph{`[SEP]'} is used to separate different segments.
\item Each token $a_t$ is represented by the sum of its input embedding with a positional embedding and a segment embedding.\footnote{BERT employs the WordPiece tokenizer and uses a vocabulary of 30,000 features \citep{Wu2016}. WordPiece \citep{Schuster2012} is a variant of the Byte-Pair Encoding (BPE) subword tokenization algorithm. (For more information on subword tokenization algorithms see Appendix \ref{appendix:subword}.) The segment embeddings allow the model to distinguish segments. All tokens belonging to the same segment have the same segment embedding.}
\item In software-based implementations, BERT-like models typically require all input sequences to have the same length \citep{Huggingface2020}. To meet this requirement, the text sequences are tailored to the same length by padding or truncation \citep{Huggingface2020}. Truncation is employed if text sequences exceed the maximum accepted sequence length. Truncation implies that excess tokens are removed. In padding, a padding token (\emph{`[PAD]'}) is repeatedly added to a sequence until the desired length is reached \citep{McCormick2019}. Note that due to memory restrictions, the maximum sequence length that BERT can process is limited to 512 tokens.
\end{itemize}

BERT is pretrained with the masked language modeling and the next sentence prediction task. As pretraining corpora the BooksCorpus \citep{Zhu2015} and the English Wikipedia are used \citep[p.~4175]{Devlin2019}. Taken together the pretraining corpus consists of $3.3$ billion tokens \citep[p.~4175]{Devlin2019}. (For details on pretraining BERT see Appendix \ref{appendix:pretrainingbert}.)

Token representations that are produced from a pretrained BERT model afterward can be extracted and taken as an input for a target task-specific architecture as in a classic feature extraction approach \citep[p.~4179]{Devlin2019}. The more common way to use BERT, however, is to fine-tune BERT on the target task. Here, merely the output layer from pretraining is exchanged with an output layer tailored for the target task \citep[p.~4173, 4184]{Devlin2019}. Other than that, the same model architecture is used in pretraining and fine-tuning (compare Figures \ref{fig:bert_pretrain} and \ref{fig:bert_pretrain2}) \citep[p.~4173, 4184]{Devlin2019}. If the target task is to classify single input sequences into a set of predefined categories (see Figure \ref{fig:bert_pretrain2}), the hidden state vector generated by the last Transformer encoder for the \emph{[CLS]} token, $\bm{h}^*_{1}$, enters the following output layer to generate output vector $\bm y$ \citep{BERTCode2018}:
\begin{equation} \label{eq:bertoutput}
\bm y = softmax(\tanh(\bm{h}^*_{1}\bm{W}_1 + \bm{b}_1)\bm{W}_2 + \bm{b}_2)
\end{equation}
$\bm y$'s dimensionality corresponds to $C$---the number of categories in the target classification task. The $c$th element of $\bm y$ gives the predicted probability of the input sequence belonging to the $c$th class. During fine-tuning, not only the weight matrices and bias terms in Equation \ref{eq:bertoutput} but \emph{all} parameters of BERT are updated \citep[p.~6]{Devlin2018}.\footnote{Based on their experiences with adapting BERT on various target tasks, the authors recommend to use for fine-tuning a mini-batch size of 16 or 32 sequences and a global Adam learning rate of 5e-5, 3e-5, or 2e-5 \citep[p.~4183-4184]{Devlin2019}. They also suggest to set the number of epochs to 2, 3 or 4 \citep[p.~4184]{Devlin2019}.}

\subsection{More Transformer-Based Pretrained Models} \label{sec:moremodels}
A helpful way to describe and categorize the various Transformer-based models for transfer learning is to differentiate them according to their pretraining objective and their model architecture \citep{Huggingface2020a}. The major groups of models in this categorization scheme are autoencoding models, autoregressive models, and sequence-to-sequence models \citep{Huggingface2020a}.

\subsubsection{Autoencoding Models}
In their pretraining task, autoencoding models are presented with input sequences that are altered at some positions \citep[p.~5753-5755]{Yang2019}.
The task is to correctly predict the uncorrupted sequence \citep[p.~5753-5755]{Yang2019}.
The models' architecture is typically composed of the encoders of the Transformer which implies that autoencoding models can access the entire set of input sequence tokens and can learn bidirectional token representations \citep{Huggingface2020a}. Autoencoding models tend to be especially high performing in sequence or token classification target tasks \citep{Huggingface2020a}. BERT with its masked language modeling pretraining task is a typical autoencoding model \citep[p.~5753]{Yang2019}. 

Among the various extensions of BERT that have been developed since its introduction, RoBERTa \citep{Liu2019} is widely known. RoBERTa makes changes in the pretraining and hyperparameter settings of BERT. For example, RoBERTa is only pretrained on the masked language modeling and not the next sentence prediction task \citep[p.~4-6]{Liu2019}. Masking is performed dynamically each time before a sequence is presented to the model instead of being conducted once in data preprocessing \citep[p.~4, 6]{Liu2019}. 
Moreover, RoBERTa is pretrained on more data and more heterogeneous data (e.g.~also on web corpora) \citep[p.~5-6]{Liu2019}.\footnote{On ALBERT \citep{Lan2020} and ELECTRA \citep{Clark2020}, two further well-known autoencoding models, see Appendix \ref{appendix:albert}.}

One major disadvantage of pretrained models that are based on the self-attention mechanism in the Transformer is that currently available hardware does not allow Transformer-based models to process long text sequences \citep[p.~1]{Beltagy2020}. The reason is that the memory and time required increase quadratically with sequence length \citep[p.~1]{Beltagy2020}. Long text sequences thereby quickly exceed memory limits of presently existing graphics processing units (GPUs) \citep[p.~1]{Beltagy2020}. Transformer-based pretrained models therefore typically induce a maximum sequence length. For BERT and related models this maximum length usually is 512 tokens. Simple workarounds for processing sequences longer than 512 tokens (e.g.~truncating texts or processing them in chunks) lead to information loss and potential errors \citep[p.~2-3]{Beltagy2020}. To solve this problem, various works present procedures for altering the Transformer architecture such that longer text documents can be processed \citep{Child2019, Dai2019, Beltagy2020, Kitaev2020, Wang2020, Zaheer2020}.

Here, one of these models, the Longformer \citep{Beltagy2020}, is presented in more detail. The Longformer introduces a new variant of the attention mechanism such that time and memory complexity does not scale quadratically but linearly with sequence length and thus longer texts can be processed \citep[p.~3]{Beltagy2020}. The attention mechanism in the Longformer is composed of a sliding window as well as global attention mechanisms for specific preselected tokens \citep[p.~3-4]{Beltagy2020}. In the sliding window, each input token $a_t$---instead of attending to all tokens in the sequence---attends only to a fixed number of tokens to the left and right of $a_t$ \citep[p.~3]{Beltagy2020}. In order to learn representations better adapted to specific NLP tasks, the authors use global attention for specific tokens on specific tasks (e.g. for the \emph{`[CLS]'} token in sequence classification tasks) \citep[p.~3-4]{Beltagy2020}. These preselected tokens directly attend to all tokens in the sequence and enter the computation of the attention vectors of all other tokens \citep[p.~3-4]{Beltagy2020}. The Longformer allows processing text sequences of up to 4,096 tokens \citep[p.~6]{Beltagy2020}. This Longformer-specific attention mechanism can be used as a plug-in replacement of the original attention mechanism in any Transformer-based model \citep[p.~6]{Beltagy2020}. \citet[p.~2]{Beltagy2020} insert the Longformer attention mechanism into the RoBERTa architecture and then continue to pretrain RoBERTa with the Longformer attention mechanism on the masked language modeling task \citep[p.~2]{Beltagy2020}.

\subsubsection{Autoregressive Models}
Autoregressive models are pretrained on the classic language modeling task (see Equations \ref{eq:lm} and \ref{eq:lm2}) \citep[p.~5753-5755]{Yang2019}. 
They are trained to predict the next token given all the preceding tokens in a sequence, $P(a_t|a_1, \dots, a_{t-1})$, and/or to predict the next token given all succeeding tokens, $P(a_t|a_{T}, \dots, a_{t+1})$ \citep[p.~5753]{Yang2019}. 
Hence, autoregressive models are not capable of learning genuine bidirectional representations that draw from left and right contexts \citep[p.~5753]{Yang2019}. In correspondence with this pretraining objective, their architecture is typically based only on the decoders of the Transformer (without encoder-decoder attention) \citep{Huggingface2020a}. 
Due to their decoder-based architecture and the characteristics of their pretraining task, autoregressive models are typically very good at target tasks in which they have to generate text \citep{Huggingface2020a}. Autoregressive models, however, can be successfully fine-tuned to a large variety of downstream tasks \citep{Huggingface2020a}. An elementary autoregressive model is the GPT \citep{Radford2018}. Its successors GPT-2 \citep{Radford2019a} and GPT-3 \citep{Brown2020} are well-known and also play a role in the context of zero-shot learning (see Appendix \ref{appendix:gpt}). Another model using the autoregressive language modeling framework is the XLNet \citep{Yang2019}. (On the XLNet see Appendix \ref{appendix:xlnet}).

\subsubsection{Sequence-to-Sequence Models}
The architecture of sequence-to-sequence models contains Transformer encoders and decoders \citep{Huggingface2020a}. They tend to be pretrained on sequence-to-sequence tasks (e.g.~translation) and, consequently, are especially suited for sequence-to-sequence-like downstream tasks such as translating or summarizing input sequences \citep{Huggingface2020a}. The Transformer itself is a sequence-to-sequence model for translation tasks. BART \citep{Lewis2020} and the T5 \citep{Raffel2020} are further well-known sequence-to-sequence models applicable to a large variety of target tasks \citep{Huggingface2020a}. (On the T5 and BART see Appendix \ref{appendix:seqseq}.)

\subsection{Foundation Models: Concept, Limitations, and Issues}

Whilst this article focuses on transfer learning with Transformer-based models in the context of text classification tasks for the purpose of measurement, the 
application of transfer learning as a mode of learning, as well as the application of the Transformer as a deep neural network architecture, is not restricted to classification tasks, text data, or the purpose of measurement. \citet{Porter2021}, for example, apply the GPT-2 to generate placebos texts in social science survey experiments. Moreover, transfer learning and the Transformer architecture are at the heart of fundamental changes within the entire field of NLP and beyond: AI research previously had moved from classic machine learning (in which a function between representations of data and outputs are learned but representations still have to be engineered) to the era of deep learning (in which deep neural networks learn representations of data but still typically one model is trained for one specific task) \citep[p.~3-4]{Bommasani2021}. Now, AI research seems to move toward the widespread use of pretrained deep neural networks that function as highly general all-purpose models \citep[p.~3-6]{Bommasani2021}. Such deep neural networks 
that are pretrained in a self-supervised fashion on large amounts of data and then can be adapted to a wide spectrum of target tasks are also called foundation models \citep{Bommasani2021}. During the last few years, foundation models not only have taken hold in NLP but across the field of AI research \citep{Bommasani2021}. Thus, transfer learning with (Transformer-based) deep neural networks is not only applied to text data but also, for example, to images \citep{Dosovitskiy2021, Goyal2021}, videos \citep{Sun2019a}, audio data \citep{Baevski2020}, 
and data in tabular form \citep{Yin2020}. Moreover, an increasing number of models are multimodal \citep[e.g.][]{Bapna2021, Fu2021, Radford2021, Ramesh2021}. 
Summarizing current developments, the emerging mode of learning is characterized by 
a deep neural network that
\begin{itemize}
\item is frequently based on the Transformer architecture.\footnote{Due to its self-attention mechanisms, the Transformer is more flexible and general than convolutional or recurrent neural networks \citep[p.~75-76]{Bommasani2021}. The Transformer, however, is not a defining feature of foundation models and at some point may be superseded by new neural network architectures \citep{Jaegle2021a, Jaegle2021b}.}
\item has been pretrained---typically in self-supervised learning mode---on massive amounts of data from various sources and domains. (Pretraining does not have to be constrained to one pretraining task conducted on one type of data in one language. Rather, increasingly general models are developed by pretraining on multiple tasks \citep{Wei2022, Aribandi2022}, pretraining on data in multiple languages \citep{Conneau2020, Babu2021}, or pretraining on data from multiple modes \citep{Luo2020, Bapna2021, Radford2021}.)
\item can process---and learn representations for---these data inputs (probably across domains, languages, and modes) \citep{Tenney2019, Radford2021},
\item after adaptation can be applied to a wide spectrum of tasks, for example, various language understanding tasks 
\citep{Devlin2019, Liu2019}, different language understanding plus language generation tasks 
\citep{Lewis2020, Wei2022}, or tasks related to the understanding and/or generation of data in multiple modes \citep{Luo2020, Fu2021, Ramesh2021}.
\end{itemize}

Foundation models have triggered significant performance enhancements but also come with substantive limitations and problematic issues.\footnote{For an elaborate discussion see \citet{Bommasani2021}.} One major concern, for example,  is that the amounts of resources required in pretraining (especially in terms of data and compute) are so massively large that academic institutions and the scientific research community struggle (or are not able) to pretrain the largest foundation models (but see https://bigscience.huggingface.co/) \citep[p.~11]{Bommasani2021}. Moreover, the data used in pretraining and the model source code are not always publicly available (\citealp[p.~4]{Assenmacher2020}; \citealp{Riedl2020}). This raises deep concerns regarding accessibility and traceability. Another problematic aspect is that these models reflect the representational biases (e.g.~stereotypes, underrepresentations) encoded in the data they have been pretrained on \citep[p.~129-131]{Bommasani2021}. As soon as a model is adapted to some target task, these biases can materialize with serious negative consequences \citep[p.~130]{Bommasani2021}. A further problem 
is the fixed (typically relatively small) maximum sequence length that Transformer-based models can process. 
Whatever the given current computational restrictions, efficient modifications of the self-attention mechanism, as for example presented by the Longformer \citep{Beltagy2020}, allow for longer sequences to be processed than with the original Transformer and thereby constitute important steps toward alleviating this major drawback.\footnote{For an evaluative overview of efficient Transformer-based models see \citet{Tay2021}.}



Besides these issues related to large pretrained representation models, the mere application of deep neural networks is likely to pose further difficulties for social science researchers. One issue is interpretability: If a researcher applies a learning method to measure an a priori-defined concept from text, the ability to as closely as possible imitate human codings on yet unseen test data is arguably the most important goal because this ability indicates the measure's validity. In this very context, a model's prediction performance thus is considered more important than a model's interpretability. Yet interpretability (i.e.~the human-understandable and accurate representation of a model's decision process) can be highly important \citep{Miller2019, Jacovi2020}. Interpretability makes a model's predictions more transparent and more human-retraceable. If a researcher can understand why a model made predictions in a particular way and if a researcher can estimate the effect of input features for a model's predictions, this can enhance the researcher's trust in the model she applies (\citealp[p.~2]{DoshiVelez2017}; \citealp[ch.~3.1]{Molnar2022}). 
Because deep neural networks do not operate on a priori-defined, human-engineered features but have a layered architecture that enables them to learn complex representations of textual inputs, interpretation of deep neural networks is less straightforward and neural network-specific interpretation methods tend to be used (\citealp[p.~49]{Belinkov2019}; \citealp[ch.~10]{Molnar2022}). (For an overview of common methods to make neural networks interpretable see Appendix \ref{appendix:interpret}.) The open-source library Captum (https://captum.ai/) is likely to be a useful interpretability tool. Captum 
implements several attribution algorithms that allow researchers to examine how predicted outputs relate to input features \citep[p.~3]{Kokhlikyan2020}. Captum furthermore provides tools for analyzing attention patterns (see \url{https://captum.ai/tutorials/Bert_SQUAD_Interpret2}).

Another issue is reproducibility: As for conventional models, reproducibility issues with deep neural networks typically arise from random elements that are used during optimization and/or when sampling data (e.g.~in cross-validation, or batch allocation). In both cases, sources of randomness usually can be controlled. Yet in practice, this often proves to be more difficult for deep neural networks than for conventional models. Note, furthermore, that full reproducibility \emph{across} different computing platforms and environments cannot be ensured \citep{NVIDIA2021, PyTorchContributors2021}.

\section{Applications}  \label{sec:applications}

Researchers who wish to apply sequential transfer learning can pretrain a model on a suitable source task by themselves and then finetune the pretrained model to their target task of interest. 
Because pretraining tends to be very expensive, however, the much more convenient, cost-effective, and common approach for applied researchers is to make use of an already pretrained model and then to merely adapt the pretrained model to the target task. Hence, to fully leverage the power of neural transfer learning, researchers require access to already pretrained models that they can fine-tune on their specific tasks. Such access is provided by Hugging Face's Transformers \citep{Wolf2019} which is an open-source library that contains thousands of pretrained NLP models ready to download and use: https://huggingface.co/. The Hugging Face library contains pretrained versions of the models discussed here and a great many models more. Most of the available pretrained models in the Hugging Face library have been pretrained on English texts, yet there are numerous monolingual models pretrained in other languages. Moreover, the library also comprises several models for cross-lingual learning that have been pretrained on text in several languages. The pretrained models can be accessed via the respective Transformers Python package that also provides compatibility with PyTorch \citep{Paszke2019} and TensorFlow \citep{Abadi2015}.\footnote{For basic guidance on deep learning and transfer learning in practice see Appendix \ref{appendix:practice}.} 

In the applications presented in the following neural transfer learning is conducted in Python 3 \citep{vanRossum2009} making use of PyTorch \citep{Paszke2019} and pretrained models from Hugging Face's Transformers \citep{Wolf2019}. The code is executed in Google Colab.\footnote{https://colab.research.google.com/notebooks/intro.ipynb} Whenever a GPU is used, an NVIDIA Tesla T4 is employed. The source code for this study is openly available in figshare at \url{https://doi.org/10.6084/m9.figshare.14394173}. Especially the shared Colab Notebooks serve as templates that other researchers can easily adapt for their NLP tasks.\footnote{More specifically, bag-of-words and word vector-based text preprocessing is implemented in R \citep{RCoreTeam2020} using the packages quanteda \citep{Benoit2018}, stringr \citep{stringr2019}, text2vec \citep{text2vec2020}, and rstudioapi \citep{rstudioapi2020}. Training and evaluating the pretrained Transformer models and the conventional machine learning algorithms is conducted in Python 3 \citep{vanRossum2009} employing the modules and packages gdown \citep{Kentaro2020}, imbalanced-learn \citep{Lemaitre2017}, matplotlib \citep{Hunter2007}, NumPy \citep{Oliphant2006}, pandas \citep{McKinney2010}, seaborn \citep{Waskom2020}, scikit-learn \citep{sklearn2011}, PyTorch \citep{Paszke2019}, watermark \citep{Raschka2020}, Hugging Face's Transformers \citep{Wolf2019}, and the XGBoost Python package \citep{Chen2016}.}

\subsection{Models, Data Sets, and Tasks} 

The aim of this applied section is to explore the use of transfer learning with Transformer-based models for text analyses in social science contexts. To do so, the prediction performances of BERT, RoBERTa, and the Longformer are compared to the performances of two conventional machine learning algorithms: Support vector machines (SVMs) \citep{Boser1992, Cortes1995} and the gradient tree boosting algorithm XGBoost \citep{Chen2016}. SVMs have been widely used in social science text applications \citep[e.g.][]{Diermeier2011, DOrazio2014, Ramey2019, Miller2020, Seboek2020}. As a tree-based (boosting) method XGBoost represents a type of algorithm also commonly utilized \citep[e.g.][]{Katagiri2019, Anastasopoulos2020, Park2020}. The comparisons are conducted on the basis of three different data sets of varying sizes and textual styles:

{\bfseries\sffamily{1. The Ethos Dataset}} \citep{Duthie2018} is a corpus of 3,644 sentences from debates in the UK parliament (train: 2,440; test: 1,204). \citet{Duthie2018} gathered 90 debate transcripts from the period Margaret Thatcher served as Prime Minister (1979-1990). In each debate, they recorded for each spoken sentence whether the sentence refers to the ethos (i.e.~the character) of another politician or party, and if so whether the other's ethos is supported or attacked \citep[p.~4042]{Duthie2018}. The task associated with this data set thus is to as precisely as possible measure the concept of ethos from text. 
With $82.5\%$ of the sentences being non-ethotic, $12.9\%$ attacking and $4.6\%$ supporting another's ethos, the data are quite imbalanced.

{\bfseries\sffamily{2. The Legalization of Abortion Dataset}} comprises 933 tweets (train: 653; test: 280). The data set is a subset of the Stance Dataset \citep{Mohammad2017} that was used for detecting the attitude toward five different targets from tweets. \citet{Mohammad2017} collected the tweets via hashtags and let CrowdFlower workers annotate the tweets regarding whether the tweeter is in favor, against, or neutral toward the target of interest \citep[p.~4-7]{Mohammad2017}. The Legalization of Abortion Dataset used here contains those tweets that refer to the target `legalization of abortion'. The task associated with this data set thus is to measure attitudes toward a policy issue from text. $58.3\%$ of the tweets express an opposing and $17.9\%$ a favorable position toward legalization of abortion whilst $23.8\%$ express a neutral or no position.

{\bfseries\sffamily{3. The Wikipedia Toxic Comment Dataset}} \citep{Jigsaw2018} contains 159,571 comments from Wikipedia Talk pages that were annotated by human raters for their toxicity. On Wikipedia Talk pages contributors discuss changes to Wikipedia pages and articles.\footnote{https://en.wikipedia.org/wiki/Help:Talk\_pages} Toxic comments are comments that are obscene, threatening, insulting, or express hatred toward social groups and identities \citep{Jigsaw2018}. The task here 
is to separate toxic from non-toxic comments. Tasks in which the aim is to separate documents in which a concept (here: toxicity) occurs from documents in which the concept does not occur are common in text-based social science applications. Such tasks often constitute a first step in a text analysis in which documents that refer to concepts or entities that are of interest to the analysis have to be singled out from a large heterogeneous corpus \citep{Wankmueller2022}. Frequently, such tasks are imbalanced classification problems \citep[p.~155]{Manning:2008vf}. Here, $9.6\%$ of the comments in the data are toxic. 
The Wikipedia Toxic Comment Dataset is used to assess in how far the algorithms' performances vary with training set size. To do so, five training data sets of sizes 10,000, 5,000, 2,000, 1,000, and 500 and a test set comprising 1,000 comments are sampled uniformly at random from the 159,571 comments in the Wikipedia Toxic Comment Dataset. To account for the uncertainty induced by operating on samples of training sets, five iterations are performed. This is, the sampling is repeated five times, such that there are five sets comprising five training data sets of varying sizes.\footnote{More precisely: To get five differently sized training data sets evaluated on the same test set, the following steps are conducted:
\begin{enumerate}
\item A set of 11,000 comments is sampled uniformly at random from the 159,571 comments in the Wikipedia Toxic Comment Dataset. 
\item A random sample of 1,000 comments is drawn from the set of 11,000 comments to become the test data set. The remaining 10,000 comments constitute the first training data set.
\item From the training set of 10,000 comments, a subset of 5,000 comments is randomly drawn to become the second training set. From this subset again a smaller training subset of 2,000 texts is sampled from which a subset of 1,000 and then 500 comments are drawn.
\item To account for the induced uncertainty, steps (a) to (c) are repeated five times.
\end{enumerate}
}

\begin{figure}[!ht]
\begin{center}
\begin{tabular}{ccc}
Ethos & \hspace{-0.5cm} Abortion & \hspace{-0.5cm} Toxic \\
9a: & \hspace{-0.5cm} 9d: & \hspace{-0.5cm} 9g: \\
 \hspace{-1.2cm}  \includegraphics[width=0.36\textwidth]{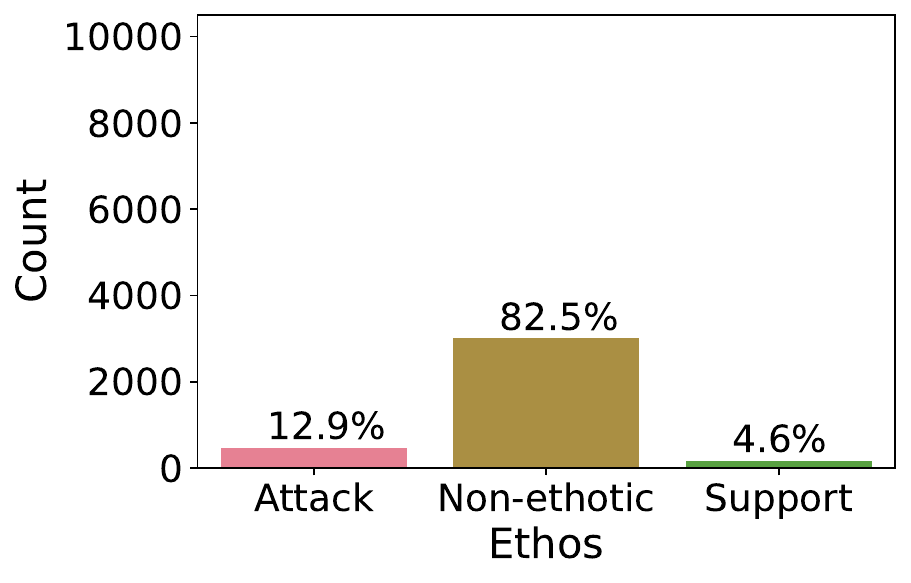} &  \hspace{-1.1cm} \includegraphics[width=0.36\textwidth]{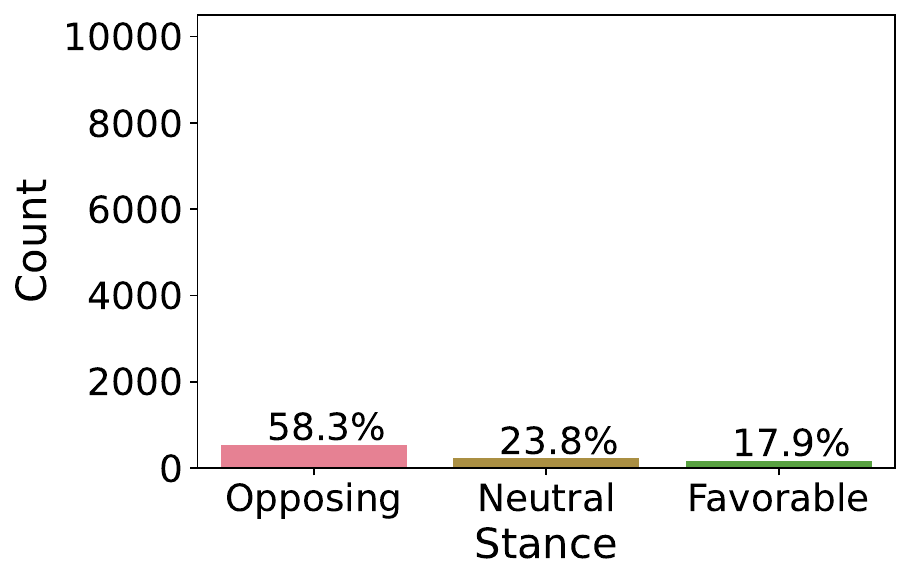} &  \hspace{-1cm} \includegraphics[width=0.36\textwidth]{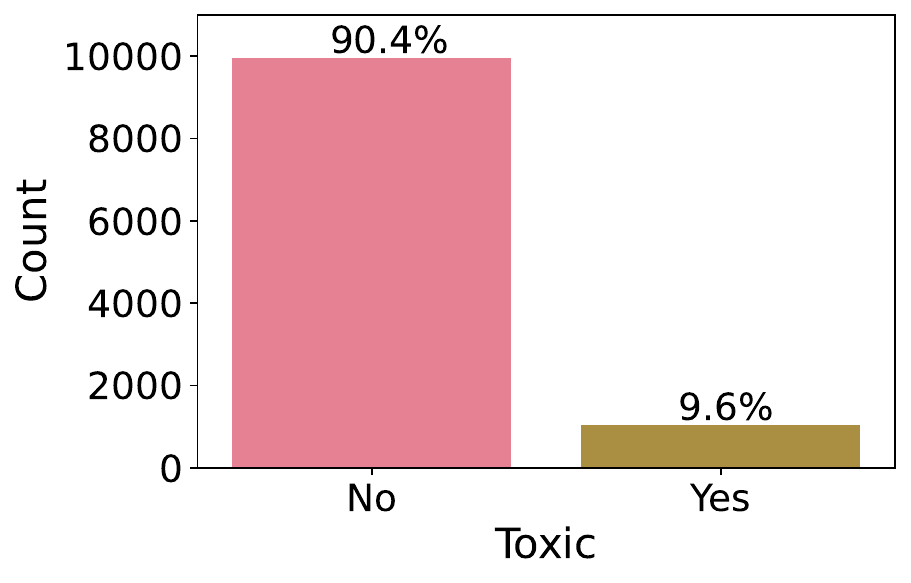} \\
 9b: & \hspace{-0.5cm} 9e: & \hspace{-0.5cm} 9h: \\
\includegraphics[width=0.32\textwidth]{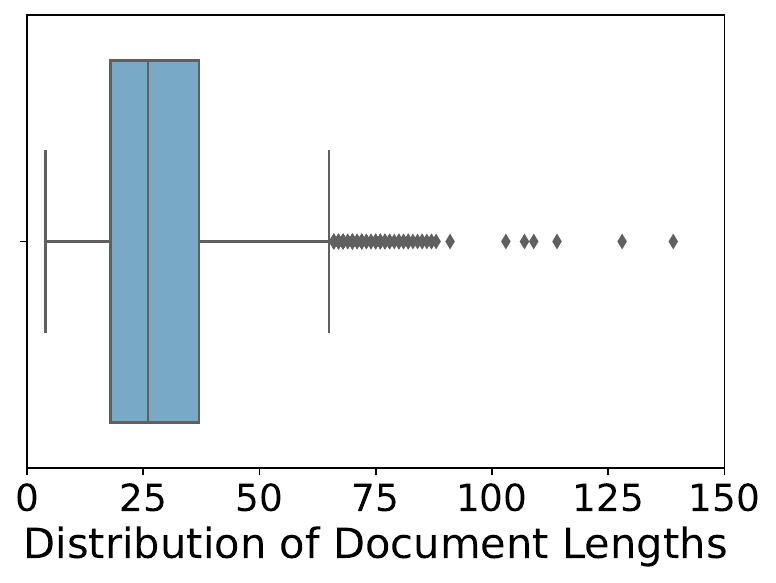} & \includegraphics[width=0.32\textwidth]{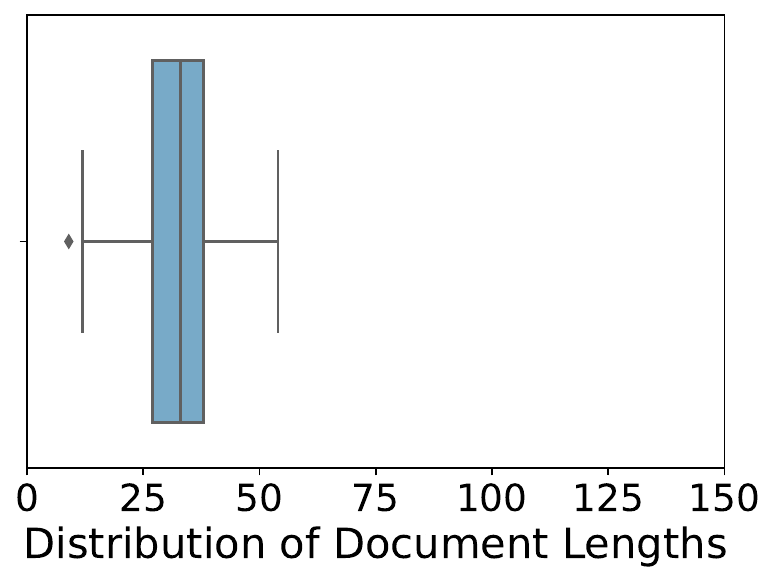} & \includegraphics[width=0.32\textwidth]{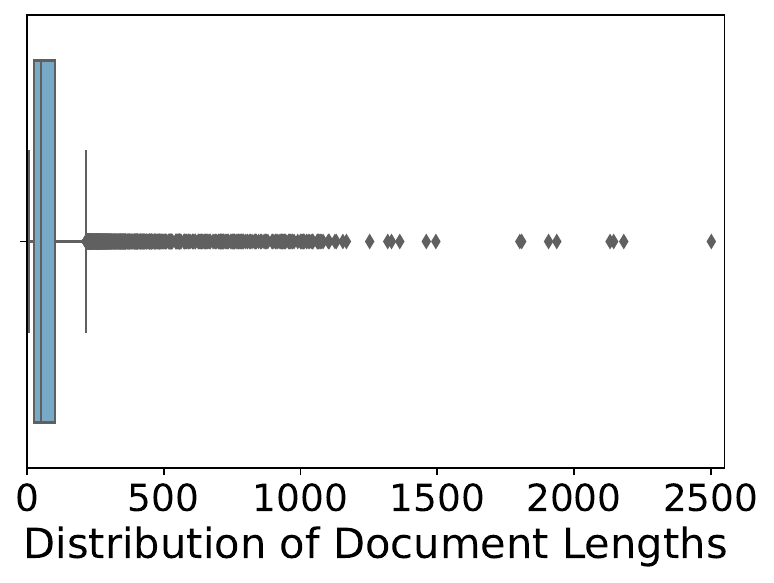} \\
 9c: & \hspace{-0.5cm} 9f: & \hspace{-0.5cm} 9i: \\
 \hspace{-1.2cm}  \includegraphics[width=0.36\textwidth]{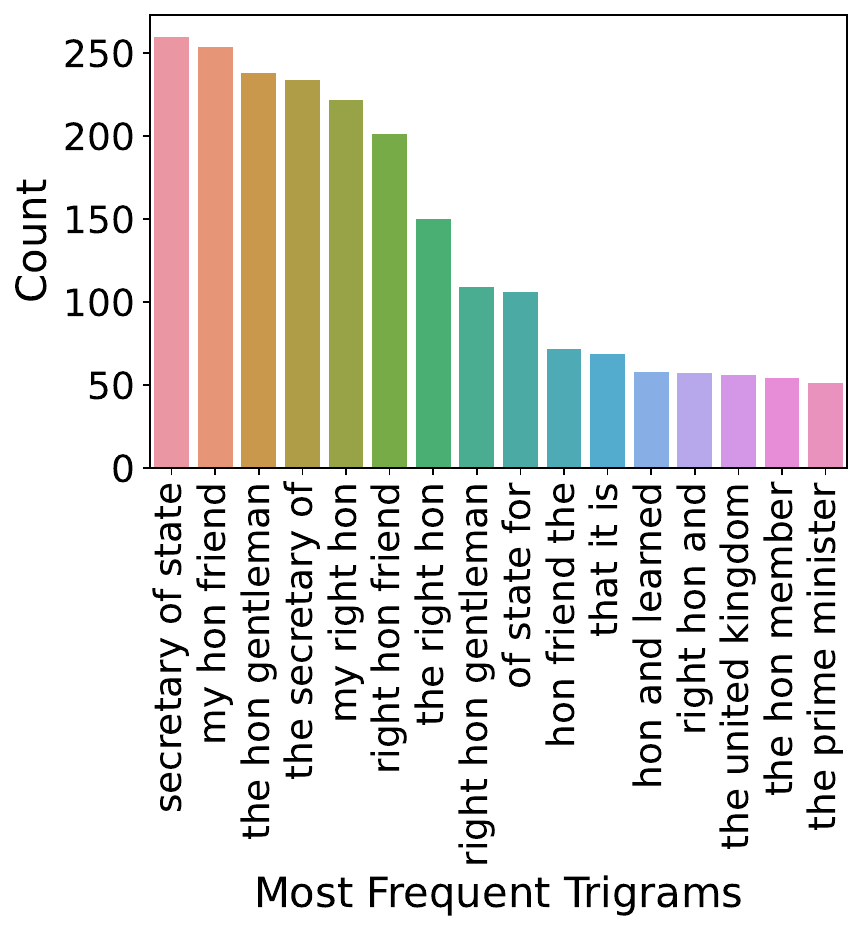} &  \hspace{-1.1cm}  \includegraphics[width=0.36\textwidth]{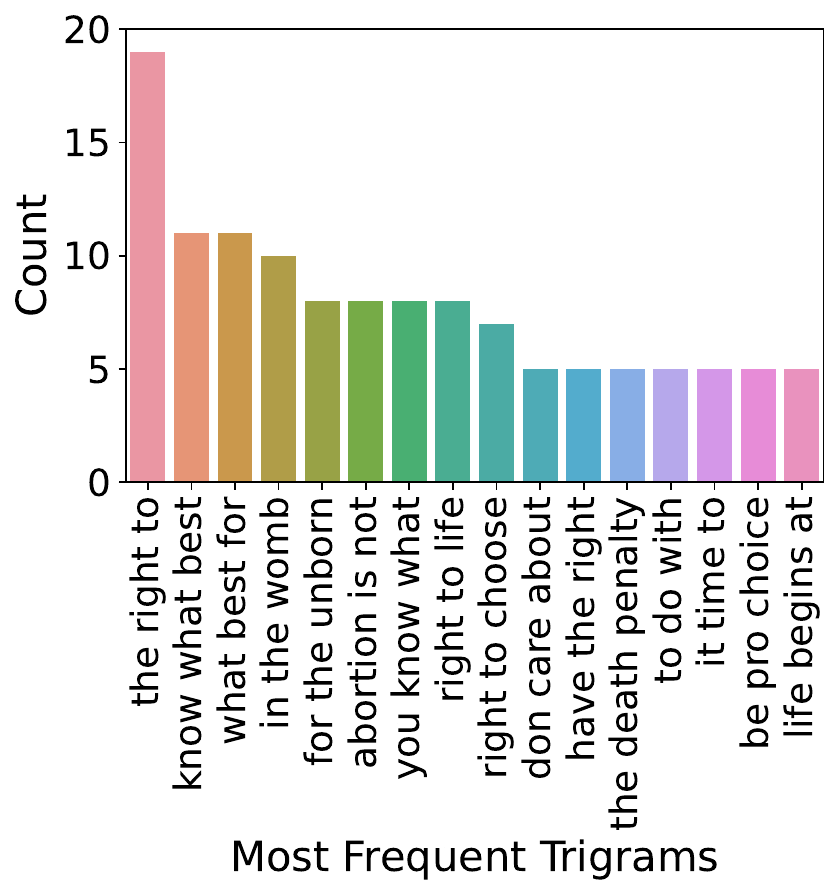} & \hspace{-1cm}  \includegraphics[width=0.36\textwidth]{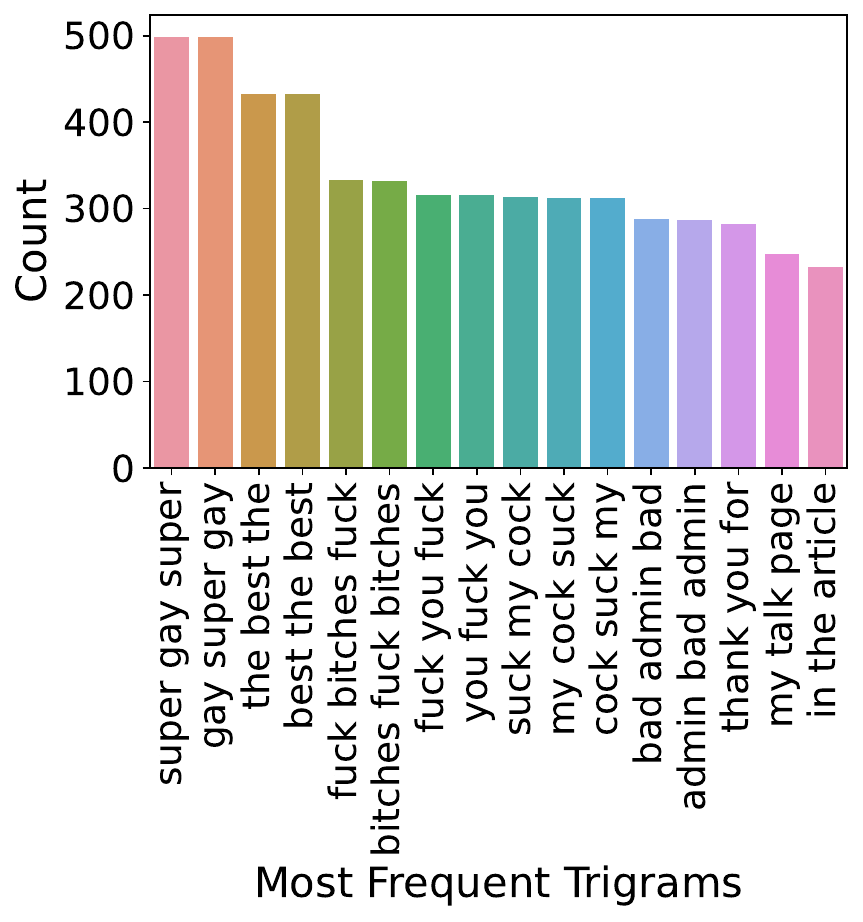} \\
\end{tabular}
\caption[Description of the Corpora.]{\textbf{Description of the Corpora.} \small{Figures 9a-9c: Ethos Datatset. Figures 9d-9f: Legalization of Abortion Dataset. Figures 9g-9i: Sample of 11,000 comments from the Wikipedia Toxic Comment Dataset. Figures 9a-9g: Class label distributions. Figures 9b-9h: Boxplots visualizing the distribution of the number of tokens per document. Figures 9c-9i: Most frequent trigrams. If possible and reasonable, the figures' axes have the same scale.}}
\label{fig:tokenlen}
\end{center}
\end{figure}

The three applications---Ethos, Abortion, and Toxic---are selected so that the methods are applied on text data that, on the one hand, represent types of texts that are often used within social science and, on the other hand, vary regarding core characteristics (for a comparison see also Figures \ref{fig:tokenlen}a to \ref{fig:tokenlen}i). Across the three applications, textual style ranges from the formal, rule-based, courteous language of parliamentary speeches over the short, statement-like nature of tweets to informal, interrelating (and at times disrespectful) comments from online discussions (to get an impression see the most frequent trigrams in Figures \ref{fig:tokenlen}c, \ref{fig:tokenlen}f, \ref{fig:tokenlen}i). The tasks associated with the applications vary with regard to the number of class labels (binary vs.~three-class classification) and the distribution over these labels (see Figures \ref{fig:tokenlen}a, \ref{fig:tokenlen}d, \ref{fig:tokenlen}g). The data sets are furthermore characterized by different document lengths (see Figures \ref{fig:tokenlen}b, \ref{fig:tokenlen}e, \ref{fig:tokenlen}h) and vary with regard to their sizes (and hence the number of data available for training).

The fewer training data, the more class labels, and the more imbalanced the distribution over class labels, the more difficult a task is likely to be. Especially with regard to imbalanced classification problems when the proportion of training instances in the minority class is small, it can be difficult to have enough training data to train an adequately performing deep neural network from scratch. As transfer learning reduces the number of required training instances, transfer learning is likely to facilitate the training of neural networks in situations of imbalance. Nevertheless, random oversampling is additionally applied here (see Section \ref{sec:ro} below). 

\subsection{Text Preprocessing for the Conventional Models} 

Two types of preprocessing procedures are employed on the raw texts to provide data representation inputs for the conventional models SVM and XGBoost:

{\bfseries\sffamily{1. Basic BOW:}} The texts are tokenized into unigrams. Punctuation, numbers, and symbols are removed in the Ethos application but kept in the other applications. Afterward, the tokens are lowercased and stemmed. Then, tokens occurring in less than a tiny share of documents (e.g.~$0.1\%$ in the Ethos application) and more than a large share of documents (e.g.~$33\%$ in the Ethos application) are excluded. Finally, the elements in the document-feature matrix are weighted such that the mere presence (1) vs.~absence (0) of each feature within each document is recorded. 

{\bfseries\sffamily{2. GloVe Representation:}} 
For each unigram that occurs at least 3 (Ethos, Abortion) or 5 (Toxic) times in the respective corpus, the 300-dimensional pretrained GloVe word vector is identified \citep{Pennington2014}.\footnote{GloVe embeddings are pretrained based on a web data corpus from CommonCrawl comprising 42 billion tokens \citep[p.~1538]{Pennington2014}.} Each document then is represented by the mean over its unigrams' GloVe word vectors. Due to making use of pretrained feature representations that are not updated during training, GloVe Representation constitutes a transfer learning approach with feature extraction. By averaging over the unigrams' word embeddings, the word order, however, is not taken into account.

\subsection{Text Preprocessing for the Transformer-Based Models} 

Before the Transformer-based models are applied, the documents are transformed to the required input format. In each document, the tokens are lowercased and the special \emph{`[CLS]'} and \emph{`[SEP]'} tokens are added. Then, each token is converted to an index identifying its input embedding and is associated with an index identifying its segment embedding. Additionally, each document is padded to the same length. In the Ethos and Legalization of Abortion corpora, this length corresponds to the maximum document length among the training set documents, which is $139$ and $54$ tokens respectively. The comments from Wikipedia Talk pages pose a problem here: An inspection of the distribution of sequence lengths in the sampled subsets of the Wikipedia Toxic Comment Dataset (see Figure \ref{fig:tokenlen}h) shows that the vast majority of comments are shorter than the maximum number of $512$ tokens that BERT and RoBERTa can distinguish---but there is a long tail of comments exceeding $512$ tokens. To address this issue, two different approaches are explored: For BERT, following the best strategy identified by \citet{Sun2019}, in each comment that is longer than 512 tokens, only the first 128 and the last 382 tokens are kept while the tokens positioned in the middle are removed. RoBERTa, in contrast, is replaced with the Longformer in the Toxic application. For the Longformer the sequence length is set to $2*512 =$ 1,024 tokens. This ensures that in each run only a small one- or two-digit number of sequences that are longer than 1,024 tokens are truncated by removing tokens from the middle whilst padding the texts to a shared length that still can be processed with given memory restrictions.

\subsection{Training on the Target Tasks} \label{sec:ro}

Pretrained BERT, RoBERTa and Longformer models are accessed from the Hugging Face's Transformers library (\url{https://huggingface.co/}) and then are adapted to each of the target tasks. When doing so, the Adam algorithm as introduced by \citet{Loshchilov2019} with a linearly decaying global learning rate, no warmup, and no weight decay is employed. Dropout is set to $0.1$. To fine-tune the models within the memory resources provided by Colab, small batch sizes are used. In the Ethos and Abortion applications, a batch size of $16$ is selected. A batch in the Toxic application comprises $8$ (and for the Longformer $4$) text instances.\footnote{Note that when selecting a small batch size (e.g.~because of memory restrictions) this is not a disadvantage but rather the opposite: Research suggests that smaller batch sizes not only require less memory but also have better generalization performances \citep{Keskar2017, Masters2018}. To ensure that the learning process with small batch sizes does not get too volatile, one merely has to account for the fact that smaller batch sizes require correspondingly smaller learning rates \citep{Brownlee2020a}.} Moreover, for the pretrained models, the base size of the model architecture is used instead of the large or extra large model versions. So, for example, BERT\textsubscript{BASE} instead of BERT\textsubscript{LARGE} is applied. Larger models are likely to lead to higher performances. Yet, because they have more parameters, it takes more computing resources to fine-tune them and---especially for small data sets---fine-tuning might lead to results that vary more noticeably across random restarts \citep[p.~4176]{Devlin2019}. The training data in the Ethos and Wikipedia Toxic Comment Datasets are randomly oversampled.\footnote{In random oversampling, instances of the minority classes are randomly sampled with replacement and added as identical copies to the training data such that the training data become more balanced \citep{Brownlee2020}. The presence of multiple minority class copies in the training data increases the loss caused by misclassifying minority class instances and hence induces the algorithm to put a stronger focus on correctly classifying minority class examples.} To address the class imbalances but also to prevent too strong overfitting on the training data, the minority classes are moderately oversampled such that the size of the minority classes is $1/4$th the size of the majority class. 

For each evaluated combination of an algorithm and a preprocessing procedure, a grid search across sets of hyperparameter values is performed via five-fold cross-validation on the training set. For the Transformer-based models, the hyperparameter grid search explores model performances across combinations of different learning rates and epoch numbers. Accounting for the fact that in the optimization process the gradient updates are conducted based on small batches, relatively small global Adam learning rates $\{$1e-05, 2e-05, 3e-05$\}$ are inspected. The number of epochs explored is $\{2, 3, 4\}$.\footnote{Note that for the Longformer (for which a batch size of 4 is used) the learning rate is set to 1e-05 and the number of epochs explored is $\{2, 3\}$. Hyperparameter tuning for the SVMs compares a linear kernel and a Radial Basis Function kernel. The explored values are $\{0.1, 1.0, 10.0\}$ for penalty weight $\mathcal C$, and---in the case of the Radial Basis Function kernel---values of $\{0.001, 0.01, 0.1\}$ are inspected for parameter $\gamma$, that specifies the radius of influence for single training examples. Regarding the XGBoost algorithms, the grid search explores $50$ vs.~$250$ trees, each with a maximum depth of $5$ vs.~$8$, and XGBoost learning rates of $0.001, 0.01,$ and $0.1$. For details on SVM and XGBoost hyperparameters see also \citet{SklearnUserGuide2020b, SklearnUserGuide2020a} and \citet{XGBoostUserGuide2020}.}

At the end of hyperparameter tuning, the best performing set of hyperparameters according to the macro-averaged $F_1$-Score and overfitting considerations is selected. Then the model with the chosen hyperparameter setting is trained on the entire training data set and evaluated on the test set via the macro-averaged $F_1$-Score.

\subsection{Results} \label{sec:res097}

\begin{figure}[htb]
\begin{center}
\includegraphics[width=0.6\paperwidth]{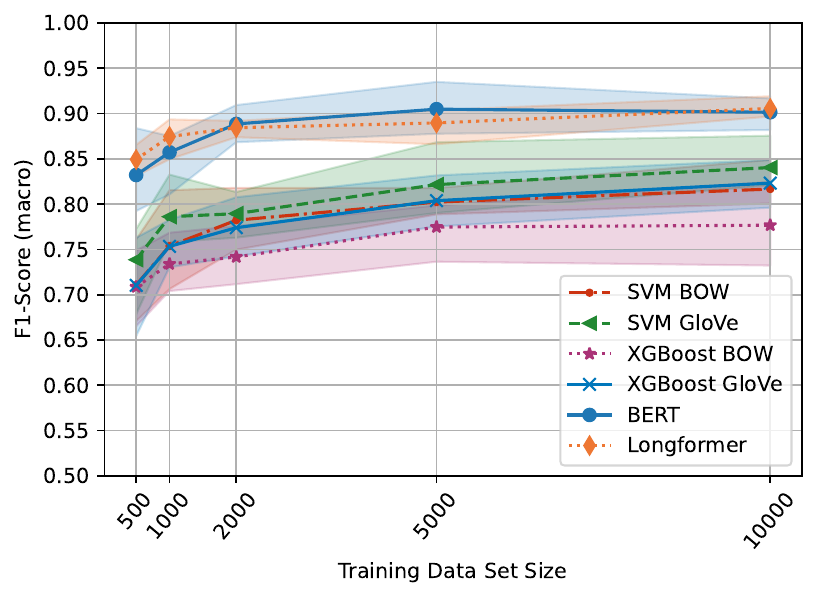} \\
\caption[Performances on Toxic Application with Varying Training Data Set Sizes.]{\textbf{Performances on Toxic Application with Varying Training Data Set Sizes.} \small{For each training data set size and each model, the plotted symbols indicate the mean of the test set macro-averaged $F_1$-Scores across the five iterations. The shaded areas range from the minimum to the maximum macro-averaged $F_1$-Score obtained across the five iterations.}}
\label{fig:wikipedia}
\end{center}
\end{figure}

The results for the Ethos, Abortion, and Toxic classification tasks are presented in Table \ref{tab:validation_results}. 
Figure \ref{fig:wikipedia} additionally visualizes the results for the Toxic application. 

\begin{table}[htb]
\begin{center}
\begin{small}
\begin{tabular}{lrrrrrrr}
\toprule
{} &  Ethos &  Abortion &  Toxic0.5K &  Toxic1K &  Toxic2K &  Toxic5K &  Toxic10K \\
\midrule
SVM BOW            &  0.566 &     0.526 &     0.711 &      0.754 &      0.782 &      0.802 &       0.817 \\
SVM GloVe          &  0.585 &     0.545 &     0.739 &      0.786 &      0.789 &      0.822 &       0.840 \\
XGBoost BOW        &  0.563 &     0.540 &     0.709 &      0.734 &      0.742 &      0.775 &       0.777 \\
XGBoost GloVe      &  0.513 &     0.506 &     0.710 &      0.753 &      0.774 &      0.804 &       0.823 \\
BERT                             &  0.695 &     0.593 &     0.832 &      0.857 &      \maxf{0.888} &      \maxf{0.905} &   0.901 \\
RoBERTa/Longf. &  \maxf{0.747} &     \maxf{0.617} &     \maxf{0.849} &      \maxf{0.875} &      0.884 &      0.890 &      \maxf{0.906} \\
\bottomrule
\end{tabular}
\end{small}
\end{center}
\caption[Macro-Averaged $F_1$-Scores]{\textbf{Macro-Averaged $F_1$-Scores.} \small{Macro-averaged $F_1$-Scores of the evaluated models for the Ethos, Abortion and Toxic classification tasks. If there are $C$ classes such that $y_i \in \{\mathcal{G}_1, \dots, \mathcal{G}_c, \dots, \mathcal{G}_C\}$, the $F_1$-Score for a particular class $\mathcal{G}_c$ is the harmonic mean of precision and recall for this class \citep[p.~156]{Manning:2008vf}. Recall indicates what proportion of instances that truly belong to class $\mathcal{G}_c$ have been correctly classified as being in $\mathcal{G}_c$. Precision informs about what share of instances that have been predicted to be in class $\mathcal{G}_c$ truly belong to class $\mathcal{G}_c$. The $F_1$-Score can range from 0 to 1 with 1 being the highest value signifying perfect classification. The macro-averaged $F_1$-Score is the unweighted mean of the $F_1$-Scores of each class \citep{SklearnUserGuide2020}. By not weighting the $F_1$-Scores according to class sizes, algorithms that are bad at predicting the minority classes are penalized more severely \citep{SklearnUserGuide2020}. In the Toxic application, for each tested training data set size, $\{$500, 1,000, 2,000, 5,000, 10,000$\}$, the mean of the macro-averaged $F_1$-Scores across the five iterations is shown. The column labeled Toxic0.5K gives the mean of the macro-averaged $F_1$-Scores for the Toxic classification task with a training set size of $500$ instances. SVM BOW and XGBoost BOW denote SVM and XGBoost with bag-of-words preprocessing. SVM GloVe and XGBoost GloVe refer to SVM and XGBoost with GloVe representations. In RoBERTa/Longf., RoBERTa is applied for the Ethos and the Abortion target tasks whereas the Longformer is used for the Toxic comment classification tasks. Gray colored cells highlight the best performing model for the task.}}
\label{tab:validation_results}
\end{table}

Across all evaluated classification tasks and training data set sizes, the Transformer-based models for transfer learning tend to achieve higher macro-averaged $F_1$-Scores than the conventional machine learning algorithms SVM and XGBoost. As has been observed before, the classic machine learning algorithms produce acceptable results given the relatively simple representations of text they are applied on. However, when compared on the basis of the (mean) macro-averaged $F_1$-Scores presented in Table \ref{tab:validation_results}, BERT, RoBERTa, and the Longformer consistently outperform the best performing conventional model by a margin of at minimum $0.05$ to $0.11$. These moderate to considerably higher prediction performances across all evaluated textual styles, sequence lengths, and especially the smaller training data set sizes, demonstrate the potential benefits that neural transfer learning with Transformers can bring to analyses in which a researcher aims at having a valid text-based measure of a concept and thus seeks to replicate human codings as accurately as possible. Even if only a small to medium-sized training data set is available, social scientists that apply Transformer-based models in a transfer learning setting are likely to obtain more valid measures for concepts that they measure from texts. A detailed examination of the macro-averaged $F_1$-Scores reveals further findings: 
\begin{itemize}
\item Averaged GloVe representations partly, though not consistently, produce a slight advantage over basic BOW preprocessing. This emphasizes that employing transfer learning on conventional machine learning algorithms by extracting pretrained features (here: GloVe embeddings) and taking them as the data representation input might be beneficial---even if averaging over the embeddings erases information on word order and dependencies.

\item For the Ethos and Abortion applications, RoBERTa outperforms BERT to a small extent. This finding is consistent with previous research \citep[p.~7]{Liu2019}. In general, it is difficult to disentangle the effects of single modifications of the original BERT architecture and pretraining settings that BERT-extensions as RoBERTa implement \citep{Assenmacher2020}. It is likely, however, that one important contribution is the longer pretraining on more and more varied data. Whereas BERT is pretrained on a corpus of books and Wikipedia articles, RoBERTa is additionally pretrained on three more large data sets that are based on text passages from the web \citep[p.~5-6]{Liu2019}. The larger and more heterogeneous pretraining corpus is likely to enable RoBERTa to produce representations that better generalize across a diverse set of target task corpora as inspected here. 

\item In the Ethos application, BERT and RoBERTa do not only exceed the performances of the other evaluated models but also the best performing model developed by \citet{Duthie2018}. To differentiate non-ethotic from positive and from negative ethotic sentences, \citet{Duthie2018} had created an elaborate NLP pipeline including a POS tagger, dependency parsing, anaphora resolution, entity extraction, sentiment classification, and a deep RNN. \citet[p.~4045]{Duthie2018} report a macro-averaged $F_1$-Score of $0.65$ for their best model. BERT and RoBERTa here surpass this performance. As the pretrained BERT and ROBERTa models are simply fine-tuned to the Ethos classification target task without implementing (and having to come up with) an extensive and complex preprocessing pipeline, this demonstrates the efficiency and power of transfer learning.
 
\item With all models achieving only mediocre performances, the Abortion classification task, for which only $653$ short Tweets are available as training instances, seems to be especially difficult. BERT and RoBERTa still surpass SVM and XGBoost but with a slightly smaller margin. By applying an SVM with a linear kernel based on word and character n-gram feature representations, \citet[p.~13]{Mohammad2017} reach classification performance levels that are higher than the ones reached by the models presented here.\footnote{\citet{Mohammad2017} merely compute the $F_1$-Score for the favorable and opposing categories leaving out the neutral position. They report a score of $0.664$ for their $N$-gram based SVM classifier \citep[p.~13]{Mohammad2017}. Here the corresponding score values are $0.633$ for BERT, $0.648$ for RoBERTa as well as $0.616$ for the best performing conventional model SVM GloVe.} The Abortion classification task with short tweets in which the mere $N$-grams tend to be indicative of the stance toward the issue \citep[p.~13]{Mohammad2017}, seems to be an example of a task in which deep learning models only produce a moderate advantage or---if it is easy to select BOW representations that very well capture linguistic variation that helps in discriminating the texts into the categories
---even no advantage over traditional machine learning algorithms.

\item Across all evaluated training data set sizes, the Transformer-based models with transfer learning tend to be better at solving the Toxic comment classification task compared to the conventional algorithms (see Figure \ref{fig:wikipedia}). As is to be expected, the performance levels for all models decrease with decreasing training data set sizes. Yet although the neural models have much more parameters to learn, their macro-averaged $F_1$-Scores do not decrease more sharply than those of the traditional machine learning algorithms. Especially as training data sets become small, the effectiveness of representations from pretrained models becomes salient. Here, the pretrained models seem to function as a quite effective input to the target task.

\item Whereas the Longformer processes text sequences of 1,024 tokens, the input sequences for BERT were truncated at 512 tokens for the Toxic application. Despite this large difference in sequence lengths, BERT only slightly underperforms compared to the Longformer---and matches the Longformer for larger training data set sizes. As only a small share of comments in the Wikipedia Toxic Comment Dataset are longer than 512 tokens (see again Figure \ref{fig:tokenlen}h), the Longformer's advantage of being able to process longer text sequences does not materialize here. Removing tokens from the middle of comments that exceed 512 tokens does not harm BERT's prediction performance and is an effective workaround in this application. For applications based on corpora in which the mass of the sequence length distribution is above 512 tokens, however, the Longformer's ability to process and capture the information contained in these longer documents, is likely to be important for prediction performance.

\item The time consumed during training differs substantively between the conventional and the Transformer models. Larger training data sets and smaller batch sizes increase the time required for fine-tuning the pretrained Transformer models. 
Across the applications presented here, the absolute training time varies between 1 and 276 seconds for SVM BOW, between 32 and 2,272 seconds for BERT and 31 to 9,707 seconds for RoBERTa/Longformer. 

\item An additional analysis that explores the effectiveness of zero-shot learning is conducted (see Appendix \ref{appendix:zslimpl}). Across all applications, across both employed pretrained models (RoBERTa and BART), and across all explored hypothesis formulations\footnote{A hypothesis formulation is an additional textual input required in the employed natural language inference (NLI) framework for zero-shot learning (see Appendix \ref{appendix:zslimpl}).}, the macro-averaged $F_1$-Scores are mediocre and substantially lower than for the fine-tuned models. The highest macro averaged $F_1$-Scores from zero-shot learning are 0.200 (Ethos), 0.455 (Abortion), and 0.470 (Toxic). Even if the prediction performances of the here implemented zero-shot learning framework are not sufficiently high in order to be applied in research projects in which researchers seek to as accurately as possible measure a priori-defined concepts from texts, this analysis nevertheless demonstrates what can be achieved with representations from pretrained models alone.
\end{itemize}

\section{Discussion}  \label{sec:conclusion2}

Advances in NLP research on transfer learning and the attention mechanism, that is incorporated in the Transformer, have paved the way to a new mode of learning in which researchers can hope to achieve higher prediction performances by taking a readily available pretrained model and fine-tuning it, with a manageable amount of resources, to their NLP task of interest \citep{Bommasani2021}. These advances are of interest to social scientists that attempt to have valid measures of concepts from text data but may have limited amounts of training data and resources. To use the potential advantages for social science text analysis, this study has presented and applied Transformer-based models for transfer learning. In the supervised classification tasks evaluated in this study, transfer learning with Transformer models outperformed traditional machine learning across all tasks and data set sizes.

Employing transfer learning with Transformer-based models, however, will not always perform better compared to other machine learning algorithms and is not the most adequate strategy for each and every text-based research question. As the attention mechanism is specialized in capturing dependencies and contextual meanings, these models are likely to generate more accurate predictions if contextual information and long-range dependencies between tokens are relevant for the task at hand. They are less likely to provide much of an advantage if the function to be learned between textual inputs and desired outputs is less complex---for example because single $N$-grams are strongly indicative of class labels (see e.g.~the Abortion application).

Transformer-based models for transfer learning furthermore are useful for supervised classification tasks in which the aim is to achieve an as high as possible prediction performance rather than having an interpretable model. Social scientists whose primary goal is to have as precise as possible text-based measures for concepts they employ may find Transformer-based models for transfer learning highly useful, whereas researchers whose primary goal is to know which textual features are most important in discriminating between class labeled documents \citep[e.g.][]{Slapin2020} are likely to be better served with directly interpretable models. 

Moreover, due to the sequence length limitations of Transformer-based models, the applicability of these models is currently restricted to NLP tasks that operate on only moderately long text sequences. Research that seeks to reduce the memory resources consumed by the attention mechanism and thus allows for processing longer text sequences is highly important because it opens up the potential of Transformers for a wider range of social science text analyses.

As neural transfer learning with Transformers is the basis of larger developments within AI research \citep{Bommasani2021}, it is important that social scientists understand these new learning modes and models---such that these learning modes and models can be correctly and fruitfully applied and their risks critically assessed.

\newpage
\appendix
{\LARGE{\bfseries\sffamily{Appendix}}}

\section{Introduction to Deep Learning}  \label{appendix:introdl} 
This section provides an introduction to the basics of deep learning. First, based on the example of feedforward neural networks the core elements of neural network architectures are explicated. Then, the optimization process via stochastic gradient descent with backpropagation \citep{Rumelhart1986} will be presented. Subsequently, the architecture of recurrent neural networks (RNNs) \citep{Elman1990} is outlined. 

\subsection{Feedforward Neural Network}
\begin{figure}[htb]
\begin{center}
\footnotesize
\includegraphics[width=0.75\textwidth]{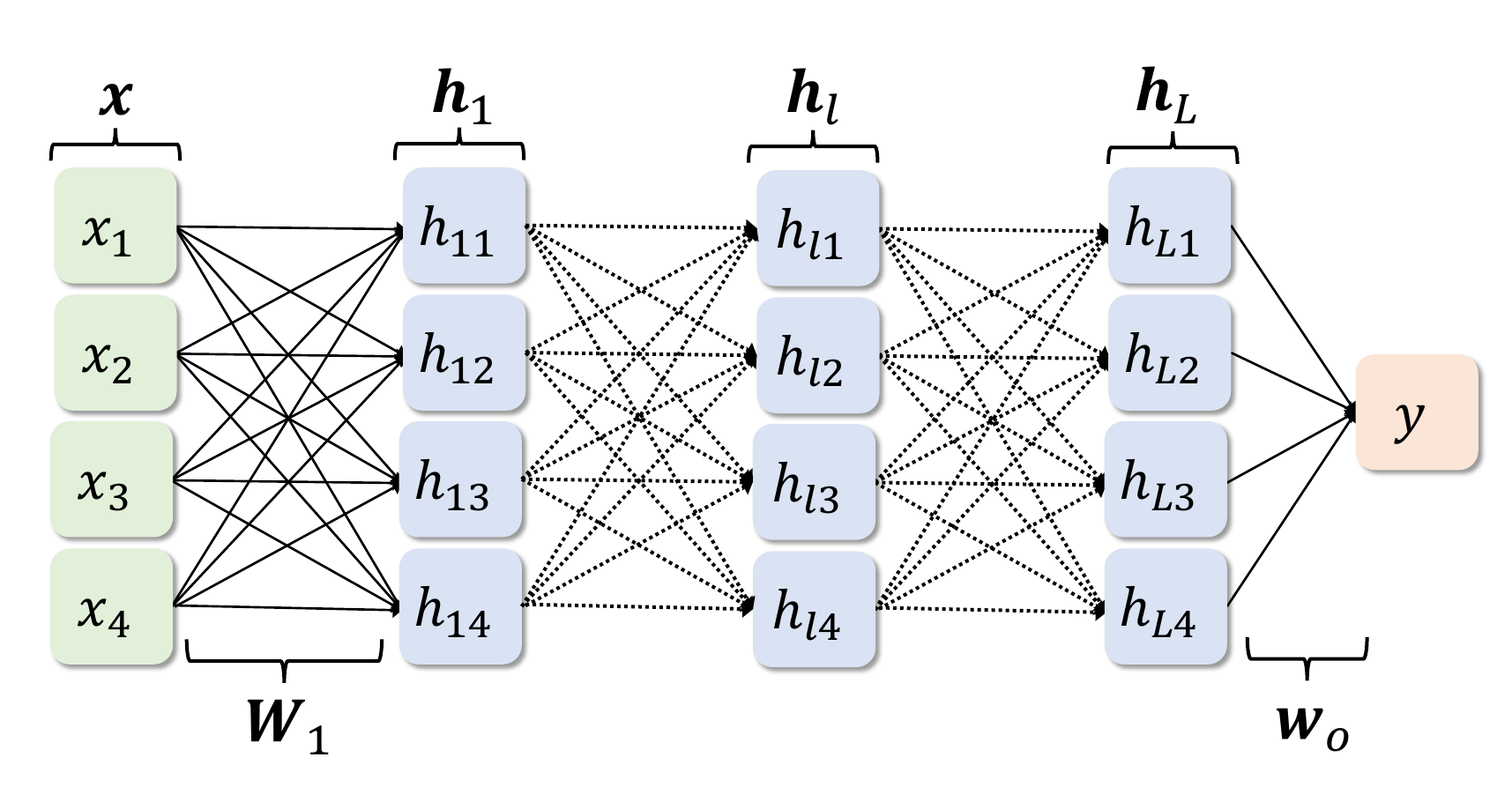}\\
\vspace{0.5cm}
\caption[A Feedforward Neural Network]{\textbf{A Feedforward Neural Network.} \small{Feedforward neural network with $L$ hidden layers, four units per hidden layer and scalar output $y$. The solid lines indicate the linear transformations of weight matrix $\bm W_1$. The dotted lines indicate the connections between several consecutive hidden layers.}}
\label{fig:fnn}
\end{center}
\end{figure}

The most elementary deep learning model is a feedforward neural network \citep[p.~164]{Goodfellow2016}. A feedforward neural network with $L$ hidden layers, vector input $\bm{x}$, and a scalar output $y$ can be visualized as in Figure \ref{fig:fnn} and be described as follows:\footnote{Note that here, in accordance with standard notation in machine learning and NLP \citep[p.~346]{Goldberg2016}, column vectors are used.}
\begin{equation} \label{eq:fnn1}
\bm{h}_{1} = \sigma_1(\bm{W}_1\bm{x} + \bm{b}_1)
\end{equation}
\begin{equation} \label{eq:fnn2}
\bm{h}_{2} = \sigma_2(\bm{W}_2\bm{h}_{1} + \bm{b}_2)
\end{equation}
\begin{equation*}
\dots
\end{equation*}
\begin{equation} \label{eq:fnn3}
\bm{h}_{l} = \sigma_l(\bm{W}_l\bm{h}_{l-1} + \bm{b}_l)
\end{equation}
\begin{equation*}
\dots
\end{equation*}
\begin{equation} \label{eq:fnn4}
y = \sigma_o(\bm{w}_o\bm{h}_{L} + b_o)
\end{equation}

The input to the neural network is the $K_0$-dimensional vector $\bm{x}$ (see Equation \ref{eq:fnn1}). $\bm{x}$ enters an affine function characterized by weight matrix $\bm{W}_1$ and bias vector $\bm{b}_1$, where $\bm{W}_1 \in \mathbb{R}^{K_1 \times K_0}$, and $\bm{b}_1 \in \mathbb{R}^{K_1}$. $\sigma_1$ is a nonlinear activation function and $\bm{h}_1 \in  \mathbb{R}^{K_1}$ is the $K_1$-dimensional representation of the data in the first hidden layer. This is, the neural networks takes the input data $\bm{x}$ and via combining an affine function with a nonlinear activation function generates a new, transformed representation of the original input: $\bm{h}_1$. The hidden state $\bm{h}_1$ in turn serves as the input for the next layer that produces representation $\bm{h}_2 \in \mathbb{R}^{K_2}$. This continues through the layers until the last hidden representation, $\bm{h}_L \in \mathbb{R}^{K_L}$, enters the output layer (see Equation \ref{eq:fnn4}). 

The activation functions in neural networks are typically chosen to be nonlinear \citep[p.~168]{Goodfellow2016}. The reason is that if the activation functions were set to be linear, the output of the neural network would merely be a linear function of $\bm x$ \citep[p.~168]{Goodfellow2016}. Hence, the use of nonlinear activation functions is essential for the capacity of neural networks to approximate a wide range of functions and highly complex functions \citep[p.~31]{Ruder2019}. 

In the hidden layers, the Rectified Linear Unit (ReLU) \citep{Nair2010} is often used as an activation function $\sigma_l$ \citep[p.~171]{Goodfellow2016}. If $\bm q = [q_1, \dots, q_{k}, \dots, q_{K}]$ is the $K$-dimensional vector resulting from the affine transformation in the $l$th hidden layer, $\bm q = \bm{W}_l\bm{h}_{l-1} + \bm{b}_l$ (see Equation \ref{eq:fnn3}), then ReLU is applied on each element $q_k$:
\begin{equation} \label{eq:relu}
\sigma_l(\bm q)_k = max\{0,q_k\}
\end{equation}
$\sigma_l(\bm q)_k$ then is the $k$th element of hidden state vector $\bm h_l$.\footnote{Activation functions that are similar to ReLU are the Exponential Linear Unit (ELU) \citep{Clevert2016}, Leaky ReLU and the Gaussian Error Linear Unit (GELU) \citep{Hendrycks2016}. The latter is used in BERT \citep{Devlin2019}.} 

In the output layer, the activation function $\sigma_o$ is selected so as to produce an output that matches the task-specific type of output values. In binary classification tasks with $y_i \in \{0,1\}$ the standard logistic function, often simply referred to as the sigmoid function, is a common choice \citep[p.~179-180]{Goodfellow2016}. For a single observational unit $i$, the sigmoid function's scalar output value gives the probability that $y_i = 1$. If $y_i$, however, can assume one out of $C$ unordered response category values, $y_i \in \{\mathcal{G}_1, \dots, \mathcal{G}_c, \dots, \mathcal{G}_C\}$, then the softmax function (which is a generalization of the sigmoid function that takes as an input and produces as an output a vector of length $C$) is typically employed \citep[p.~180-181]{Goodfellow2016}. For the $i$th example, the $c$th element of the softmax output vector gives the predicted probability that unit $i$ falls into the $c$th class.

\subsection{Optimization: Gradient Descent with Backpropagation}

In supervised learning tasks, a neural network is provided with input $\bm{x}_i$ and corresponding output $y_i$ for each training example. All the weights and bias terms are parameters to be learned in the process of optimization \citep[p.~165]{Goodfellow2016}. The set of parameters hence is $\bm{\theta} = \{ \bm{W}_1, \dots, \bm{W}_l, \dots, \bm{W}_L, \bm{W}_o, \bm{b}_1, \dots, \bm{b}_l, \dots, \bm{b}_L, \bm{b}_o \}$.

For a single training example $i$, the loss function $\mathcal L(y_i, \mathsf{f}(\bm{x}_i, \bm{\tilde \theta}))$ measures the discrepancy between the value predicted for unit $i$ by model $\mathsf{f}(\bm{x}_i, \bm{\tilde \theta})$, that is characterized by the estimated parameter set $\bm{\tilde \theta}$, and the true value $y_i$ \citep[p.~832]{Vapnik1991}. In the optimization process, the aim is to find the set of values for the weights and biases that minimizes the average of the observed losses over all training set instances, also known as the empirical risk: $\mathcal{R}_{emp}(\bm{\tilde \theta}) = \frac{1}{N} \sum_{i = 1}^{N} \mathcal L(y_i, \mathsf{f}(\bm{x}_i, \bm{\tilde \theta}))$ \citep[p.~272-273]{Goodfellow2016}.

Neural networks commonly employ variants of gradient descent with backpropagation in the optimization process \citep[p.~173]{Goodfellow2016}. To approach the local minimum of the empirical risk function, the gradient descent algorithm makes use of the fact that the direction of the negative gradient of function $\mathcal{R}_{emp}$ at current point $\bm{\tilde{\theta}_j}$ gives the direction in which $\mathcal{R}_{emp}$ is decreasing fastest---the direction of the steepest descent \citep[p.~83]{Goodfellow2016}. The gradient is a vector of partial derivatives. It is the derivative of $\mathcal{R}_{emp}$ at point $\bm{\tilde{\theta}_j}$ and is denoted as $\nabla_{\bm{\tilde{\theta}_{j}}} \mathcal{R}_{emp}(\bm{\tilde{\theta}_j})$ \citep[p.~2]{Johnson2017}.

In the $j$th iteration, the gradient descent algorithm computes the negative gradient of $\mathcal{R}_{emp}$ at current point $\bm{\tilde{\theta}_j}$ and then changes its position from $\bm{\tilde{\theta}_j}$ into the direction of the negative gradient \citep[p.~83-84]{Goodfellow2016}:
\begin{equation} \label{eq:learn}
\bm{\tilde{\theta}_{j+1}} = \bm{\tilde{\theta}_{j}} - \eta \nabla_{\bm{\tilde{\theta}_{j}}} \mathcal{R}_{emp}(\bm{\tilde{\theta}_j})
\end{equation}
where $\eta \in \mathbb{R}_+$ is the learning rate. If $\eta$ is small enough, then $\mathcal{R}_{emp}(\bm{\tilde{\theta}_j}) \geq \mathcal{R}_{emp}(\bm{\tilde{\theta}_{j+1}}) \geq \mathcal{R}_{emp}(\bm{\tilde{\theta}_{j+2}}) \geq \dots$. This is, repeatedly updating into the direction of the negative gradient with a suitably small learning rate $\eta$, will generate a sequence moving toward the local minimum \citep{Li2020a}. 

In each iteration, the gradients for all parameters are computed via the backpropagation algorithm \citep{Rumelhart1986}.\footnote{The backpropagation algorithm makes use of the chain rule to compute the gradients. Helpful introductions to the backpropagation algorithm can be found in \citet{Li2020a}, \citet{Li2020b} and \citet{Hansen2019}.} A very frequently employed approach, known as mini-batch stochastic gradient descent, is to compute the gradients based on a random sample, a mini-batch, of $S$ training set observations \citep[p.~275-276, 291]{Goodfellow2016}: 
\begin{equation}
\nabla_{\bm{\tilde{\theta}_{j}}} \mathcal{R}_{emp}(\bm{\tilde{\theta}_j}) = \frac{1}{S} \sum_{s = 1}^S \nabla_{\bm{\tilde{\theta}_j}} \mathcal L(y_s, \mathsf{f}(\bm{x}_s, \bm{\tilde{ \theta}_j}))
\end{equation}
The learning rate $\eta$ and the size of the mini-batch $S$ are hyperparameters in training neural networks. Especially the learning rate is often attended to carefully \citep{Li2020a}. A too high learning rate leads to large fluctuations in the loss function values, whereas a too low learning rate implies slow convergence and risks that the learning process does not move away from a non-optimal region with a high loss value \citep[p.~291]{Goodfellow2016}. Commonly, the learning rate is set to vary over the course of the training process \citep[p.~290-291]{Goodfellow2016}. Furthermore, there are variants of stochastic gradient descent, e.g.~AdaGrad \citep{Duchi2011}, RMSProp \citep{Hinton2012}, and Adam \citep{Kingma2014}, that have a different learning rate for each parameter \citep[p.~303-305]{Goodfellow2016}.

\subsection{Recurrent Neural Networks} \label{appendix:rnns}
\begin{figure}[htb]
\begin{center}
\footnotesize
\includegraphics[width=0.6\textwidth]{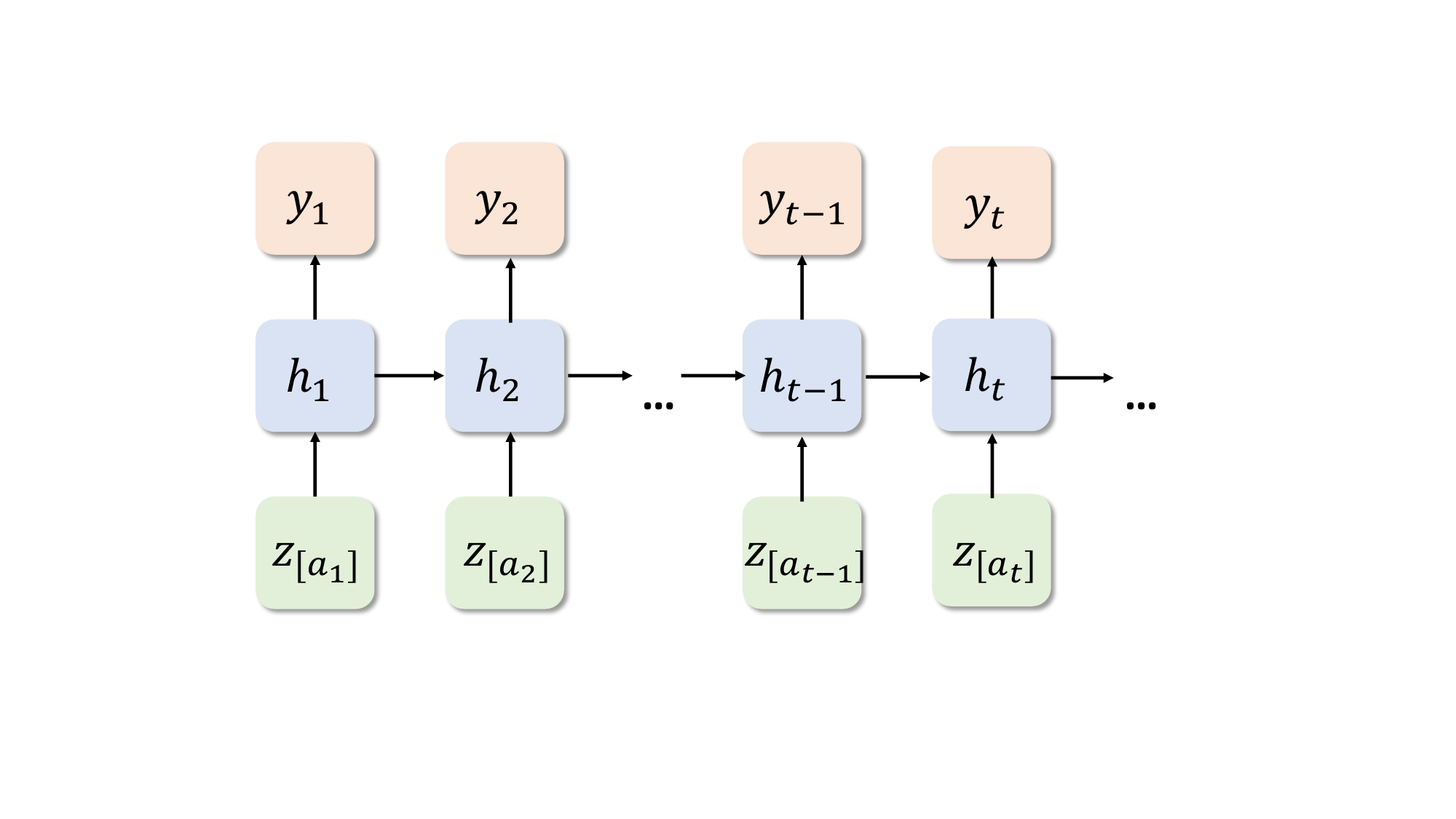}
\caption[A Recurrent Neural Network]{\textbf{A Recurrent Neural Network.} \small{Architecture of a basic RNN unfolded through time. At time step $t$, the hidden state $\bm{h}_t$ is a function of the previous hidden state, $\bm{h}_{t-1}$, and current input embedding $\bm{z}_{[a_t]}$. $\bm{y}_t$ is the output produced at $t$.}}
\label{fig:rnn}
\end{center}
\end{figure}

The recurrent neural network (RNN) \citep{Elman1990} is the most basic neural network to process sequential input data of variable length such as texts \citep[p.~367]{Goodfellow2016}. Given an input sequence of 
$T$ input embeddings $(\bm{z}_{[a_1]}, \dots, \bm{z}_{[a_t]}, \dots, \bm{z}_{[a_T]})$, RNNs sequentially process each token. Here, one input embedding $\bm{z}_{[a_t]}$ corresponds to one time step $t$ and the hidden state $\bm{h}_t$ is updated at each time step. At each step $t$, the hidden state $\bm{h}_t$ is a function of the hidden state generated in the previous time step, $\bm{h}_{t-1}$, and new input data, $\bm{z}_{[a_t]}$ (see Figure \ref{fig:rnn}) \citep{Elman1990, Amidi2019}. 

The hidden states $\bm{h}_t$, that are passed on and transformed through time, serve as the model's memory (\citealp[p.~182]{Elman1990}; \citealp[p.~32]{Ruder2019}). They capture the information of the sequence that entered until $t$ \citep[p.~391]{Goldberg2016}. Due to this sequential architecture, RNNs theoretically can model dependencies over the entire range of an input sequence \citep{Amidi2019}. But in practice, recurrent models have problems learning dependencies that extend beyond sequences of 10 or 20 tokens \citep[p.~396-399]{Goodfellow2016}. The reason is that when backpropagating the gradients through the time steps (Backpropagation Through Time (BPTT)), the gradients may vanish and thus fail to transmit a signal over long ranges \citep[p.~396-399]{Goodfellow2016}.

The long short-term memory (LSTM) model \citep{Hochreiter1997} extends the RNN with input, output, and forget gates that enable the model to accumulate, remember, and forget provided information \citep[p.~399-400]{Goldberg2016}. This makes LSTMs better suited than the basic RNNs to model dependencies stretching over long time spans \citep[p.~399-400]{Goldberg2016}.

\section{Zero-Shot Learning and the GPT-3} \label{appendix:gpt}
Ultimately the aim of the strand of NLP research focusing on zero-shot learning is to have a model that generalizes well to a wide spectrum of target tasks without being explicitly trained on the target tasks \citep{Radford2019a, Brown2020, Davison2020}. The work on the series of GPT models---OpenAI GPT \citep{Radford2018}, GPT-2 \citep{Radford2019a}, and especially GPT-3 \citep{Brown2020}---has demonstrated that large models that are pretrained on language modeling tasks on excessively large corpora can sometimes come close to achieving acceptable prediction performances without fine-tuning (i.e.~without gradient updates) on target task-specific examples \citep[p.~4]{Brown2020}. So far, the key to increasing the zero-shot no-fine-tuning learning performances seems to be an increase in the models' capacity to learn complex functions as determined by the number of model parameters \citep[p.~4]{Brown2020}. (Whilst the original OpenAI GPT comprises $117$ million parameters, GPT-2 has $1,542$ million \citep[p.~4]{Radford2019a} and GPT-3 has 175,000 million parameters \citep[p.~1, 8]{Brown2020}.) Additionally, and in correspondence with an increase in model parameters, the size of the employed training corpora increases rapidly as well (\citealp[p.~3]{Radford2019a}; \citealp[p.~5]{Brown2020}). Yet given its sheer size, re-training the GPT-3 is prohibitively expensive (\citealp[p.~9]{Brown2020}; \citealp{Riedl2020}). Moreover, whereas typically the source code of pretrained language models is open sourced by the companies (e.g.~Google, Facebook, Microsoft) that developed these models, OpenAI decided not to share the code on GPT-3 and instead allows using GPT-3 for downstream tasks via an API, thereby raising questions regarding accessibility and replicability of pretrained language models for research \citep{Brockman2020, Riedl2020}.

\section{Subword Tokenization Algorithms}  \label{appendix:subword}
Subword tokenization algorithms try to find a balance between word-level tokenization (which tends to result in a large vocabulary---and hence a large embedding matrix that consumes a lot of memory) and character-level tokenization (which generates a small and flexible vocabulary but does not yield as well-performing representations of text) (\citealp[p.~4]{Radford2019a}; \citealp{Huggingface2020b}). Subword tokenization algorithms typically result in vocabularies in which frequently occurring character sequences are merged to form words whereas less common character sequences become subwords or remain separated as single characters (\citealp[p.~4]{Radford2019a}; \citealp{Huggingface2020b}). The Byte-Pair Encoding (BPE) algorithm and variants thereof are subword tokenization algorithms employed in many Transformer-based models \citep[e.g.][]{Devlin2019, Liu2019, Radford2019a}. The base BPE algorithm starts with a list of all the unique characters in a corpus and then learns to merge the characters into longer character sequences (eventually forming subwords and words) until the desired vocabulary size is reached \citep[p.~1717-1718]{Sennrich2016}. In the WordPiece variant of BPE, the algorithm merges at each step the character pair that, when merged, results in the highest increase in the likelihood of the training corpus compared to all other pairs \citep[p.~5150]{Schuster2012}.\\

\section{Pretraining BERT} \label{appendix:pretrainingbert}
In the masked language modeling pretraining task, for each token $q$, that has been sampled for prediction, the updated token representation produced by the last encoder $\bm{h}^*_{q}$ is fed into a single-layer feedforward neural network with a softmax output layer to generate a probability distribution over the terms in the vocabulary predicting the term corresponding to $q$ (see Figure \ref{fig:bert_pretrain} in the main article) (\citeauthor{Alammar2018b}, \citeyear{Alammar2018b}; \citeauthor{Devlin2019}, \citeyear{Devlin2019}, p.~4174). For the next sentence prediction task, the representation for the \emph{[CLS]} token, $\bm{h}^*_{1}$, is processed via a single-layer feedforward neural network with a softmax output to give the predicted probability of the second segment succeeding the first segment (see Figure \ref{fig:bert_pretrain} in the main article) (\citeauthor{Alammar2018b}, \citeyear{Alammar2018b}; \citeauthor{Devlin2019}, \citeyear{Devlin2019}, p.~4174). The loss function in pretraining is the sum of the average loss from the masked language modeling task and the average loss from next sentence prediction \citep[p.~4183]{Devlin2019}.

In order to learn the parameters in pretraining, the authors use the Adam algorithm, a variant of stochastic gradient descent, in which at the $j$th iteration for each individual parameter the estimate of the gradient's average for this parameter is updated based on a parameter-specific learning rate (\citeauthor{Kingma2014}, \citeyear{Kingma2014}; \citeauthor{Devlin2019}, \citeyear{Devlin2019}, p.~4183).\footnote{Here the individual learning rate is inversely proportional to the average of the squared gradient---such that the learning rate is smaller for large gradients and higher for smaller gradients \citep[p.~303-306]{Goodfellow2016}. The gradient's average and the squared gradient's average are exponentially weighted moving averages with decay rates $\beta_1, \beta_2 \in [0,1)$ to assign an exponentially decaying weight to gradients from long ago iterations (\citeauthor{Kingma2014}, \citeyear{Kingma2014}, p.~2; \citeauthor{Goodfellow2016}, \citeyear{Goodfellow2016}, p.~303-306). \citet[p.~4183]{Devlin2019} set $\beta_1$ to $0.9$ and $\beta_2$ to $0.999$.} They use a learning rate schedule in which the global Adam learning rate (that is individually adapted per parameter) linearly increases during the first 10,000 iterations (the warmup) to reach a maximum value of 1e-4 and then is linearly decaying \citep[p.~4183]{Devlin2019}. They furthermore regularize by employing an $L^2$ weight decay (\citealp[p.~226]{Goodfellow2016}; \citealp[p.~4183]{Devlin2019}). As an additional regularization strategy they use dropout \citep{Srivastava2014} with dropout probability $p = 0.1$ \citep[p.~4183]{Devlin2019}. In dropout, units and their corresponding connections are randomly dropped during training \citep[p.~1929]{Srivastava2014}. \citet[p.~4183]{Devlin2019} select a mini-batch size of 256 sequences and conduct 1,000,000 iterations, which implies that they train the model for around $40$ epochs; i.e.~they make around $40$ passes over the entire $3.3$ billion token pretraining data set.

\section{Additional Examples for Autoencoding Models: ALBERT and ELECTRA} \label{appendix:albert}
ALBERT \citep{Lan2020}  aims at a parameter efficient design. By decoupling the size of the input embedding layers from the size of the hidden layers and by sharing parameters across all layers, ALBERT substantially reduces the number of parameters to be learned (e.g.~by a factor of 18 comparing ALBERT-Large to BERT\textsubscript{LARGE}) \citep[p.~2, 4, 6]{Lan2020}. Parameter reduction has regularizing effects, and---because it saves computational resources---allows to construct a deeper model with more and/or larger hidden layers whose increased capacity benefits performance on target tasks while still comprising fewer parameters than the original BERT\textsubscript{LARGE} \citep[p.~2, 7]{Lan2020}.

Whereas BERT, RoBERTa, and ALBERT make use of the masked language modeling task, ELECTRA introduces a new, more resource-efficient pretraining objective, named replaced token detection \citep[p.~1]{Clark2020}. ELECTRA addresses the issue that in masked language modeling for each input sequence predictions are made only for those $15\%$ of tokens that have been sampled for the task, thereby reducing the amount of what could be learned from each training sequence \citep{Clark2020a}. In pretraining, ELECTRA has to predict for each input token in each sequence whether the token comes from the original sequence or has been replaced by a plausible fake token (\citealp{Clark2020a}; \citealp[p.~1, 3]{Clark2020}). Thus, ELECTRA (the discriminator) solves a binary classification task for each token and is much more efficient in pretraining requiring fewer computational resources (\citealp{Clark2020a}; \citealp[p.~3]{Clark2020}). The plausible fake tokens come from a generator that is trained on a masked language modeling task together with the ELECTRA discriminator \citep[p.~3]{Clark2020}. After pretraining, the generator is removed and only the ELECTRA discriminator is used for fine-tuning \citep{Clark2020a}.

\section{Additional Example for an Autoregressive Model: The XLNet} \label{appendix:xlnet}
Strictly speaking, XLNet \citep{Yang2019} is not an autoregressive model \citep{Huggingface2020a}. Yet the permutation language modeling objective that it introduces builds on the autoregressive language modeling framework \citep[p.~5756]{Yang2019}. The authors of XLNet seek a pretraining objective that learns bidirectional representations as in autoencoding models whilst overcoming problems of autoencoding representations: first, the pretrain-finetune discrepancy that results from the fact that \emph{`[MASK]'} tokens only occur in pretraining, and, second, the assumption that the tokens selected for the masked language modeling task in one sequence are independent of each other \citep[p.~5754-5755]{Yang2019}. Given a sequence whose tokens are indexed $(1, \dots, T)$, the permutation language modeling objective makes use of the permutations of the token index $(1, \dots, T)$ \citep[p.~5756]{Yang2019}. For each possible permutation of $(1, \dots, T)$, the task is to predict the next token in the permutation order given the previous tokens in the permutation \citep[p.~5756]{Yang2019}. In doing so, the learned token representations can access information from left and right contexts whilst the autoregressive nature of the modeling objective avoids the pretrain-finetune discrepancy and the independence assumption \citep[p.~5756]{Yang2019}.

\section{Examples for Sequence-to-Sequence Models: The T5 and BART} \label{appendix:seqseq}
The T5 \citep{Raffel2020} is very close to the original Transformer encoder-decoder architecture. It is based on the idea to consider all NLP tasks as text-to-text problems \citep[p.~2-3]{Raffel2020}. To achieve this, each input sequence that is fed to the model is preceded by a task-specific prefix, that instructs the model what to do. For example \citep[p.~47ff.]{Raffel2020}: A translation task in this scheme has the input \emph{`translate from English to German: I love this movie.'} and the model is trained to output \emph{`Ich liebe diesen Film.'}. For a sentiment classification task on the SST-2 Dataset \citep{Socher2013}, the input would be: \emph{`sst2 sentence: I love this movie.'} and the model is trained to predict one of \emph{`positive'} or \emph{`negative'}. The fact that there is a shared scheme for all NLP tasks, allows the T5 to be pretrained on a multitude of different NLP tasks before being fine-tuned on a specific target task \citep[p.~30-33]{Raffel2020}. In the multitask pretraining mode, T5 is trained on a self-supervised objective similar to the masked language modeling task in BERT as well as various different supervised tasks (such as translation or natural language inference) \citep[p.~37]{Raffel2020}. With this multitask pretraining setting, in which the parameters learned in pretraining are shared across different tasks, the T5, rather than being a standard sequential transfer learning model, implements a softened version of multitask learning \citep[p.~30]{Raffel2020}.

BART \citep{Lewis2020}---another well-known sequence-to-sequence model---is composed of an encoder and a decoder. Just as an autoencoding model, BART in pretraining is presented with a corrupted sequence and has to predict the original uncorrupted sequence \citep[p.~2]{Lewis2020}. In pretraining, BART allows a wide range of different types of corrupting operations to be applied to the documents \citep[p.~2-3]{Lewis2020}. Due to this autoencoding-style pretraining task, BART can be considered a BERT-like bidirectional encoder followed by an autoregressive unidirectional decoder \citep[p.~1-2]{Lewis2020}. Because of the decoder, that learns to predict output tokens in an autoregressive manner, BART is better suited to perform text generation tasks than regular autoencoding models \citep[p.~1, 6]{Lewis2020}. Additionally, BART performs similarly to RoBERTa on discriminative natural language understanding tasks \citep[p.~1, 6]{Lewis2020}.

\section{Interpretability}  \label{appendix:interpret}
One common method to make neural networks more interpretable is probing (also known as auxiliary prediction tasks) \citep[p.~51]{Belinkov2019}. In probing, an element of a trained neural network (e.g.~a set of contextualized token embeddings) is extracted, fixed, and fed into a simple classifier and then is applied to some task (e.g.~POS tagging, coreference resolution) (\citealp[p.~51]{Belinkov2019}; \citealp[p.~1-3]{Tenney2019a}). If the prediction performance on the task is high, then this is taken as an indication that information required to address the task (e.g.~syntactic information for POS tagging, semantic information for coreference resolution) is encoded in the tested element of the network (\citealp[p.~51]{Belinkov2019}; \citealp[p.~2]{Tenney2019a}). Accordingly, probing is one way to inspect what information a neural network has learned and which elements capture which information.

Another important aspect of interpretability is to assess the importance of input features for predicted outputs. The open-source library Captum (https://captum.ai/) implements several attribution algorithms that allow just that \citep[p.~3]{Kokhlikyan2020}. Additionally, algorithms for attributing outputs to a hidden layer, as well as algorithms for attributing hidden layer values to feature inputs, are also provided \citep[p.~3]{Kokhlikyan2020}.

Most attribution algorithms can be considered as either being gradient-based or perturbation-based \citep[p.~110]{Agarwal2021}. Perturbation-based algorithms make use of removed or altered input features to learn about the importance of input features (\citealp[p.~2]{Ancona2018}; \citealp[p.~110]{Agarwal2021}). Gradient-based algorithms make use of the gradient of the predicted output with regard to input features (\citealp[p.~2-3]{Ancona2018}; \citealp[p.~110]{Agarwal2021}).

For models that incorporate attention mechanisms, the analysis of patterns in attention weights $\alpha_{t,t^*}$ and the probing of attention heads is another interpretability-related research area \citep[see e.g.][]{Clark2019a, Kobayashi2020}. Tools for interpreting attention matrices are also provided by Captum (see, for example, the tutorial at https://captum.ai/tutorials/Bert\_SQUAD\_Interpret2).

Another set of methods related to interpretability is behavioral testing and the construction of adversarial examples \citep[p.~54-58]{Belinkov2019}.\footnote{In NLP, an adversarial example is a text sequence $d_i^*$ that is close to a sequence $d_i$ but has a different class label than $d_i$ \citep[p.~56]{Belinkov2019}.} Here, the goal is to inspect a trained model's behavior when confronted with a challenging set of inputs or adversarial examples. In an award-winning paper, \citet{Ribeiro2020} present a methodology for behavioral testing. Their research findings emphasize that whilst the performance of NLP models as evaluated via accuracy measures on held-out test sets has risen substantially during the last years \citep[see also e.g.][p.~2]{Wang2019a}, when evaluating the models via behavioral testing, it is revealed that accuracy-based performances on common benchmark data sets overestimate the models' linguistic and language understanding capabilities. BERT, for example, is found to have high failure rates for simple negation tests (e.g.~classifying $84.4\%$ of positive or neutral tweets in which a negative sentiment expression is negated into the negative category) \citep[p.~4905-4907]{Ribeiro2020}.

\section{Deep Learning and Transfer Learning in Practice}  \label{appendix:practice}

To practically implement deep learning models, it is advisable to have access to a graphics processing unit (GPU). In contrast to a central processing unit (CPU), a GPU comprises many more cores and can conduct thousands of operations in parallel \citep{Caulfield2009}. GPUs thus handle tasks that can be broken down into smaller, simultaneously executable subtasks much more efficiently than CPUs \citep{Caulfield2009}. When training a neural network via stochastic gradient descent, every single hidden unit within a layer usually can be updated independently of the other hidden units in the same layer \citep[p.~440]{Goodfellow2016}. Hence, neural networks lend themselves to parallel processing.

A major route to access and use GPUs is via NVIDIA's CUDA framework \citep[p.~440-441]{Goodfellow2016}. But instead of additionally learning how to write CUDA code, researchers use libraries that enable CUDA GPU processing \citep[p.~441]{Goodfellow2016}. As of today, PyTorch \citep{Paszke2019} and TensorFlow \citep{Abadi2015} are the most commonly used libraries that allow training neural networks via CUDA-enabled GPUs. Both libraries have Python interfaces. Therefore, to efficiently train deep learning models via GPU acceleration, researchers can use a programming language they are familiar with.

Another obstacle is having a GPU at hand that can be used for computation. The computing infrastructures of universities and research institutes typically provide their members access to GPU facilities. Free GPU usage also is available via Google Colaboratory (or Colab for short): https://colab.research.google.com/notebooks/intro.ipynb. Colab is a computing service that allows its user to run Python code via the browser \citep{GoogleColab2020}. Here, GPUs can be used free of cost. The free resources, however, are not guaranteed and there may be usage limits. One issue researchers have to keep in mind when using Colab is that at each session another type of GPU may be assigned. Documenting the used computing environment hence is vital to ensure traceability. Note, that full reproducibility \emph{across} different computing platforms and \emph{across} different versions of PyTorch and TensorFlow cannot be guaranteed \citep{NVIDIA2021, PyTorchContributors2021}. However, there are measures that researchers can undertake to minimize nondeterministic elements (e.g.~not using nondeterministic algorithms where possible and ensuring that batch allocation is reproducible) \citep[see][]{PyTorchContributors2021}.

\section{Application: Zero-Shot Learning}  \label{appendix:zslimpl}

To explore how pretrained Transformer-based models would perform in a zero-shot learning setting, the approach of \citet{Yin2019} is followed. \citet[p.~3918-3919]{Yin2019} frame zero-shot text classification as a natural language inference (NLI) task. In NLI, a model is presented with a premise and a hypothesis and then has to decide whether the hypothesis is true given the premise (\textsc{entailment}), whether the hypothesis is false given the premise (\textsc{contradiction}), or whether the hypothesis is neither true nor false given the premise (\textsc{neutral}) (see Table \ref{tab:zsl}) \citep[p.~1112-1113]{Williams2018}.

\begin{table}[htb]
\begin{center}
\begin{small}
\begin{tabular}{l|l|l}
\toprule
 &  \multirow{2}{11.1cm}{[CLS] \hspace{0.8cm} ... Premise ...  \hspace{0.8cm}  [SEP]  \hspace{0.5cm}  ... Hypothesis ... \hspace{0.65cm} [SEP]} &  \multirow{2}{2.1cm}{predict one out of}\\
& & \\
\midrule
 \multirow{3}{0.6cm}{\textbf{NLI}}   &  \multirow{3}{11.1cm}{[CLS]  \hspace{0.5mm} I am a lacto-vegetarian. \hspace{0.005mm} [SEP]  \hspace{0.5mm} I eat one egg per week. \hspace{0.5mm} [SEP]} &  \multirow{3}{2.1cm}{{\footnotesize{\textsc{entailment}}}, {\footnotesize{\textsc{contradict.}}}, {\footnotesize{\textsc{neutral}}}} \\
& & \\
 & & \\
\hline
 \multirow{6}{0.6cm}{\textbf{ZSL}}  &  \multirow{3}{11.1cm}{[CLS] Ok, I see what you mean. [SEP]  \hspace{0.5mm} This comment is \underline{toxic}.  \hspace{1mm} [SEP]} & \multirow{3}{2.1cm}{{\footnotesize{\textsc{entailment}}}, {\footnotesize{\textsc{contradict.}}}, {\footnotesize{\textsc{neutral}}}}\\
 & & \\
 & & \\
  \cdashline{2-3}
&   \multirow{3}{11.1cm}{[CLS] Ok, I see what you mean. [SEP] This comment is \underline{not toxic}. [SEP]}  & \multirow{3}{2.1cm}{{\footnotesize{\textsc{entailment}}}, {\footnotesize{\textsc{contradict.}}}, {\footnotesize{\textsc{neutral}}}}\\
& & \\
 & & \\
\bottomrule
\end{tabular}
\end{small}
\end{center}
\caption[Scheme in NLI and ZSL.]{\textbf{Scheme in Natural Language Inference (NLI) and Zero-Shot Learning (ZSL).} \small{In NLI, the model is provided with a premise followed by a hypothesis and has to decide whether the hypothesis is true (\textsc{entailment}), false (\textsc{contradiction}), or neutral (\textsc{neutral}) given the premise. In ZSL, the sequence for which a prediction is to be made constitutes the premise. Each class label is presented to the model as a separate hypothesis. Here, the input sequence is ``Ok, I see what you mean.'' and there are $C=2$ class labels, namely: $\{$toxic, not toxic$\}$. For each sequence-hypothesis pair the model has to predict one out of $\{$\textsc{entailment}, \textsc{contradiction}, \textsc{neutral}$\}$.}}
\label{tab:zsl}
\end{table}

In the zero-shot classification framework of \citet{Yin2019}, a model is presented with the input text (taking the role of the premise) and a hypothesis that asks whether the input text belongs to a particular class. The model then has to predict whether this is the case or not. The model that \citet{Yin2019} use for zero-shot learning is a BERT model that has been trained on three different NLI data sets \citep[p.~3919]{Yin2019}. (The NLI data sets are, of course, unrelated to the target tasks \citet{Yin2019} use for zero-shot learning evaluation.)

Here, in a similar approach, two pretrained Transformer-based models---RoBERTa \citep{Liu2019} and BART \citep{Lewis2020}---that have been further trained on the Multi-Genre Natural Language Inference (MNLI) data set \citep{Williams2018} are used as models for zero-shot learning. The models are accessed from Hugging Face's Transformers and then are used in a zero-shot-classification pipeline that is based on an NLI-framework \citep[see][]{Davison2020a}. For an illustration see Table \ref{tab:zsl}. 

An important point to note is that the zero-shot performance will also depend on the textual formulation of the hypothesis the model is presented with \citep[p.~3921-3922]{Yin2019}. (The model takes as an input the text sequence for which a prediction is to be made followed by the hypothesis and thus the model learns representations for the input sequence and the hypothesis and will generate a prediction based on (the compatibility) of both inputs \citep{Davison2020}.) Here, to explore the effect of different hypothesis formulations, two different hypothesis formulations are tried in each application (see Tables \ref{tab:zslres1} and \ref{tab:zslres2}). 

Moreover, note that if one has a target task with $C$ class labels such that $y_i \in \{\mathcal{G}_1, \dots, \mathcal{G}_c, \dots, \mathcal{G}_C\}$, then each class label is presented to the model as one separate hypothesis \citep{Davison2020a}. Consequently, if there are, for example, $C = 4$ labels, the model will be fed with $C = 4$ different sequence-hypothesis pairs for each sequence. In the implementation in Hugging Face's zero-shot-classification pipeline, the model generates for each of the $C$ sequence-hypothesis pairs a prediction to belong to one of $\{$\textsc{entailment}, \textsc{contradiction}, \textsc{neutral}$\}$ \citep{Davison2020a}. Then, in order to aggregate the $C$ separate predictions into a single one, the predicted score for \textsc{entailment} is extracted for each hypothesis. Together, the $C$ entailment scores serve as the input to a softmax function that returns a $C$-dimensional vector of predicted probabilities (which sum to one) and the $c$th element gives the probability that the sequence belongs to the $c$th class \citep{Davison2020a}.

The zero-shot classification results are presented in Tables \ref{tab:zslres1} and \ref{tab:zslres2}. For each application, for each combination of an employed pretrained model and an explored hypothesis formulation, the macro-averaged $F_1$-Score for the test set is reported. Across applications, models, and hypothesis formulations, the achieved performance levels are mediocre compared to the macro-averaged $F_1$-Scores reached by the fine-tuned models (which are presented in Table \ref{tab:validation_results} in the main article). Interestingly, the smallest reduction in performance compared to the fine-tuned models can be observed for the Abortion application. In contrast to the other two applications, that seek to measure concepts from text that are relatively difficult to adequately describe in words (ethos, toxicity), hypothesis formulation in the legalization of abortion application is relatively straightforward.

{\bfseries\sffamily{Source Code.}} Just as for the other analyses presented in this paper, the code for this zero-shot learning implementation is openly available in figshare at \url{https://doi.org/10.6084/m9.figshare.14394173}.

\begin{table}[H]
\begin{center}
\begin{small}
{\large{\textbf{Ethos}}} \\
\vspace{0.2cm}
\begin{tabular}{|l|l|l|l|}
\toprule
        \textbf{Model} & \textbf{Hypothesis Formulation} & \textbf{Class Labels} & \textbf{$F_1$ (macro)} \\ 
        \midrule
       \multirow{3}{1.5cm}{RoBERTa} & \multirow{3}{7cm}{The statement \{\} the ethos of another politician or party.} & \{attacks,   &  \\
    &    & does not refer to, & \maxf{0.200}   \\
& & supports\} &  \\
        \hline
        \multirow{3}{1cm}{RoBERTa}  & \multirow{3}{7cm}{The statement expresses \{\} sentiment toward the character of another politician or party.} & \{negative,  & \multirow{3}{1.5cm}{0.188} \\
     &   & no,  & \\
& & positive\} & \\
        \hline
        \multirow{3}{1cm}{BART} & \multirow{3}{7cm}{The statement \{\} the ethos of another politician or party.} & \{attacks,  & \multirow{3}{1.5cm}{0.184} \\ 
      &  & does not refer to, & \\
& & supports\} & \\
 \hline
       \multirow{3}{1cm}{BART} & \multirow{3}{7cm}{The statement expresses \{\} sentiment toward the character of another politician or party.} & \{negative, & \multirow{3}{1.5cm}{0.181} \\
      &  & no,  & \\
 & & positive\} & \\
\bottomrule
\end{tabular}\\
%
%
%
\vspace{0.7cm}
{\large{\textbf{Abortion}}} \\
\vspace{0.2cm}
    \begin{tabular}{|l|l|l|l|}
\toprule
        \textbf{Model} & \textbf{Hypothesis Formulation} & \textbf{Class Labels} & \textbf{$F_1$ (macro)} \\ 
        \midrule
         \multirow{3}{1.5cm}{RoBERTa} & \multirow{3}{7cm}{The text is \{\} legalization of abortion.} & \{in favor of, &  \\ 
         & & neutral toward, & 0.344 \\
         & & against\}  & \\
     \hline
        \multirow{3}{1cm}{RoBERTa}  & \multirow{3}{7cm}{The text expresses a \{\} stance toward legalization of abortion.} & \{positive, &  \\ 
        & & neutral, & 0.362 \\
        & & negative\} & \\
        \hline
         \multirow{3}{1cm}{BART} & \multirow{3}{7cm}{The text is \{\} legalization of abortion.} & \{in favor of, &  \\ 
         & & neutral toward, & \maxf{0.455} \\
         & & against\} & \\
         \hline
         \multirow{3}{1cm}{BART} & \multirow{3}{7cm}{The text expresses a \{\} stance toward legalization of abortion.} & \{positive, & \\
         & & neutral, &  0.368 \\
         & & negative\} & \\
        \bottomrule
    \end{tabular}
 \end{small}
\end{center}
\caption[ZSL Results I.]{\textbf{ZSL Results I.} \small{Macro-averaged $F_1$-Scores obtained via zero-shot learning for the test sets of the Ethos and Abortion classification tasks. In each application, two pretrained models (RoBERTa and BART both trained on the MNLI) and two hypothesis formulations are explored. Gray colored numbers highlight the best performing model-hypothesis formulation combination for the task.}}
\label{tab:zslres1}
\end{table}
\newpage

\begin{table}[H]
\begin{center}
\begin{small}
{\large{\textbf{Toxic}}} \\
\vspace{0.2cm}
    \begin{tabular}{|l|l|l|l|}
\toprule
        \textbf{Model} & \textbf{Hypothesis Formul.} & \textbf{Class Labels} & \textbf{$F_1$ (macro)} \\ 
        \midrule
         \multirow{3}{1.5cm}{RoBERTa} & \multirow{3}{3.5cm}{This comment is \{\}.} & \multirow{3}{6cm}{\{not toxic, toxic\}} &\\
        && & \maxf{0.470} \\
        && &\\
          \hline
         \multirow{6}{1.5cm}{RoBERTa} & \multirow{6}{3.5cm}{This comment is \{\}.} & \multirow{6}{6cm}{\{\{neither obscene, nor threatening, nor insulting, and does not expresses hatred toward social groups and identities\}, \{obscene, or threatening, or insulting, or expresses hatred toward social groups and identities\}\}} & \multirow{6}{1.5cm}{0.213} \\
        &&&\\
        &&&\\
          &&&\\
             &&&\\
          &&&\\
          \hline
        \multirow{4}{1cm}{BART} & \multirow{4}{3.5cm}{This comment is \{\}.} & \multirow{4}{6cm}{\{not toxic, toxic\}}  & \multirow{4}{1.5cm}{0.378} \\
        && &\\
        &&  &\\
          &&&\\
          \hline
       \multirow{6}{1cm}{BART} & \multirow{6}{3.5cm}{This comment is \{\}.} & \multirow{6}{6cm}{\{\{neither obscene, nor threatening, nor insulting, and does not expresses hatred toward social groups and identities\}, \{obscene, or threatening, or insulting, or expresses hatred toward social groups and identities\}\}} & \multirow{6}{1.5cm}{0.178} \\
        &&&\\
        &&&\\
           &&&\\
          &&&\\
           &&&\\
  \bottomrule
    \end{tabular}
     \end{small}
\end{center}
\caption[ZSL Results II.]{\textbf{ZSL Results II.} \small{Macro-averaged $F_1$-Scores obtained via zero-shot learning for one sampled test set ($N$ = 1,000) of the Toxic classification task. Two pretrained models (RoBERTa and BART both trained on the MNLI) and two hypothesis formulations are explored. Gray-colored numbers highlight the best performing model-hypothesis formulation combination for the task.}}
\label{tab:zslres2}
\end{table}

\newpage
\bibliography{Paper03a097_chapterdiss10}
\bibliographystyle{apalike-me2}

\end{document}